\pdfoutput=1
\documentclass[twocolumn]{wlscirep}

\usepackage{dsfont}
\usepackage{mathtools}

\usepackage{graphicx}%
\usepackage{multirow}%
\usepackage{amsmath,amssymb,amsfonts}%
\usepackage{amsthm}%
\usepackage{mathrsfs}%
\usepackage[title]{appendix}%
\usepackage{xcolor}%
\usepackage{textcomp}%
\usepackage{manyfoot}%
\usepackage{booktabs}%
\usepackage{algorithm}%
\usepackage{algorithmicx}%
\usepackage{algpseudocode}%
\usepackage{listings}%
\usepackage{placeins}
\usepackage{svg}
\usepackage{csquotes}

\usepackage[backend=biber,style=numeric, autocite=superscript,url=false]{biblatex}
\usepackage{caption}
\newcommand{\suppnote}[1]{Supplementary Note~\ref{#1}}
\newcommand{\suppnotesub}[1]{Supplementary Note~\ref{#1}, \emph{\nameref{#1}}}
\begin{document}

\title{Joint Surrogate Learning of Objectives, Constraints, and Sensitivities for Efficient Multi-objective Optimization of Neural Dynamical Systems}
\date{}

\author[1]{Frithjof Gressmann}

\author[2]{Ivan Georgiev Raikov}

\author[3]{Seung Hyun Kim}

\author[3,4,5]{Mattia Gazzola}

\author[1]{Lawrence Rauchwerger}

\author[2]{Ivan Soltesz}

\makeatletter
\renewcommand\AB@affilsepx{, \protect\Affilfont}
\makeatother

\affil[1]{\footnotesize{Siebel School of Computing and Data Science, University of Illinois Urbana-Champaign, Urbana, IL}}

\affil[2]{\footnotesize{Department of Neurosurgery, Stanford University, Stanford, CA}}

\affil[3]{\footnotesize{Carl R. Woese Institute for Genomic Biology, University of Illinois Urbana-Champaign, Urbana, IL}}

\affil[4]{\footnotesize{Mechanical Science and Engineering, University of Illinois Urbana-Champaign, Urbana, IL}}

\affil[5]{\footnotesize{National Center for Supercomputing Applications, University of Illinois Urbana-Champaign, Urbana, IL. \, Corresponding author: Frithjof Gressmann  \textless \href{mailto:fg14@illinois.edu}{fg14@illinois.edu}\textgreater}.}

\begin{abstract}

Gaussian process surrogates dominate constrained multi-objective optimization because they are effective in data-scarce regimes, but their cubic scaling in training samples limits their ability to capture shared structure between objectives and constraints as problems grow in dimensionality.
We show that deterministic neural network surrogates, equipped with feature tokenization and adaptive output normalization, match or exceed Gaussian process accuracy, while scaling to high-dimensional output spaces and training on all data including infeasible samples. Jointly training a single Feature Tokenizer Transformer to predict objectives, constraint satisfaction, and parameter sensitivities yields a unified gradient that simultaneously improves objective values, steers toward feasibility, and identifies the most influential parameters: a coherent search signal that disjoint per-output models cannot provide.
We validate this on biophysical neural optimization problems of increasing complexity. In the hardest regime, with wide, uninformed parameter bounds where random sampling finds zero feasible solutions, descending the surrogate's learned constraint gradient steers the search into the feasible region and recovers near-optimal solutions where standard surrogate optimization and constrained Bayesian optimization find none.
\end{abstract}

\keywords{Multi-objective optimization, Surrogate models, Constraint optimization, Neural networks, Neuroscience}

\maketitle

\section{Introduction}\label{sec:introduction}

Computational models of neural circuits are governed by many biophysical parameters that cannot all be determined analytically or experimentally, but must be inferred through data-driven optimization~\autocite{prinzAlternativeHandTuningConductanceBased2003,markramReconstructionSimulationNeocortical2015, bezaireInterneuronalMechanismsHippocampal2016,dura-bernalNetPyNEToolDatadriven2019}. These simulations of chaotic and tightly coupled dynamics are routinely among the most computationally demanding scientific applications, challenging even the most sophisticated optimization algorithms~\autocite{druckmannNovelMultipleObjective2007,vangeitBluePyOptLeveragingOpen2016}. Moreover, experimentalists often characterize acceptable network behavior through constraints (e.g.~bounds on firing rates or oscillation frequencies) rather than precise target values, but the resulting binary feasible/infeasible partition provides no gradient signal, and the combination of high dimensionality, expensive evaluations, and narrow feasibility regions renders standard optimization approaches intractable. 

Surrogate-assisted optimization addresses the evaluation bottleneck by replacing the original problem with computationally inexpensive models during search~\autocite{jonesEfficientGlobalOptimization1998,jinSurrogateassistedEvolutionaryComputation2011}, and has been extended to multi-objective~\autocite{knowlesParEGOHybridAlgorithm2006,voutchkov_multi-objective_2010,chughSurrogateAssistedReferenceVector2018} and constrained~\autocite{gardnerBayesianOptimizationInequality2014,gelbartBayesianOptimizationUnknown2014,hernandez-lobatoPredictiveEntropySearch2015} settings. 
The surrogate can drive the search in different ways. In Bayesian optimization (BO), a probabilistic surrogate like a Gaussian Process (GP) generates new samples by maximizing an acquisition function that trades the posterior mean against its variance. Constraint information can be incorporated by weighting the acquisition by the predicted feasibility of the samples~\autocite{gardnerBayesianOptimizationInequality2014,gelbartBayesianOptimizationUnknown2014,hernandez-lobatoPredictiveEntropySearch2015}. Evolutionary methods, on the other hand, do not require a probabilistic model and instead proceed by minimizing the surrogate's point prediction using population-based black-box optimization~\autocite{jinSurrogateassistedEvolutionaryComputation2011}. 
In both settings, GP surrogates remain the dominant choice due to their data efficiency~\autocite{shahriariTakingHumanOut2015,balandatBoTorchFrameworkEfficient2020}, but their cubic scaling in training samples limits their ability to capture shared structure between objectives and constraints as problems grow in dimensionality. The acquisition step of BO poses a further, structural limitation in that it produces one candidate per acquisition function optimization, forcing BO to proceed sequentially in small batches~\autocite{wilsonMaximizingAcquisitionFunctions2018,daultonParallelBayesianOptimization2021}. Population-based sampling by contrast is easier to parallelize and scale, but has lacked a principled way to incorporate constraint information. More broadly, the three core surrogate capabilities---objective prediction, constraint estimation, and sensitivity analysis---are still developed and deployed largely in isolation, forcing practitioners to rely on ad-hoc heuristics for combining surrogate information~\autocite{bhosekarAdvancesSurrogateBased2018}, for instance by re-ranking the candidates proposed by a black-box optimizer with a separately fitted feasibility classifier.

Neural network surrogates can unify these capabilities through multi-task learning and shared representations. Unlike the \emph{probabilistic} neural surrogates developed for Bayesian optimization~\autocite{snoek_scalable_2015,springenberg_bayesian_2016,garnelo_neural_2018}, a \emph{deterministic} network yields point estimates rather than a posterior, which is all a population-based black-box optimizer requires. They have nonetheless remained underexplored in this data-scarce setting, under the assumption that neural networks demand prohibitively large training sets~\autocite{goodfellow_deep_2016}.

\begin{figure*}[h]
\centering
\includegraphics[width=1\textwidth]{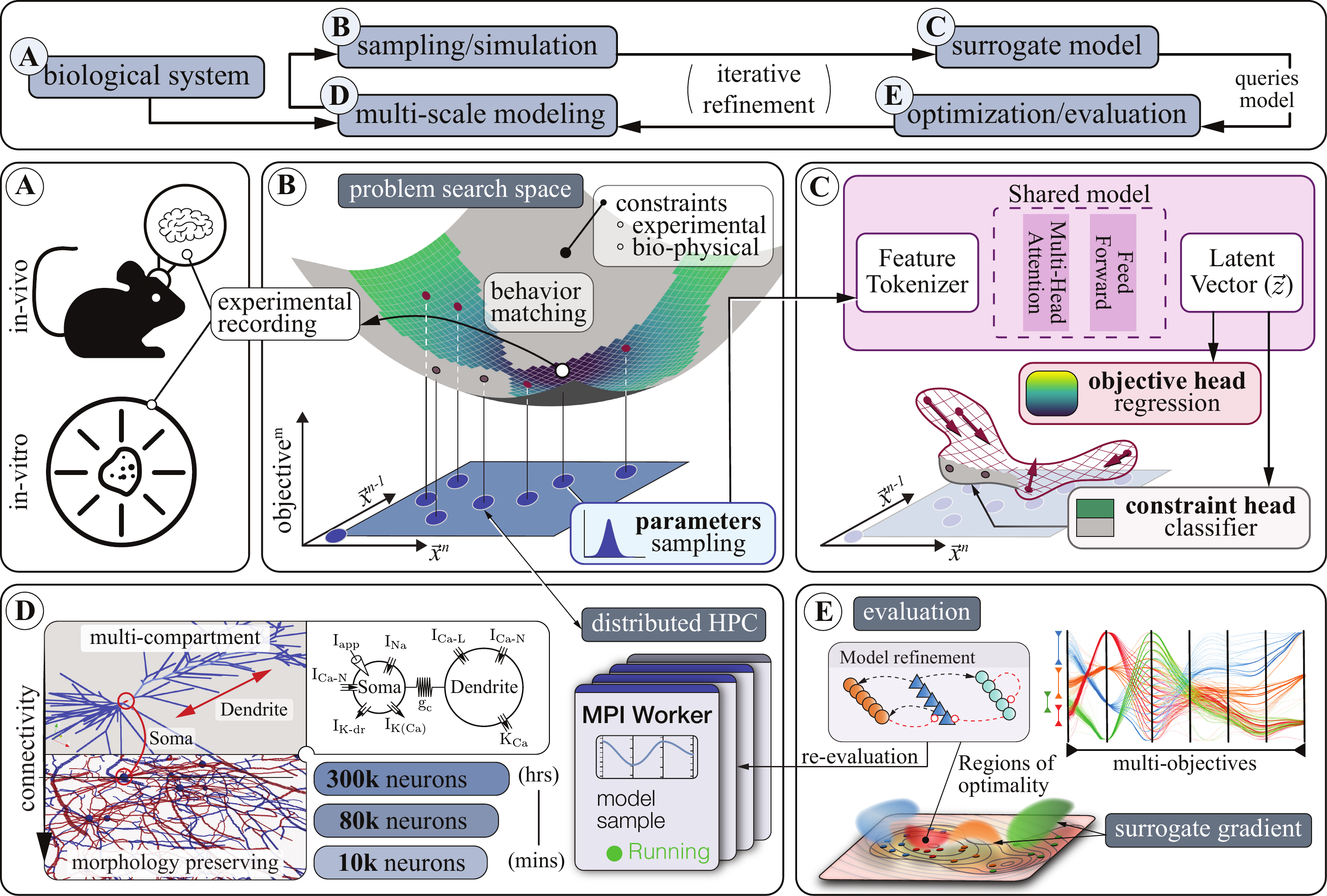}
\caption{\textbf{Surrogate-assisted optimization for multi-scale neural modeling at supercomputing scale.} Top: overview of the iterative pipeline connecting biological data acquisition \textbf{(A)}, parameter sampling and simulation \textbf{(B)}, surrogate modeling \textbf{(C)}, multi-scale neural simulation \textbf{(D)}, and optimization/evaluation \textbf{(E)}. \textbf{A}, Experimental recordings are obtained from in vivo and in vitro preparations of biological systems. \textbf{B}, Reproducing these observations requires efficient exploration of a high-dimensional parameter space subject to multiple objectives (behavior matching) and constraints (experimental and biophysical). \textbf{C}, A transformer-based surrogate model maps parameter configurations to a shared latent representation and jointly predicts objectives (regression head) and constraint satisfaction (classification head), yielding a differentiable approximation of the search landscape. \textbf{D}, Multi-compartment neuron models with detailed ion-channel dynamics are assembled into morphology-preserving networks and evaluated in parallel on distributed high-performance computing infrastructure via MPI workers. \textbf{E}, The surrogate enables iterative model refinement by identifying regions of optimality across multiple objectives.}\label{fig:overview}
\end{figure*}

Here, we show that a \emph{jointly learned, differentiable} neural network surrogate resolves the integration challenge: its gradients provide a smooth, unified signal that simultaneously steers toward improved objective values and greater constraint satisfaction, while enabling sensitivity-informed exploration. We implement this approach in \textsc{dmosopt}, an open-source framework for \textbf{D}istributed \textbf{M}ulti-\textbf{O}bjective \textbf{S}urrogate-Assisted \textbf{Opt}imization designed for large-scale problems with out-of-the-box HPC scalability via MPI (Methods; \href{https://github.com/dmosopt/dmosopt}{github.com/dmosopt}; Fig.~\ref{fig:overview}). The surrogate gradient is supplied as an informed search direction to a multi-objective evolutionary algorithm (MOEA), whose population diversity, mutation, and recombination operators provide exploratory pressure complementary to the surrogate's exploitation signal.

To validate the framework across the stages of a realistic modeling workflow, we apply it to progressively more challenging neuroscience optimization problems, from individual neuron models~\autocite{boothCompartmentalModelVertebrate1997,milesFunctionalPropertiesMotoneurons2004,bezaireInterneuronalMechanismsHippocampal2016} to a large-scale hippocampal network~\autocite{bezaireInterneuronalMechanismsHippocampal2016}, demonstrating how joint learning and surrogate-model gradients (1)~improve optimization compared to disjoint surrogates, (2)~render otherwise-intractable, highly constrained problems solvable by descending the surrogate's constraint gradient into feasible regions, and (3)~enable the optimization of large-scale network models with minimal evaluations.
While motivated by computational neuroscience, our techniques are domain-agnostic and readily transferable to constrained multi-objective optimization problems across science and engineering.

\section{Results}\label{sec:results}

Our approach centers on establishing joint surrogate model \(f\colon \mathds{R}^{n} \rightarrow \mathds{R}^{q + k}\), differentiable by construction, that simultaneously predicts \(q\) objective values \(\mathbf{y}\) and \(k\) constraint outcomes \(\mathbf{c}\) for a given parameter vector \(\mathbf{x}\).
We leverage the gradient of $f$
\begin{align}
    \mathbf{g}_\text{sopt} = \nabla_{\mathbf{x}} [  \underbrace{f_\mathbf{y}(\mathbf{x})}_\text{Objective} + \underbrace{(f_{\mathbf{c}}(\mathbf{x}) - \mathds{1})}_\text{Constraint feasibility}  ]_{\mathbf{x} = \mathbf{\bar{x}}_i}
\label{eq:sopt_grad}
\end{align}
that guides toward feasible regions of lower objective value. 
In practice, the objective and constraint terms are substituted by statistical metrics: a hypervolume-based loss and binary focal cross-entropy, with norm-balanced gradients (\suppnote{sec:supp-gradient}). The partial derivatives \(\tfrac{\partial f}{\partial x_j}\) yield parameter sensitivity estimates at no extra cost. We parameterize $f$ as a \emph{deterministic} deep neural network producing point predictions, which scales readily to large evaluation budgets and high-dimensional, multi-output settings in contrast to the Gaussian process surrogates. Rather than optimizing $f$ directly, which risks convergence to regions where the model is confidently wrong~\autocite{jinSurrogateassistedEvolutionaryComputation2011}, we supply \(\mathbf{g}_\text{sopt}\) as an informed search direction to a multi-objective evolutionary algorithm (MOEA), whose population diversity, mutation, and recombination operators provide the exploratory pressure that posterior uncertainty provides in Bayesian optimization, and which proposes an entire population of candidates per epoch for concurrent evaluation. All algorithmic details are reported in the Methods.

In the following, the capabilities of \textsc{dmosopt} are demonstrated on a suite of progressively more challenging neuroscientific problems ranging from individual neuron dynamics to a full-scale hippocampal network simulation, establishing each component of the framework in turn.

\subsection{Neural network surrogates outperform baselines on single-cell optimization}

\begin{figure*}[h]
\centering
\includegraphics[width=0.95\textwidth]{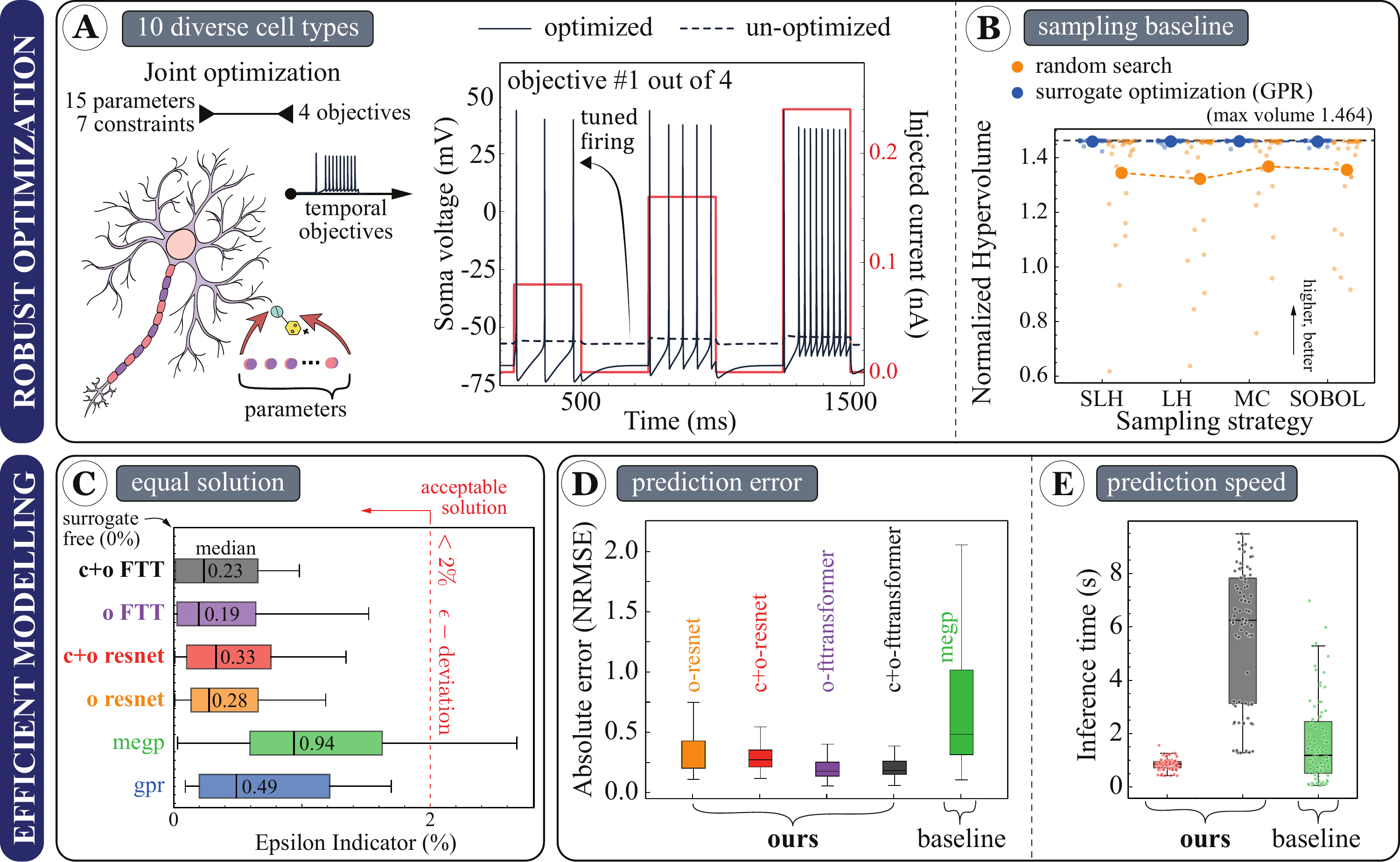}
\caption{\textbf{Neural network surrogate performance on single-cell optimization.} \textbf{A}, To benchmark surrogate performance across diverse dynamical systems, 9 morphologically distinct CA1 interneuron types are jointly optimized over 10-15 parameters against 4 electrophysiological objectives (e.g., somatic firing response to current injection, objective \#1 shown) subject to 7-8 constraints. Solid and dashed traces show optimized and un-optimized voltage responses, respectively. \textbf{B}, Standard space-filling sampling strategies (symmetric Latin hypercube, SLH; Latin hypercube, LH; Monte Carlo, MC; Sobol sequences) fail to approach the theoretical maximum normalized hypervolume, whereas Gaussian process regression (GPR)-based surrogate optimization consistently reaches it. \textbf{C}, All tested neural network surrogates (ResNet and FTTransformer variants with objective-only ``o-'' or joint constraint-and-objective ``c+o''-heads) achieve Pareto-front $\varepsilon$-indicator values well within the 2\% acceptable-solution threshold and outperform GPR and multitask exact Gaussian process (MEGP) baselines. \textbf{D}, This improvement is explained by lower absolute prediction error (normalized root-mean-square error, NRMSE) of the neural network architectures relative to the GP baselines throughout the optimization. \textbf{E}, ResNet variants achieve higher accuracy at lower inference cost than the GP baselines; the FTTransformer incurs moderately higher inference times but delivers the largest accuracy gains. Since accuracy gains translate to faster convergence, reduced evaluations relative to GPs outweigh any increase in inference overhead across all configurations. Statistics: in \textbf{B}, markers show the mean and error bars $\pm 1$ SEM over $n = 27$ optimization runs per strategy (9 populations $\times$ 3 independent trials with distinct random seeds), and small dots the individual runs. In \textbf{C}, boxplot of the $\varepsilon$-indicator for $n = 27$ runs per method (9 populations $\times$ 3 trials). In \textbf{D}, boxplot across 81 fits (9 populations $\times$ 9 epoch checkpoints of one completed optimization run). In \textbf{E}, boxplot and dots show the individual timings of all $n = 81$ fits per model.}
\label{fig:figure2-surrogate-opt}
\end{figure*}

We begin by considering the optimization of individual CA1 hippocampal interneuron models, each having a distinct electrophysiological fingerprint that must be faithfully reproduced as a prerequisite for assembling a realistic network model. Because individual single-cell simulations are computationally tractable even without surrogates, this problem class also provides a suitable testbed for benchmarking surrogate approaches before scaling to the full network. To this end, we jointly optimized 9 morphologically distinct CA1 hippocampal interneuron types using a two-compartment Pinsky-Rinzel neuron formalism~\autocite{pinskyIntrinsicNetworkRhythmogenesis1994a, atherton_bifurcation_2016} implemented in the NEURON simulator~\autocite{jaeger_neuron_2022} (Figure~\ref{fig:figure2-surrogate-opt}A; \suppnotesub{sec:supp-ca1-single}). Each cell is comprised of a somatic compartment (fast $\mathrm{Na^{+}}$ and delayed rectifier $\mathrm{K^{+}}$ currents) and a dendritic compartment (high-threshold $\mathrm{Ca^{2+}}$, $\mathrm{Ca^{2+}}$-activated $\mathrm{K^{+}}$, and afterhyperpolarization $\mathrm{K^{+}}$ currents), coupled by an adjustable soma-dendrite conductance. To ensure numerical stability under surrogate-guided optimization, we adopt the continuous reformulation of Atherton et al. \autocite{atherton_bifurcation_2016}, which replaces the discontinuous gating functions of the original Pinsky-Rinzel model with smooth exponential and sinusoidal approximations while preserving its qualitative dynamics (\suppnotesub{sec:supp-ca1-single}). Each model has 10-15 biophysical parameters governing compartment geometry, ion channel conductances, and passive membrane properties, and is optimized against four electrophysiological objectives (input resistance, membrane time constant, frequency-current relationship, and spike amplitude) subject to 7-8 binary constraints encoding biophysical validity requirements such as monotonic frequency-current responses and absence of spontaneous spiking.

We first verified that surrogate-assisted search is necessary for this problem class. Standard space-filling sampling strategies (symmetric Latin hypercube, SLH; Latin hypercube, LH; Monte Carlo, MC; Sobol sequences) fail to approach the theoretical maximum normalized hypervolume even after thousands of evaluations, whereas Gaussian process regression (GPR)-based surrogate optimization consistently reaches it (Figure~\ref{fig:figure2-surrogate-opt}B). This confirms that the objective landscape is too complex for passive sampling and motivates the use of surrogates to guide the search.

We then compared four neural network surrogate configurations. ResNet and FT-Transformer~\autocite{gorishniyRevisitingDeepLearning2021} backbones, each in objective-only (\texttt{o-}) and joint constraint-and-objective (\texttt{c+o-}) variants are benchmarked against the GPR and multitask exact Gaussian process (MEGP) baselines. All neural network surrogates achieved additive $\varepsilon$-indicator values well within the 2\% acceptable-solution threshold, indicating that their Pareto fronts nearly dominate those of the no-surrogate baseline (Figure~\ref{fig:figure2-surrogate-opt}C). This improvement is explained by systematically lower prediction error: the neural networks achieve lower normalized root-mean-square error (NRMSE) relative to the GP baselines throughout the optimization (Figure~\ref{fig:figure2-surrogate-opt}D). Importantly, this higher accuracy does not come at the cost of increased computation; the neural network surrogates incur comparable or lower inference times than the GP baselines, yielding a more favorable accuracy-to-compute trade-off (Figure~\ref{fig:figure2-surrogate-opt}E). Per-population breakdowns confirm that these trends hold across cell types with varying constraint difficulty (Supplementary Figures~\ref{fig:samplers},~\ref{fig:epsilon-performance}). Furthermore, per-constraint accuracy reveals substantial variation across the individual constraints, with firing behavior feasibility being predicted most reliably (\suppnotesub{sec:supp-constraint-accuracy}).

These results establish neural networks as a practical surrogate backbone for this problem class. Beyond their empirical advantages, neural networks are natively differentiable and straightforward to extend with additional output heads, properties that we exploit in the following sections to build the joint modeling framework.

\subsection{Joint surrogate learning yields more effective optimization than disjoint models}

Having established the neural network surrogate backbone, we next investigated whether joint learning of objectives and constraints affects optimization quality. We compared four neural network configurations (objective-only and joint variants of ResNet and FT-Transformer) and two GP baselines across all 10 single-cell optimization problems.

\begin{figure*}[h]
\centering
\includegraphics[width=0.95\textwidth]{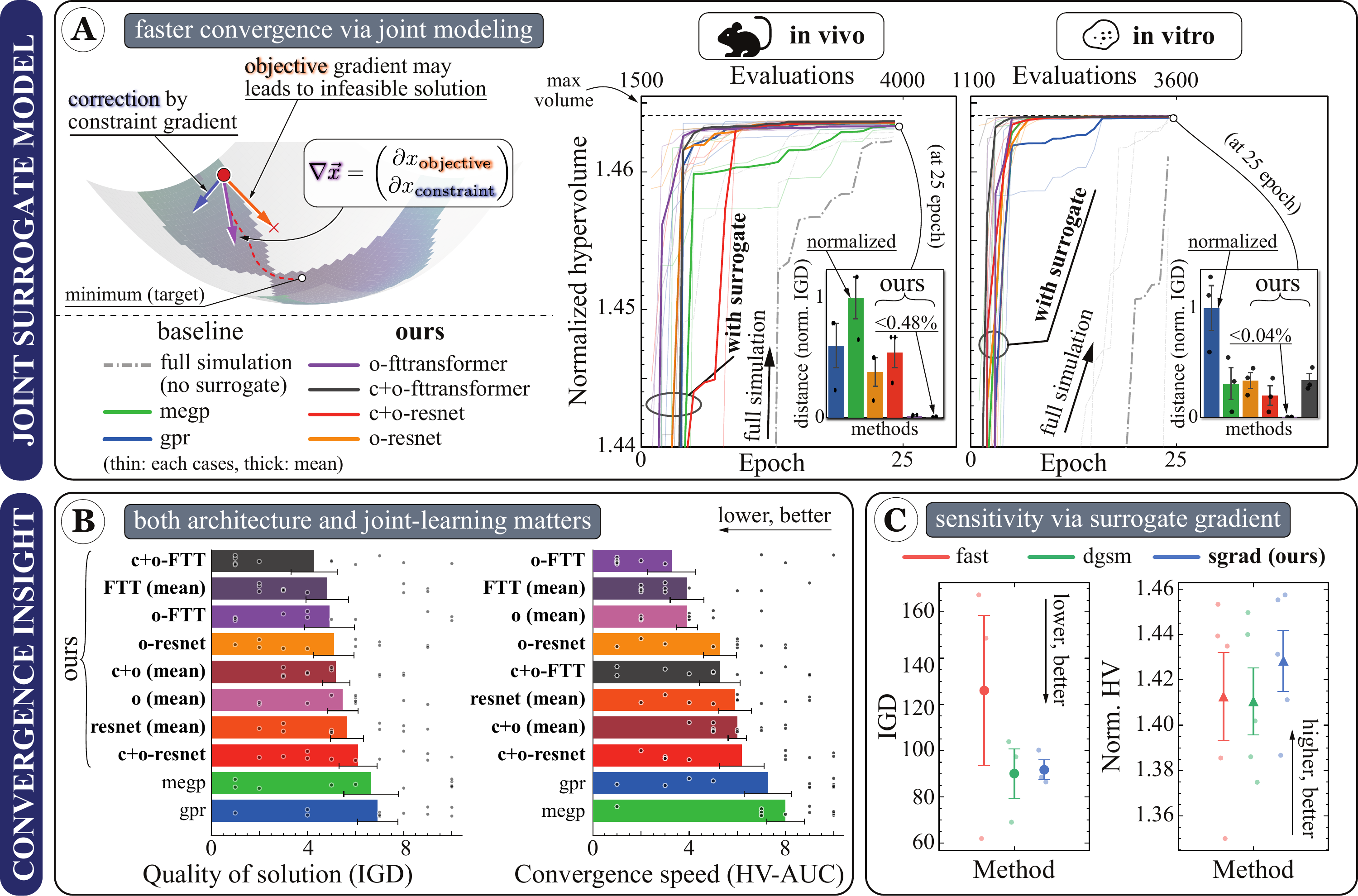}
\caption{\textbf{Joint differentiable representation of objectives, constraints and sensitivities.} \textbf{A}, Left: schematic illustration how the joint surrogate gradient - combining objective and constraint partial derivatives - corrects the search direction away from infeasible regions toward optimal solutions. Right: normalized hypervolume over optimization epochs for representative in vivo and in vitro cell types. All surrogate-assisted strategies (neural network and GP baselines) converge faster and with fewer evaluations than full simulation alone (dashed line). Insets show normalized Inverted Generational Distance (IGD) to full-simulation solution at epoch 25; surrogate-assisted methods achieve \(\leq 0.48\%\) (in vivo) and \(\leq 0.04\%\) (in vitro) deviation from the full-simulation Pareto front. \textbf{B}, Global ranking of surrogate strategies across all single-cell problems (lower is better). Left: joint constraint-and-objective learning (c+o) on an FTTransformer backbone achieves the highest solution quality (IGD). Right: objective-only architectures (o-) converge fastest (hypervolume area under the curve, HV-AUC), revealing a trade-off between convergence speed and final solution quality. The FTTransformer backbone consistently outranks the GP baselines on both metrics. \textbf{C}, Surrogate-gradient-based sensitivity analysis (sgrad) yields IGD and normalized hypervolume comparable to established methods (Fourier Amplitude Sensitivity Test, FAST; Derivative-based Global Sensitivity Measure, DGSM) while being a direct by-product of the surrogate model, requiring no additional simulation budget. Statistics: in \textbf{A}, thick lines show the mean normalized hypervolume over $n = 3$ independent trials (distinct random seeds) and thin lines the individual trials; inset bars show the mean IGD $\pm 1$ SEM with dots for the $n = 3$ trials. In \textbf{B}, bars show the mean rank $\pm 1$ SEM over $n = 10$ optimization problems (9 CA1 populations and the motoneuron) and dots the per-problem ranks. In \textbf{C}, markers show the mean $\pm 1$ SEM and dots the individual trials; $n = 3$ trials for IGD and $n = 5$ trials for normalized hypervolume.}\label{fig:figure3-joint-modelling}
\end{figure*}

Figure~\ref{fig:figure3-joint-modelling}A illustrates the core mechanism: the joint surrogate gradient, which combines partial derivatives with respect to both objectives and constraints, provides a corrected search direction that steers away from infeasible regions while simultaneously improving objective values. In practice, this translates to faster convergence. Normalized hypervolume curves for representative in vivo and in vitro cell types show that all surrogate-assisted strategies converge faster and with fewer evaluations than the full simulation alone, with the joint \texttt{c+o} variants reaching equivalent or superior hypervolume in fewer epochs. At epoch 25, surrogate-assisted methods achieve normalized inverted generational distance (IGD) values of $\leq 0.48\%$ (in vivo) and $\leq 0.04\%$ (in vitro) relative to the full-simulation Pareto front, confirming that surrogate-guided search closely approximates the true optimum. Per-population convergence curves for all 9 interneuron types are provided in Supplementary Figure~\ref{fig:all-convergence}; for the BS population, where the surrogate initially trails the full simulation, augmenting the search with the surrogate gradient restores faster convergence (Supplementary Figure~\ref{fig:bs-gradient}). During surrogate training, the objective and constraint gradients are positively aligned (mean cosine similarity $0.61$), pointing in compatible directions (\suppnotesub{sec:supp-gradient-geometry}).

To obtain a global ranking across cell types, we aggregated performance on two complementary metrics (Figure~\ref{fig:figure3-joint-modelling}B). For final solution quality (IGD), the joint \texttt{c+o}-FT-Transformer achieved the best mean rank across populations, demonstrating that shared learning of objectives and constraints produces better Pareto fronts. For convergence speed (hypervolume area under the curve, HV-AUC), the objective-only (\texttt{o-}) architectures converged fastest, revealing a trade-off between early convergence and final solution quality. This trade-off is expected: the joint model must also learn the constraint landscape, slightly slowing early convergence but ultimately producing solutions that are both more feasible and of higher quality. Across both metrics, the FT-Transformer backbone consistently outranked the GP baselines.

The jointly trained differentiable surrogate provides an additional benefit: its partial derivatives $\tfrac{\partial f}{\partial x_j}$ yield per-parameter sensitivity estimates at no extra computational cost. We compared these surrogate-gradient-derived sensitivity indices (\texttt{sgrad}) against two established methods, namely the Fourier Amplitude Sensitivity Test (FAST)~\autocite{mcraeGlobalSensitivityAnalysis1982} and the Derivative-based Global Sensitivity Measure (DGSM)~\autocite{sobolDerivativeBasedGlobal2010}, when used to inform the NSGA-II distribution indices that control crossover and mutation spread. Across representative cell populations, surrogate-gradient sensitivity yielded IGD and normalized hypervolume comparable to FAST and DGSM (Figure~\ref{fig:figure3-joint-modelling}C; Supplementary Figure~\ref{fig:sensitivity-all}), while being a direct by-product of the surrogate model and requiring no additional simulation budget. This result demonstrates that joint learning not only improves objective and constraint prediction but also furnishes actionable sensitivity information that further focuses the search on the most influential parameters.

\subsection{Surrogate constraint gradients guide optimization toward feasible regions in highly constrained problems}

The results established so far concern the surrogate's ability to reproduce the objective and constraint landscape. We now turn to the capability enabled by the surrogate's differentiability that allows computing a smooth descent direction toward the feasible set. This is especially useful when expert-curated parameter ranges are unavailable and modelers must begin from broad, naively specified bounds that span the full biophysical range of each parameter, relying on the optimizer to find the feasible subspace. To evaluate such a scenario, we optimized a spinal motoneuron model~\autocite{boothCompartmentalModelVertebrate1997} under both search-space configurations: the narrow, biologically informed range (presented as the representative in vitro cell type in Figure~\ref{fig:figure3-joint-modelling}; full comparison in Supplementary Figure~\ref{fig:baseline-sopt}) and a deliberately widened range with uninformed parameter bounds (\suppnotesub{sec:supp-motoneuron}).

\begin{figure*}[h]
\centering
\includegraphics[width=0.95\textwidth]{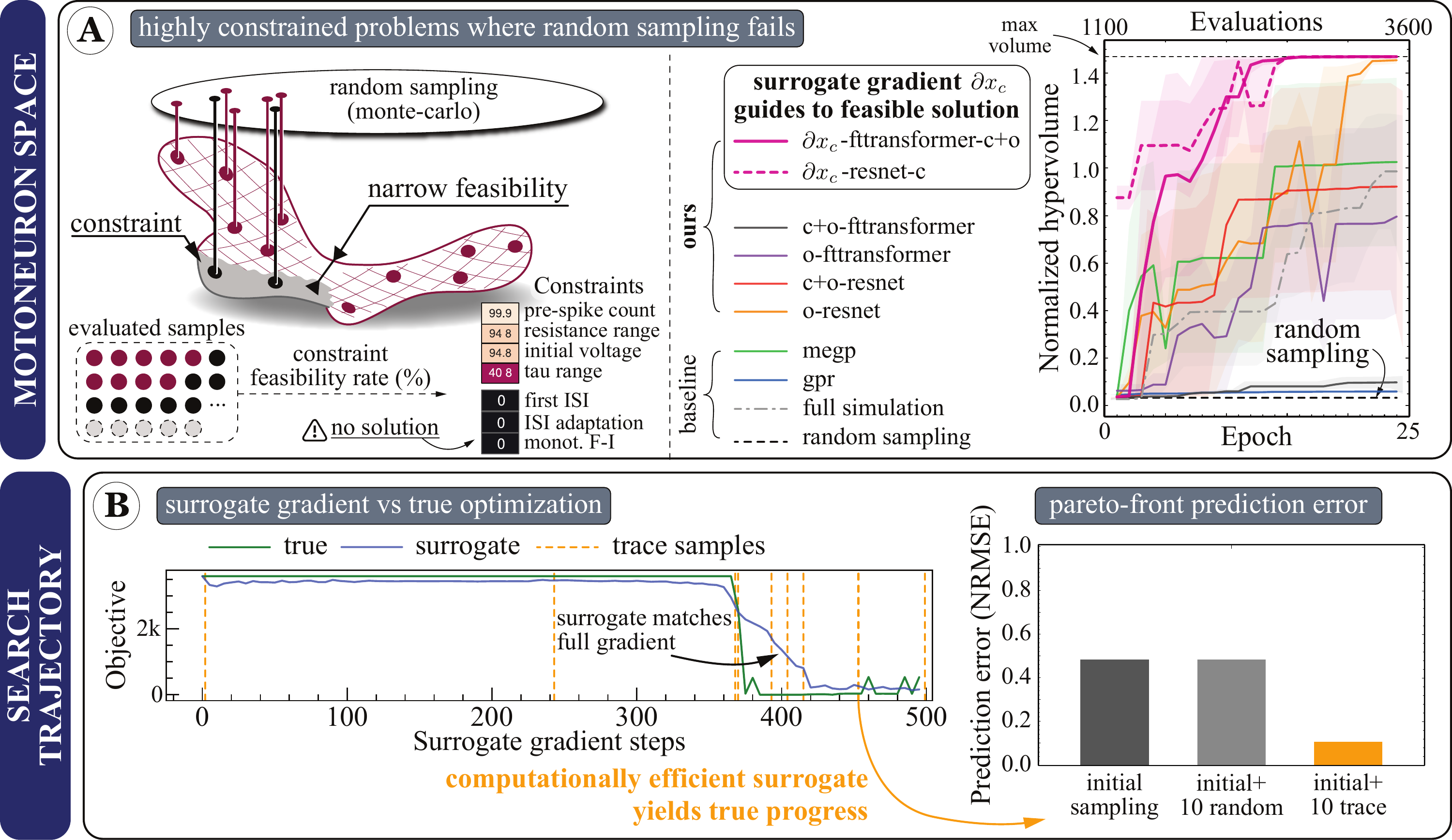}
\caption{\textbf{Solving highly constrained problems with constraint-surrogate gradients.} \textbf{A}, Left: characterization of constraint feasibility for a motoneuron optimization problem via random (Monte Carlo) sampling.  While four of seven constraints are satisfied at appreciable rates (pre-spike count, 99.9\%; resistance range, 94.8\%; initial voltage, 94.8\%; tau range, 40.8\%), three constraints - first inter-spike interval (ISI), ISI adaptation, and monotonic frequency-current (F-I) relationship - yield 0\% feasibility, preventing random sampling from discovering any valid solution. Right: computing the surrogate gradient with respect to the constraint classification head ($\nabla_{\mathbf{x}} f_{\mathbf{c}}$) guides the optimizer into feasible regions. Constraint-gradient-augmented strategies achieve rapid convergence in normalized hypervolume, outperforming both standard surrogate methods and the full-simulation baseline over 25 epochs. Random sampling (dashed black) fails to make meaningful progress. \textbf{B}, Left: analysis of the surrogate gradient-descent trajectory (blue) alongside its evaluation on the full simulation (green), confirming that the surrogate approximates the true objective landscape. Trace samples (orange dashed) are selected along the descent path by maximizing predicted diversity across all objectives. Right: augmenting the initial sampling set with 10 surrogate-descent trace samples substantially reduces Pareto-front prediction error (NRMSE) compared with the initial sampling alone or augmentation with 10 additional random samples, demonstrating the informativeness of gradient-guided trajectories for bootstrapping surrogate training. Statistics: in \textbf{A} (right), lines show the mean normalized hypervolume over $n = 3$ independent trials (distinct random seeds) and shaded bands $\pm 1$ SEM. \textbf{B} (left) shows a single gradient-descent trajectory; \textbf{B} (right) reports the NRMSE of one surrogate fit per condition evaluated on 511 Pareto-front points, so no error bars are shown.}\label{fig:figure4-highly-constraint}
\end{figure*}

Characterizing constraint feasibility via Monte Carlo sampling reveals the severity of the wide-range problem (Figure~\ref{fig:figure4-highly-constraint}A, left). While four of the seven constraints are satisfied at appreciable rates, three constraints yield exactly 0\% feasibility. Because joint satisfaction of all constraints is required, random sampling fails to discover even a single valid solution, rendering standard optimization ineffective. This combinatorial infeasibility is not an artifact of undersampling, as the three zero-feasibility constraints impose narrow manifolds in parameter space that cannot be located without directed search (Supplementary Figure~\ref{fig:random-constraints}).

Computing the surrogate gradient with respect to the constraint classification head ($\nabla_{\mathbf{x}} f_{\mathbf{c}}$) provides a directed signal to overcome this barrier. By appending gradient-descent steps on the frozen surrogate to the evolutionary search, the optimizer is actively steered into feasible parameter regions. Constraint-gradient-augmented strategies achieve rapid convergence in normalized hypervolume, outperforming both standard surrogate methods and the full-simulation baseline over 25 optimization epochs (Figure~\ref{fig:figure4-highly-constraint}A, right). In contrast, random sampling (dashed black line) fails to make meaningful progress, confirming that the smooth gradient signal from the surrogate is what allows the search to navigate the discontinuous feasibility landscape.

It is worth noting that the three constraints not satisfied by any training sample present the surrogate with a degenerate target of all-zero labels. Since the binary cross-entropy loss drives the corresponding logits uniformly downward, its gradient with respect to the parameters nearly vanishes and offers no direction toward its own feasible region. The loss trajectories do not let us observe the mechanism directly, but the most plausible explanation for successful descent in spite of this is signal from the other heads, aided by the diversity and mutation of the evolutionary population. This moves candidates through parameter space until one of them satisfies a previously unsatisfied constraint and starts contributing an informative gradient of its own.

To verify that the surrogate gradient provides a reliable approximation of the true objective landscape, we analyzed the gradient-descent trajectory in detail (Figure~\ref{fig:figure4-highly-constraint}B, left). Surrogate-predicted objective values (blue) were evaluated in parallel on the full NEURON simulation (green), confirming close agreement along the descent path. Occasional divergence, particularly near parameter configurations that produce non-viable simulations, highlights the importance of periodically re-anchoring the surrogate to true evaluations. The full multi-objective gradient-descent traces for all four objectives are provided in Supplementary Figure~\ref{fig:sopt-grad-sampling}.

We further investigated whether the gradient-descent trajectory could serve as an informative source of training data for bootstrapping the surrogate. Along the descent path, trace samples were selected by maximizing predicted diversity across all objectives via iterative distance filtering (\suppnote{sec:supp-gradient}). Augmenting the initial training set with just 10 trace samples substantially reduced Pareto-front prediction error (NRMSE) compared with either the initial sampling alone or augmentation with 10 additional random samples (Figure~\ref{fig:figure4-highly-constraint}B, right). This demonstrates that gradient-guided trajectories traverse informative regions of the search space and can meaningfully accelerate surrogate training at minimal evaluation cost.
Notably, a common alternative for this kind of problem is constrained BO~\autocite{gardnerBayesianOptimizationInequality2014,gelbartBayesianOptimizationUnknown2014}, which  explores through posterior uncertainty and proposes candidates by traversing an acquisition function one point at a time. The wide-range motoneuron problem with no feasible sample in the initial draw presents a challenging test for it. With the expected improvement undefined, the search must begin by maximizing the predicted probability of feasibility alone~\autocite{gelbartBayesianOptimizationUnknown2014}. We ran a Gardner-style constrained BO baseline (independent GPs per objective, GP classifiers per constraint, feasibility-weighted expected hypervolume improvement; Methods, \emph{Constrained Bayesian optimization baseline}) using three batch sizes that span the trade-off between sample efficiency and parallelism: $q=100$, matching the evaluation parallelism of \textsc{dmosopt}; $q=16$, a compute-efficient batch; and $q=1$, the sample-efficient sequential setting, run with a reduced evaluation budget because it serializes every simulation (\suppnotesub{sec:supp-cbo}). For the three batch sizes, constrained BO found no jointly feasible sample within 2{,}400 evaluations in any of three trials, leaving the normalized hypervolume at the value after initial sampling ($0.032 \pm 0.007$). The constraint-gradient-augmented \texttt{c+o}-FT-Transformer reached $1.4640 \pm 0.0002$ over the same evaluation budget. Three of the seven constraints were satisfied by no sample at all, so the corresponding per-constraint classifiers never saw a positive example and the search did not leave the cold-start phase. On the narrow search space, where each constraint is individually satisfiable, the same baseline does reach the optimum, but requires more evaluations than any black-box method (\suppnotesub{sec:supp-cbo}).

\subsection{Joint surrogate optimization enables multi-fidelity large-scale network simulation with minimal evaluations}

\begin{figure*}[h]
\centering
\includegraphics[width=\textwidth]{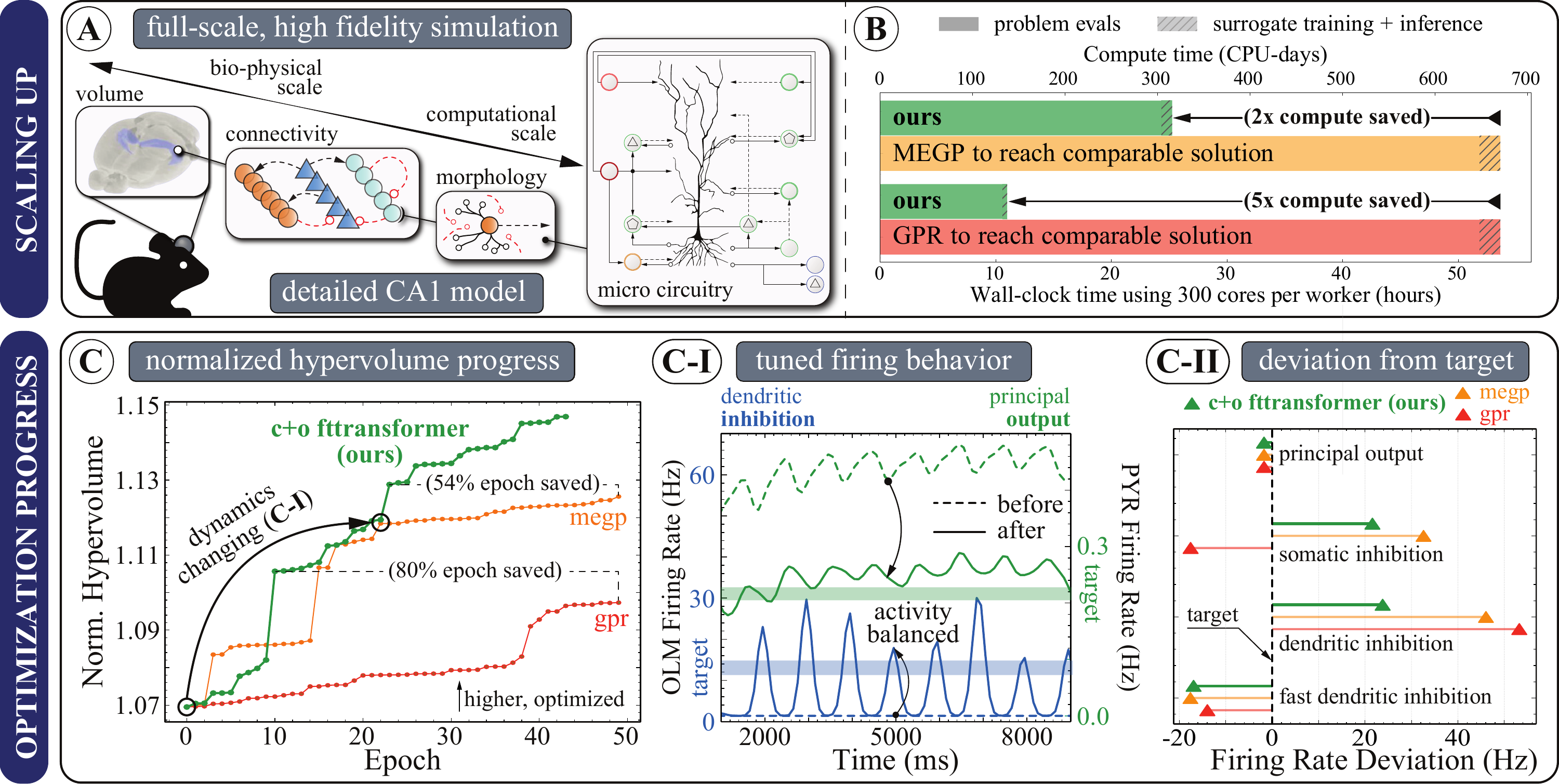}
\caption{\textbf{Joint-surrogate optimization of large-scale network problems.} \textbf{A}, The multi-scale CA1 hippocampal network optimization problem spans biophysical scales (brain volume $\rightarrow$ connectivity  $\rightarrow$ cell morphology) to computational scales (detailed micro-circuitry simulations), requiring high-fidelity, full-scale simulation. \textbf{B}, Total compute cost, measured in CPU-days (top axis) and wall-clock hours on 300 cores per worker (bottom axis), broken down into problem evaluations (solid bars) and surrogate training plus inference (hatched bars). The joint c+o-FTTransformer reaches solutions comparable to the MEGP baseline with 2x less compute and to the GPR baseline with 5x less compute. \textbf{C}, Normalized hypervolume over 50 optimization epochs. The c+o-FTTransformer (ours) and MEGP show similar initial convergence rates, both dominating the GPR baseline; after approximately 20 epochs the joint transformer advances to higher hypervolume, saving 54\% of epochs relative to MEGP and 80\% relative to GPR to reach equivalent solution quality. \textbf{C-I} Instantaneous population firing rates before (dashed) and after (solid) optimization for oriens-lacunosum moleculare (OLM) interneurons (dendritic inhibition, left axis) and pyramidal (PYR) cells (principal output, right axis). Optimization brings both populations into their respective target ranges (shaded bands), yielding balanced network activity. \textbf{C-II} Firing-rate deviation from target (0 = ideal) across four functional cell classes (principal output, somatic inhibition, dendritic inhibition, fast dendritic inhibition). The joint c+o-FTTransformer consistently achieves smaller deviations than the MEGP and GPR baselines, producing network dynamics closer to the biological target. All panels derive from a single optimization run per surrogate method ($n = 1$) because of the computational cost. To gauge the run-to-run variability of the trajectory in \textbf{C}, we run $n = 3$ independent replicates (distinct random seeds) of the c+o-FTTransformer configuration for the first 5 epochs under a revised objective formulation whose absolute hypervolume values are not directly comparable to \textbf{C}, yielding two dimensionless spread proxies: (i) The SEM of the fraction of attainable hypervolume gained since epoch 0 is $0.011$ at epoch 5, corresponding to $\pm 0.022$ normalized hypervolume in \textbf{C}. (ii) The SEM of the hypervolume gain relative to its mean is $27\%$ at epoch 5, corresponding to $\pm 0.001$ at epoch 5 and $\pm 0.021$ at the final epoch of \textbf{C} if the relative variability is held constant. Because the three methods in \textbf{C} start from the same initial sample, both proxies exclude initial-sampling variance and quantify the stochasticity of the surrogate optimization itself.}\label{fig:figure5-large-scale}
\end{figure*}

Following the single-cell benchmarks, we applied our framework to a full-scale biophysical simulation of the hippocampal CA1 circuit~\autocite{bezaireInterneuronalMechanismsHippocampal2016,upadhyayMiVSimulatorComputationalFramework2023}, a problem that is out of reach for non-surrogate optimization. 
The full-scale network model comprises $836{,}970$ neurons across 15 cell populations connected by almost 10 billion synaptic connections
(\suppnotesub{sec:supp-network}), spanning from brain-volume connectivity down to single-cell ion-channel dynamics (Figure~\ref{fig:figure5-large-scale}A).

Because multiple evaluations of the complete network within an iterative surrogate optimization loop are computationally intractable, we employed a multi-fidelity strategy in which optimization is performed on a network slice of approximately 8,700 neurons extracted from the full-scale network (\suppnotesub{sec:supp-network}). This computational strategy mirrors the use of acute hippocampal slice preparations in experimental neuroscience, where a spatially restricted section of tissue preserves local circuit connectivity and cellular physiology sufficiently to characterize network dynamics, with results that inform understanding of the intact circuit. The slice preserves the full complement of cell populations, projection pathway types, and single-cell biophysical detail of the full model, allowing surrogate-guided search at manageable cost. Synaptic weight parameters optimized at slice scale are subsequently transferred to the full-scale model for validation.

 Each simulation evaluation of the slice requires ${\sim}42.3$\,s on 300 CPU cores (${\sim}5.4$ TACC Frontera supercomputer nodes), and the optimization targets (per-population firing rates and fractions of active cells) make for a high-dimensional objective space that is challenging to scale with GP-based surrogates. Because of the computational cost (${\sim}60$--$700$ CPU-days per run), the network optimization was run only once per surrogate method.
The total computational cost, measured in both CPU-days and wall-clock hours, shows efficiency gains for the joint surrogate approach (Figure~\ref{fig:figure5-large-scale}B). The \texttt{c+o}-FT-Transformer reached solutions of comparable quality to the MEGP baseline with $2\times$ less total compute and to the GPR baseline with $5\times$ less compute, with surrogate training and inference overhead being negligible relative to simulation time.

Normalized hypervolume convergence over 50 optimization epochs shows that the \texttt{c+o}-FT-Transformer and MEGP exhibit similar initial convergence rates, both dominating the GPR baseline (Figure~\ref{fig:figure5-large-scale}C). After approximately 20 epochs, however, the joint transformer advances to higher hypervolume, saving 54\% of epochs relative to MEGP and 80\% relative to GPR to reach equivalent solution quality.  It should be noted that these figures derive from a single optimization per surrogate method, and should thus be read as a demonstration of scale rather than a replicated benchmark. The late-stage acceleration reflects the increasing accuracy of the neural network surrogate as it accumulates more training data, a regime where GP-based methods, whose computational cost scales super-linearly with sample size, become increasingly uncompetitive. 

The optimized network produces biologically realistic dynamics: instantaneous population firing rates for oriens-lacunosum moleculare (OLM) interneurons (dendritic inhibition) and pyramidal (PYR) cells (principal output) are brought into their respective experimentally determined target ranges, yielding balanced excitatory-inhibitory network activity (Figure~\ref{fig:figure5-large-scale}C-I). Quantifying firing-rate deviation from target across four functional cell classes confirms that the \texttt{c+o}-FT-Transformer consistently achieves smaller deviations than both GP baselines, producing network dynamics that more closely match the biological target across all cell types (Figure~\ref{fig:figure5-large-scale}C-II).

These results demonstrate that the combined framework of joint surrogate learning, constraint-gradient-guided search, and sensitivity-informed sampling translates the efficiency gains observed on single-cell benchmarks to large-scale network models, reducing the computational burden of large-scale neural circuit optimization from hundreds to tens of CPU-days.

\section{Discussion}\label{sec:discussion}

The central finding of this work is that deterministic neural network surrogates match Gaussian process accuracy from as few as 1,000 training samples (100 Latin-hypercube samples per parameter, less one cross-validation fold; \suppnote{sec:supp-loop}) and, when objectives and constraints are learned jointly in a shared differentiable model, enable gradient-based feasibility solving that disjoint models do not provide. That deterministic networks are competitive in this data-scarce regime is not obvious since they forgo strong probabilistic priors~\autocite{williams_gaussian_2006} and inductive bias that learning theory identifies as the basis for generalization from few examples~\autocite{wilson_bayesian_2020}. Instead, the feature representation is learned generically from data, trading sample efficiency for flexibility. Our results indicate that, for these problems, learned features can substitute for a hand-specified prior. The joint advantage, in turn, is structural: in biophysical neuron models, the same ion channel conductances that govern firing-rate objectives also determine whether physiological constraints such as monotonic frequency-current relationships are satisfied. A joint model captures these shared dependencies in its latent representation, producing more accurate predictions precisely in the constraint-relevant regions of the search space.

A second, distinct advantage follows from differentiability itself that allows gradient-based descent towards the feasible set. For the wide-range motoneuron benchmark, whose initial sampling yielded no feasible point, constraint-gradient descent recovered feasible, near-optimal solutions, whereas constrained Bayesian optimization found none within the same evaluation budget.

At scale, the neural network parameterization becomes necessary. The $\mathcal{O}(n^3)$ cost of Gaussian process surrogates restricts their use to modest evaluation budgets~\autocite{liuWhenGaussianProcess2020}, whereas the network optimization problem, with its high-dimensional objective space and thousands of accumulated training samples, requires a surrogate that scales gracefully with both data and output dimensionality. The stationarity of standard kernels compounds this limitation. Since objective errors on infeasible solutions are effectively unbounded, a stationary GP fitted to all samples predicts the feasible candidates less accurately while its fitting cost grows by an order of magnitude, so it is fitted to the feasible subset without further target engineering (Methods, \emph{Gaussian process baselines}; \suppnotesub{sec:supp-gp-infeasible}).
Moreover, alternative probabilistic approaches like constrained BO~\autocite{gardnerBayesianOptimizationInequality2014,gelbartBayesianOptimizationUnknown2014,hernandez-lobatoPredictiveEntropySearch2015} are limited by the sequential nature of acquisition optimization.
The accompanying trade-off, the loss of calibrated uncertainty estimates, is partially compensated by the intrinsic exploration of the multi-objective evolutionary algorithm through population diversity, mutation, and recombination (in this work concretely implemented by NSGA-II-type optimizers~\autocite{debFastElitistMultiobjective2002a}). Although our benchmarks are drawn from computational neuroscience, the framework requires only a black-box objective function with optional binary constraints. The same approach could serve constrained optimization problems in engineering design, where aerodynamic or structural objectives must satisfy manufacturing and safety constraints over high-dimensional shape parameterizations~\autocite{forresterEngineeringDesignSurrogate2008,kozielSurrogateBasedModelingOptimization2013}, molecular and drug design, where candidate molecules must satisfy pharmacokinetic and synthesizability constraints~\autocite{gomez-bombarelliAutomaticChemicalDesign2018,griffithsConstrainedBayesianOptimization2020}, or materials discovery, where compositions and processing conditions span high-dimensional spaces subject to stability and performance bounds~\autocite{lookmanActiveLearningMaterials2019}.

Several limitations and corresponding future directions merit discussion. The surrogate may overfit to early training data, directing the search toward spurious optima (\suppnotesub{sec:bs-fail}); we mitigate this via cross-validated epoch selection, periodic retraining, and elite preservation in the evolutionary population (Supplementary Algorithm~\ref{alg:hybrid}), but more principled trust-region strategies could further improve robustness. The smooth approximation of the constraint landscape, while essential for gradient computation, may oversmooth sharp feasibility boundaries; adaptive smoothing or constraint-specific heads could help. Relatedly, the constraint gradient encodes direction but not distance. The constraint head is a smooth stand-in for the all-or-nothing membership of the feasible set with $0.5$ marking the feasibility boundary. However, unlike the signed-distance surfaces of level-set methods, where the gradient is held at unit magnitude, the steepness and scaling of our fitted surface carry no such meaning and are artifacts of the fit. The gradient can therefore be trusted for which way to move, not how far, which is why it is used as a search direction, norm-balanced against the objective, rather than as a calibrated step. It can also point the wrong way where training data are sparse, since nothing there forces feasibility onto the correct side of the surface; periodic re-anchoring to true simulations, elite preservation, and population diversity are what keep such occasional misdirection from derailing the search. Hypervolume computation scales exponentially with the number of objectives, limiting applicability for many simultaneous objectives. Looking ahead, integrating uncertainty quantification through evidential deep learning~\autocite{sensoyEvidentialDeepLearning2018} or deep ensembles~\autocite{ganaieEnsembleDeepLearning2022} could combine the scalability of neural networks with the principled exploration of Bayesian optimization. Pre-training surrogates on libraries of related problems, warm-starting from partially differentiable simulators, and extending to a high number of objectives settings are natural next steps.

In summary, by jointly capturing objectives, constraints, and sensitivities in a single differentiable model, the framework lets practitioners specify what they know in terms of constraints on acceptable behavior rather than exact target values for optimization to handle the rest, shifting the burden from manual parameter tuning to principled, scalable, automated search.

\section{Methods}\label{sec:methods}

\subsection{Problem formulation}

We consider a black-box constrained multi-objective optimization problem of the form
\begin{align}
    \min_{\mathbf{x} \in \mathcal{X}} \; &\mathbf{y}(\mathbf{x}) = \bigl(y_1(\mathbf{x}), \ldots, y_q(\mathbf{x})\bigr), \notag \\
    \text{s.t.} \; &c_j(\mathbf{x}) \in \{0,1\}, \quad j = 1,\ldots,k, \notag \\
    &\mathbf{x}_{lb} \leq \mathbf{x} \leq \mathbf{x}_{ub},
\end{align}
where $\mathbf{x} \in \mathds{R}^n$ is the parameter vector, $\mathcal{X} = [\mathbf{x}_{lb}, \mathbf{x}_{ub}]$ is the bounded search space, $\mathbf{y}\colon \mathds{R}^n \to \mathds{R}^q$ are $q$ objective functions to be minimized, and $c_1, \ldots, c_k$ are $k$ binary constraint functions, where $c_j(\mathbf{x}) = 1$ indicates feasibility and $c_j(\mathbf{x}) = 0$ indicates infeasibility. Each evaluation of $\mathbf{y}$ and $\mathbf{c}$ requires an expensive simulation (e.g., a biophysical neuron model solved by a numerical integrator), and no analytical gradients are available. A parameter vector is considered feasible if and only if all constraints are simultaneously satisfied: $\prod_{j=1}^k c_j(\mathbf{x}) = 1$. Using as few simulation evaluations as possible, the goal is to approximate the constrained Pareto front, i.e. the set of feasible solutions for which no other feasible solution improves one objective without worsening another.

\subsection{Joint surrogate model}

The core of the framework is a joint surrogate model $f\colon \mathds{R}^n \to \mathds{R}^{q+k}$ that simultaneously predicts objective values $\hat{\mathbf{y}} = f_{\mathbf{y}}(\mathbf{x}) \in \mathds{R}^q$ and constraint satisfaction probabilities $\hat{\mathbf{c}} = f_{\mathbf{c}}(\mathbf{x}) \in [0,1]^k$ from a shared learned representation. We parameterize $f$ as a deep neural network with a shared backbone and two task-specific output heads: a regression head for objectives (linear activation) and a classification head for constraints (sigmoid activation).
We evaluated two backbone architectures from the tabular deep learning literature~\autocite{gorishniyRevisitingDeepLearning2021}: a ResNet with batch-normalized residual blocks, and a FT-Transformer that processes learned token embeddings via multi-head self-attention.
The model is trained with Adam ($\eta = 0.001$, batch size 2048) on an unweighted sum of mean squared error (objectives) and binary cross-entropy (constraints). Weighting the two heads does not improve the surrogate's ability to identify feasible parameter sets or predict objectives more accurately (\suppnotesub{sec:supp-loss-weighting}). The number of training epochs is selected via 3-fold cross-validation; note that early stopping monitors validation loss but does not restore best-epoch weights, relying instead on epoch averaging to avoid overfitting to a single fold. Full architecture specifications, hyperparameters, normalization procedures, and adaptive scaling rules are provided in \suppnote{sec:supp-architectures}.

\subsection{Gradient-based feasibility solving}

The differentiability of the joint surrogate enables gradient-based steering of candidate solutions toward feasible, high-performing regions of the search space. Given a batch of candidate parameters $\mathbf{X}$ and the frozen trained surrogate, we perform iterative gradient descent on $\mathbf{X}$ with respect to a composite loss that combines objective minimization (hypervolume maximization) and constraint feasibility (binary focal cross-entropy) terms, optionally augmented by an exploration loss and a non-negativity penalty. When multiple terms are active, their gradients are $L^2$-norm-balanced to the first term. 
This procedure augments, rather than replaces, the evolutionary optimizer: in each epoch, the NSGA-II~\autocite{debFastElitistMultiobjective2002a} population is partitioned so that the elite half is preserved while the remaining candidates are refined via gradient descent before evaluation on the true simulator (Supplementary Algorithm~\ref{alg:hybrid}). Full loss definitions, optimizer settings, and convergence criteria are provided in \suppnote{sec:supp-gradient}.

\subsection{Sensitivity-informed sampling}

The partial derivatives of the trained surrogate provide a computationally free estimate of parameter sensitivity. Per-parameter elasticities $S_j^{(k)} = \tfrac{\partial \hat{y}_k}{\partial x_j} \cdot x_j$, averaged in absolute value across training samples, are mapped to NSGA-II distribution indices so that sensitive parameters undergo smaller perturbations while insensitive ones are explored more broadly. This requires only forward and backward passes over the training data with no additional simulation evaluations, in contrast to FAST or DGSM which require dedicated sampling designs (\suppnote{sec:supp-sensitivity}).

\subsection{Optimization loop}

The complete optimization follows an iterative surrogate-assisted loop (Supplementary Algorithm~\ref{alg:loop}; \suppnote{sec:supp-loop}). An initial set of parameter vectors is drawn via symmetric Latin hypercube sampling with 100 samples per parameter for the single-cell problems (${\sim}1{,}000$--$1{,}500$ across their 10--15 parameters), and 2 samples per parameter for the network ($284$ total), evaluated on the true simulator. In each subsequent epoch: (1)~the surrogate is retrained on \emph{all} accumulated data, (2)~sensitivity indices are optionally computed, (3)~NSGA-II generates candidate solutions using the surrogate as a fast proxy for the true objective function; all generations within an epoch operate entirely on surrogate predictions, with only the final candidate set evaluated on the true simulator, (4)~candidates are optionally refined via gradient-based feasibility solving, and (5)~results are appended to the training set. The loop terminates after a fixed epoch budget (25 for single-cell, 50 for network) or upon a convergence criterion. Because the surrogate's training epochs are selected by 3-fold cross-validation, each fit uses two-thirds of the accumulated data. The cross-validation also enables dynamic initial sampling  that determines a minimum workable count without manual tuning (\suppnote{sec:supp-loop}).

\subsection{Gaussian process baselines}\label{sec:methods-gp}

Two Gaussian process (GP) surrogates serve as baselines. Gaussian Process Regression (GPR) fits an independent GP per objective implemented in scikit-learn~\autocite{pedregosaScikitlearnMachineLearning2011}, each using a Mat\'ern~5/2 kernel so predictions are independent per objective. The multitask exact Gaussian process (MEGP) on the other hand fits a single GP jointly over all objectives in GPyTorch~\autocite{gardnerGpytorchBlackboxMatrixmatrix2018}, combining a Mat\'ern kernel over the parameters with a rank-1 coregionalization kernel over the objectives. MEGP is thus the joint multi-output baseline, modelling cross-objective correlations that GPR cannot represent. Unlike the neural network surrogates, the GP surrogates are trained exclusively on the samples that satisfy all constraints. This is due to the effectively unbounded value range of the objective error on infeasible solutions. For example, a model that fails to spike may produce objective errors that exceed those of any feasible solution by orders of magnitude. Since the Mat\'ern kernel global length-scale and variance is stationary in the input space~\autocite{gibbs_bayesian_1998,williams_gaussian_2006}, considerable representational capacity is drawn into predicting infeasible samples with very large objective error. This is independent of normalization or standard transformations but a consequence of the non-uniformity of the local variation of the objective landscape as a function of the inputs. While non-stationary remedies exist, including input-dependent length-scales~\autocite{gibbs_bayesian_1998,paciorek_nonstationary_2003}, warped inputs or outputs~\autocite{snelson_warped_2003}, and deep kernel learning~\autocite{wilson_deep_2016}, they still suffer from the $\mathcal{O}(n^3)$ scaling in the number of samples and require problem-specific feature engineering or additional model selection. Therefore, we adopt the simpler standard remedy of using the feasible subset~\autocite{gardnerBayesianOptimizationInequality2014}. To quantify the effect on our problems, we additionally ran the MEGP surrogate on all samples, including the infeasible ones (\suppnotesub{sec:supp-gp-infeasible}).

\subsection{Constrained Bayesian optimization baseline}\label{sec:methods-cbo}

To contrast the framework with uncertainty-driven search, we implemented a Gardner-style constrained multi-objective BO baseline in BoTorch~\autocite{balandatBoTorchFrameworkEfficient2020}. Objectives are modeled by independent single-task GPs (Mat\'ern~5/2 kernel, standardized outputs) fitted to the feasible subset of the data~\autocite{gardnerBayesianOptimizationInequality2014}. Each binary constraint is modeled by an independent sparse variational GP classifier (128 inducing points, Bernoulli likelihood) fitted to all data~\autocite{gelbartBayesianOptimizationUnknown2014} and refitted every 10 iterations. Candidates maximize the noisy expected hypervolume improvement (qLogNEHVI~\autocite{daultonParallelBayesianOptimization2021,amentUnexpectedImprovementsExpected2023}), weighted by the product of the per-constraint feasibility probabilities~\autocite{gelbartBayesianOptimizationUnknown2014}. The batches of size $q>1$ are assembled by sequential greedy acquisition optimization (4 restarts, 128 raw samples per point). Total count matches the NSGA-II design, with simulation batch sizes of $q=1, 16, 100$.

\subsection{Software framework}

The framework is implemented in \textsc{dmosopt}, an open-source Python package following a controller-worker architecture communicating via MPI. The controller manages the optimization state and surrogate; workers execute simulations and return results. The framework supports interchangeable optimizers (NSGA-II, AGE-MOEA~\autocite{panichella_adaptive_2019}, SMPSO~\autocite{nebroSMPSONewPSObased2009}, CMA-ES~\autocite{hansenCMAEvolutionStrategy2006}), surrogates (GPR, MEGP, ResNet, FT-Transformer, and custom models via a plugin interface), samplers, and sensitivity methods. A full specification is provided in \suppnote{sec:supp-framework}. All experiments used Python 3.12 with \textsc{dmosopt} 0.78 (commit \texttt{f034729}), NEURON 8.2.6, MiV-Simulator 0.2.1 (commit \texttt{e440eaa}), TensorFlow 2.16.2 for the neural network surrogates, PyTorch 2.5.1 and GPyTorch 1.14.2 for the MEGP baseline, scikit-learn 1.7.2 for the GPR baseline and metrics, BoTorch 0.18.1 for the constrained BO baseline, SALib 1.5.1 for FAST and DGSM, NumPy 1.26.4, SciPy 1.16.2, mpi4py 4.1.1, h5py 3.15.1, and \textsc{machinable} 4.10.6 for experiment tracking (Supplementary Table~\ref{tab:software}).

\subsection{Benchmark problems}

We evaluated the framework on three benchmark problems of increasing complexity, each representing a stage in a neural circuit modeling workflow.

\paragraph{CA1 hippocampal interneuron single-cell models.}
Nine morphologically distinct CA1 interneuron types (SCA, IVY, PVBC, CCKBC, AAC, BS, OLM, NGFC, IS) were optimized using two-compartment Pinsky--Rinzel models~\autocite{pinskyIntrinsicNetworkRhythmogenesis1994a} in the NEURON simulator~\autocite{jaeger_neuron_2022}. Each model has 10-15 parameters (compartment geometry, coupling and ion channel conductances, calcium dynamics), 4 objectives (input resistance, membrane time constant, f--I curve, and spike amplitude errors computed as squared distance to experimentally determined target ranges), and 7-8 binary constraints (monotonic f-I, resistance and time constant ranges, spike detection, ISI properties, absence of spontaneous spiking, and initial voltage accuracy). Optimization ran for 25 epochs with 100 evaluations per epoch (population size 100, 10 NSGA-II generations per epoch, 100 initial samples per parameter via symmetric Latin hypercube sampling, i.e.\ 1,000--1,500 across the 10--15 parameters). Full parameter specifications and electrophysiological targets are provided in \suppnotesub{sec:supp-ca1-single}.

\paragraph{Motoneuron model.}
A two-compartment motoneuron model with 11 parameters (coupling, somatic Na$^+$/K$^+$/KCa/CaN, dendritic CaL/CaN/KCa, leak conductances, capacitance ratio) ~\autocite{boothCompartmentalModelVertebrate1997} was optimized against the same 4 types of objectives and 7 types of constraints as the CA1 interneuron models. Objective targets were based on experimental characterizations of spinal motoneuron cultures \autocite{milesFunctionalPropertiesMotoneurons2004}.  Two search-space configurations were used: a narrow, biologically informed range (baseline) and a deliberately widened range in which 3 of 7 constraints yield 0\% feasibility under random sampling, serving as a test of gradient-based feasibility solving. Optimization ran for 25 epochs (\suppnotesub{sec:supp-motoneuron}).

\paragraph{CA1 hippocampal network model.}
A full-scale biophysical simulation of the CA1 microcircuit~\autocite{bezaireInterneuronalMechanismsHippocampal2016} implemented in the MiV-Simulator~\autocite{upadhyayMiVSimulatorComputationalFramework2023} with $836{,}970$ neurons across 15 populations and almost 10 billion synaptic connections. 
Large numbers of evaluations of the full-scale model within the surrogate optimization loop are infeasible on the HPC hardware allocation available to us. Optimization was therefore performed on a subset (``slice'') of the network comprising approximately 8,700 neurons distributed across all 15 cell populations, with per-population counts listed in Supplementary Table~\ref{tab:ca1-slice-counts} and approximately 265 million connections.
The optimization adjusts $n = 142$ synaptic weight parameters across all projection pathways and comprises 36 objectives (per-population firing rate, mean fraction of active cells, and standard deviation of the fraction of active cells across 12 target populations) and 12 constraints (positive firing rates). Each evaluation required ${\sim}42.3$\,s on 300 CPU cores on the Frontera supercomputer (Texas Advanced Computing Center). Optimization ran for 50 epochs with population size 100 and 10 NSGA-II generations per epoch (\suppnotesub{sec:supp-network}).

\paragraph{Metrics.}
Solution quality was assessed via normalized hypervolume (HV), hypervolume area under the curve (HV-AUC), inverted generational distance (IGD), and the additive $\varepsilon$-indicator; surrogate accuracy via normalized root-mean-square error (NRMSE). All metrics are defined in \suppnote{sec:supp-metrics}. Single-cell and motoneuron experiments were replicated 3-5 times with independent random seeds; the network problem was run once per method due to computational cost (\suppnote{sec:supp-statistics}).

\paragraph{Computational resources.}
All experiments were executed on the Frontera supercomputer at the Texas Advanced Computing Center (TACC). Single-cell and motoneuron optimizations each required $<$1 CPU-day. The network optimization required ${\sim}60$-700 CPU-days per surrogate method ($5{,}000$ evaluations $\times$ $\sim$42\,s $\times$ 300 cores). The total computational budget across all experiments, including all replicates, surrogates, and baselines, was approximately 1{,}500 CPU-days.

\subsection*{Data availability}\label{sec:data-availability}

The study generated simulation data only. The electrophysiological target values that define the objectives and constraints are taken from previously published experimental studies, cited in the respective benchmark descriptions (\suppnote{sec:supp-benchmarks}). The processed data underlying every figure (per-trial normalized hypervolume, IGD and $\varepsilon$-indicator values, surrogate prediction and timing metrics, constraint feasibility rates, and network firing-rate summaries) are provided as JSON files in the \texttt{figures/data} directory of the analysis repository~\autocite{gressmann2026msoCode} and are archived with it on Zenodo. Source data for the main figures are provided with this paper. The raw optimization records (one HDF5 file per run containing all evaluated parameter vectors, objective values, constraint outcomes and features, as well as surrogate checkpoints) are available from the corresponding author (F.G.) on request without restriction.

\subsection*{Code availability}\label{sec:code-availability}

All code is openly available. \textsc{dmosopt} (version 0.78, commit \texttt{f034729}; GPL-3.0 license) can be installed via pip; its source code, documentation, tutorials and examples are available at \href{https://github.com/dmosopt/dmosopt}{github.com/dmosopt/dmosopt}.
The experiment definitions, surrogate model implementations, constrained Bayesian optimization baseline, analysis notebooks and figure-generation code for this paper are available at \href{https://github.com/GazzolaLab/MultiObjectiveSurrogateOptimization}{github.com/GazzolaLab/MultiObjectiveSurrogateOptimization} (MIT license) and archived on Zenodo~\autocite{gressmann2026msoCode}. Exact package versions are pinned in the repository lock file; all optimizer, surrogate and benchmark settings are specified in the experiment definitions, and the default \textsc{dmosopt} configuration was used unless stated otherwise (\suppnote{sec:supp-loop}).

\subsection*{Acknowledgements}

The authors wish to thank Prannath Moolchand and Darian Hadjiabadi for discussions and work on early versions of the electrophysiological objective functions for neuron model optimization and connectivity structure of the CA1 network model.

\subsection*{Funding}

This work was funded by the NSF Expedition ``Mind in Vitro'' award \#IIS-2123781 and supported through NSF LSCP allocation IBN22011 on the TACC Frontera supercomputer.

\subsection*{Author contributions}

F.G.: Conceptualization; Software; Methodology; Investigation; Formal analysis; Visualization; Writing -- original draft; Writing -- review \& editing.
I.R.: Conceptualization; Software; Methodology; Resources; Writing -- review \& editing.
S.K.: Software; Data curation; Visualization; Writing -- review \& editing.
M.G.: Supervision; Funding acquisition; Project administration; Writing -- review \& editing.
L.R.: Supervision; Funding acquisition; Project administration; Writing -- review \& editing.
I.S.: Supervision; Funding acquisition; Project administration; Writing -- review \& editing.

\subsection*{Competing interests}

The authors declare no competing interests.

\bigskip

\clearpage
\twocolumn[{\centering\LARGE\bfseries Supplementary Information\par\vspace{2em}}]
\setcounter{subsection}{0}
\renewcommand{\theHsubsection}{S.\arabic{subsection}}
\renewcommand{\theHfigure}{S\arabic{figure}}
\renewcommand{\theHtable}{S\arabic{table}}
\renewcommand{\theHalgorithm}{S\arabic{algorithm}}

\setcounter{secnumdepth}{0}
\renewcommand{\thesubsection}{\arabic{subsection}}
\newcommand{\suppnotesec}[2]{\refstepcounter{subsection}\subsection{Supplementary Note~\thesubsection: #1}\label{#2}}
\renewcommand{\thefigure}{S\arabic{figure}}
\renewcommand{\thetable}{S\arabic{table}}
\renewcommand{\thealgorithm}{S\arabic{algorithm}}
\setcounter{figure}{0}\setcounter{table}{0}\setcounter{algorithm}{0}

\suppnotesec{Surrogate Model Architectures and Training Procedure}{sec:supp-architectures}

This section provides the full specification of the neural network surrogate models used throughout the paper. Both architectures share a common base that handles input/output normalization, loss computation, automatic epoch selection, and joint output heads for objectives and constraints. All hyperparameters are summarized in Table~\ref{tab:hyperparameters}.

\subsubsection{FT-Transformer}

The FT-Transformer~\autocite{gorishniyRevisitingDeepLearning2021} adapts the Feature Tokenizer Transformer architecture for tabular regression and classification. Each input parameter is projected into a learned token embedding, and a learnable CLS (classification) token is prepended. The token sequence is processed by a stack of pre-norm Transformer blocks, and the CLS token representation is used for prediction.

\paragraph{Embedding layer.}
A dense layer projects the $n$-dimensional input $\mathbf{x} \in \mathds{R}^{n}$ into $n$ token embeddings of dimension $d$:
\begin{equation}
    \mathbf{E} = \text{Reshape}_{n \times d}\!\left( \mathbf{W}_e \mathbf{x} + \mathbf{b}_e \right), \quad \mathbf{W}_e \in \mathds{R}^{(n \cdot d) \times n},
\end{equation}
where weights and biases are initialized uniformly in $[-d^{-1/2},\, d^{-1/2}]$.

\paragraph{Transformer blocks.}
Each of the $L$ blocks follows a pre-norm residual architecture:
\begin{align}
    \mathbf{h} &= \mathbf{E} + \text{Dropout}_{p_r}\!\left(\text{MHA}\!\left(\text{LN}(\mathbf{E})\right)\right), \\
    \mathbf{E}' &= \mathbf{h} + \text{Dropout}_{p_r}\!\left(\text{FFN}\!\left(\text{LN}(\mathbf{h})\right)\right),
\end{align}
where $\text{LN}$ denotes layer normalization, $\text{MHA}$ is multi-head attention with $H$ heads and key dimension $d_k = \max(8,\, d/H)$, and the feed-forward network is
\begin{equation}
    \text{FFN}(\mathbf{z}) = \text{Dropout}_{p_f}\!\left(\mathbf{W}_2\, \text{GELU}\!\left(\mathbf{W}_1 \mathbf{z} + \mathbf{b}_1\right) + \mathbf{b}_2\right),
\end{equation}
with hidden dimension $d_{\text{ff}} = \lfloor r_{\text{ff}} \cdot d \rfloor$. The final block extracts the CLS token: $\mathbf{z}_{\text{out}} = \mathbf{E}'_{0}$.

\paragraph{Output heads.}
After a final layer normalization, the CLS representation is fed to task-specific linear heads:
\begin{itemize}
    \item \textbf{Objectives:} $\hat{\mathbf{y}} = \mathbf{W}_o \mathbf{z}_{\text{out}} + \mathbf{b}_o \in \mathds{R}^{q}$ (linear activation).
    \item \textbf{Constraints:} $\hat{\mathbf{c}} = \sigma\!\left(\mathbf{W}_c \mathbf{z}_{\text{out}} + \mathbf{b}_c\right) \in [0,1]^{k}$ (sigmoid activation).
\end{itemize}
In joint mode (\texttt{c+o}), both heads are active; in objective-only (\texttt{o}) or constraint-only (\texttt{c}) mode, only the respective head is used.

\paragraph{Adaptive scaling.}
For problems with a large number of input parameters ($n > 64$), tokens are grouped by averaging over groups of size $\lceil n/64 \rceil$ to cap the effective sequence length at 64. For problems with a large number of targets ($q + k > 100$), the architecture is automatically scaled down: the number of blocks is capped at 2, the embedding dimension at 128, and the parameter group size is increased (Table~\ref{tab:fttransformer-adaptive}).

\begin{table}[h]
\centering
\caption{Adaptive scaling rules for FT-Transformer.}
\label{tab:fttransformer-adaptive}
\begin{tabular}{lcc}
\toprule
Condition & $q+k > 100$ & $q+k > 200$ \\
\midrule
Max blocks $L$ & 2 & 2 \\
Max embedding dim $d$ & 128 & 96 \\
Max heads $H$ & 4 & 3 \\
Min group size & 2 & 4 \\
Pool every $n$ blocks & 1 & 1 \\
\bottomrule
\end{tabular}
\end{table}

\subsubsection{ResNet}

The ResNet surrogate follows the simple residual network design for tabular data~\autocite{gorishniyRevisitingDeepLearning2021}.

\paragraph{Architecture.}
An input projection maps $\mathbf{x}$ to a $d_{\text{block}}$-dimensional representation. This is followed by $B$ residual blocks, each consisting of:
\begin{equation}
    \mathbf{x}' = \mathbf{x} + \text{Dropout}_{p_2}\!\left(\mathbf{W}_2\, \text{Dropout}_{p_1}\!\left(\text{ReLU}\!\left(\mathbf{W}_1\, \text{BN}(\mathbf{x}) + \mathbf{b}_1\right)\right) + \mathbf{b}_2\right),
\end{equation}
where $\text{BN}$ is batch normalization, $\mathbf{W}_1 \in \mathds{R}^{d_h \times d_{\text{block}}}$ with $d_h = \lfloor m \cdot d_{\text{block}} \rfloor$, and $\mathbf{W}_2 \in \mathds{R}^{d_{\text{block}} \times d_h}$. The output heads are identical to those of the FT-Transformer.

\subsubsection{Training Procedure}\label{sec:supp-training}

\paragraph{Optimizer.} Adam with learning rate $\eta = 0.001$.

\paragraph{Batch size.} 2048 samples.

\paragraph{Automatic epoch selection.}
The number of training epochs is determined automatically via $K$-fold cross-validation ($K=3$, standard KFold). The procedure is:
\begin{enumerate}
    \item Compute a timeout budget: $E_{\max} = \max\!\left(25,\, \min\!\left(\lfloor 10^8 / N \rfloor,\, 10{,}000\right)\right)$, where $N$ is the number of training samples.
    \item Set epoch increment: $\Delta E = \max\!\left(10,\, \lfloor E_{\max} / 10 \rfloor\right)$.
    \item For each fold, train in increments of $\Delta E$ epochs with early stopping (patience $= \min(250, E_{\max})$ epochs, monitoring validation loss). Training also terminates on NaN.
    \item The final epoch count is the mean of the stopped epochs across all folds.
\end{enumerate}

\paragraph{Loss functions.}
\begin{itemize}
    \item \textbf{Objectives:} Mean squared error (MSE) by default. Alternative losses available include Huber, Log-Cosh, weighted Log-Cosh ($\sum_i w_i \cdot \text{log\_cosh}(y_i - \hat{y}_i)$ with $w_i = 1/(|y_i|+1)$), distance-weighted MSE ($\frac{1}{1+|y|} \cdot (y-\hat{y})^2$), and relative error ($|y-\hat{y}|/(|y|+\epsilon)$).
    \item \textbf{Constraints:} Binary cross-entropy.
    \item \textbf{Joint mode:} The total loss is the sum of the objective and constraint losses: $\mathcal{L} = \mathcal{L}_{\text{obj}} + \mathcal{L}_{\text{constr}}$. The two terms are weighted equally, see \suppnotesub{sec:supp-loss-weighting}.
\end{itemize}

\paragraph{GradNorm multi-task balancing (optional).}
When enabled, per-objective learnable weights $w_i$ balance the multi-task loss via the GradNorm algorithm~\autocite{chenGradnormGradientNormalization2018}. Training rates $r_i(t) = \frac{L_i(t)/L_i(0)}{\bar{L}(t)/\bar{L}(0)}$ are computed, and the gradient norm loss
\begin{equation}
    \mathcal{L}_{\text{grad}} = \sum_{i} \left| \|G_{W}^{(i)}\| - \overline{\|G_W\|} \cdot r_i^{\alpha} \right|, \quad \alpha = 0.5,
\end{equation}
where $G_W^{(i)}$ is the gradient of $w_i L_i$ with respect to the last shared layer, is minimized with SGD at learning rate $10^{-4}$. Weights are re-normalized after each step to sum to the number of objectives. GradNorm was not enabled for any of the experiments reported in this paper; it was explored during development but did not yield consistent improvements over the default equal-weighted loss for the benchmark problems considered here.

\subsubsection{Loss Weighting}\label{sec:supp-loss-weighting}

Objective and constraint loss are weighted equally by default. We assessed different weighting choices by re-fitting the surrogate offline on the samples of a completed optimization to ensure an exactly comparable training and test set. We swept the ratio $w_c\!:\!w_o$ of the constraint to the objective weight over a 16-fold range ranging from $0.25$ (objective-dominated) to $4$ (constraint-dominated). The weights are normalized to keep the scale of the loss and effective step size constant. The models were trained for a fixed budget of 600 epochs and scored on the candidates the optimized proposed in the following epoch. Objective prediction is measured by NRMSE, and constraint accuracy is measured by AUROC, which is insensitive to the strong class imbalance of the individual constraints (\suppnotesub{sec:supp-constraint-accuracy}).

Table~\ref{tab:loss-weighting} reports the outcome. Notably, the weighting varies the constraint cross-entropy monotonically as expected (up to a factor of $1.38$ between the extreme ends of the sweep). However, constraint AUROC stays within $1\%$ of the equally weighted loss across the entire range while the objective NRMSE shows no systematic trend of variation. Thus, driving the constraint loss lower does not yield a better feasibility model and the objective prediction is not measurably harmed, suggesting that the two heads are not competing for backbone capacity at these ratios. We consequently use the unweighted loss throughout, without treating the ratio as a hyperparameter.

\begin{table*}[htbp]
\centering
\caption{\textbf{Effect of head weighting on the joint loss.} Constraint and objective prediction under a 16-fold sweep of the constraint-to-objective loss ratio $w_c\!:\!w_o$, normalized to a constant sum. All values are ratios to the equally weighted model fit on the same data with the same seed with $1.0$ denoting no change (mean $\pm$ standard deviation over $n = 9$ fits from 3 CA1 populations $\times$ 3 seeds). The constraint cross-entropy (BCE) is reported at the converged validation value while AUROC and NRMSE are the best values scored on the set of candidates the optimizer proposed in the following epoch. The loss term the weighting targets moves monotonically across the sweep while neither accuracy does. Fits use the ResNet backbone at the production batch size on the accumulated samples ($N = 2300$ per population) of a completed run.}
\label{tab:loss-weighting}
\begin{tabular}{lrrr}
\toprule
$w_c\!:\!w_o$ & Constraint BCE $\downarrow$ & Constraint AUROC $\uparrow$ & Objective NRMSE $\downarrow$ \\
\midrule
0.25 & $1.135 \pm 0.078$ & $0.993 \pm 0.009$ & $1.044 \pm 0.156$ \\
0.5 & $1.060 \pm 0.054$ & $1.002 \pm 0.011$ & $0.960 \pm 0.171$ \\
1 (unweighted) & $1.000 \pm 0.000$ & $1.000 \pm 0.000$ & $1.000 \pm 0.000$ \\
2 & $0.904 \pm 0.085$ & $1.011 \pm 0.012$ & $0.963 \pm 0.161$ \\
4 & $0.822 \pm 0.108$ & $1.009 \pm 0.023$ & $0.873 \pm 0.198$ \\
\bottomrule
\end{tabular}
\end{table*}

\subsubsection{Input Normalization}

When parameter bounds $\mathbf{x}_{lb}$ and $\mathbf{x}_{ub}$ are available (as in all our experiments), inputs are normalized via bounds normalization:
\begin{equation}
    \tilde{x}_j = \frac{x_j - x_{lb,j}}{x_{ub,j} - x_{lb,j}}.
\end{equation}
Otherwise, an adaptive layer normalization is applied (z-score based on the training data statistics).

\subsubsection{Output Normalization}

Objective targets are normalized before training using a ``range'' scheme that provides a continuous down-weighting of large objective values to focus most of the learning on predicting lower objective values correctly (since that is the region of the search space most relevant to make optimization progress):
\begin{enumerate}
    \item Per-objective min-max normalization: $\bar{y}_j = \frac{y_j - y_{\min,j}}{y_{\max,j} - y_{\min,j}}$, clipped to $[0, 1]$.
    \item Global rescaling to the range $[y_{\min}^*, y_{\max}^*]$ where $y_{\min}^* = \max_j y_{\min,j}$ and $y_{\max}^* = \max_j y_{\max,j}$.
    \item Log-transform: $\tilde{y} = \log(1 + \bar{y})$.
\end{enumerate}
The inverse transform is applied at prediction time. Constraint targets (binary 0/1) are not normalized.

\subsubsection{Data Preprocessing}

\paragraph{NaN handling.} Samples with any NaN objective value are removed (default behavior). Alternatively, NaN values can be replaced with $2 \times \max_j y_j$ per objective.

\paragraph{Outlier filtering.} When enabled (configurable threshold $\theta$), log-transformed objectives $\log(y+1)$ are z-scored, and samples with $|z| > \theta$ are removed.

\paragraph{Infeasible exclusion.} Optionally, samples that violate any constraints can be excluded from training.

\begin{table*}[h]
\centering
\caption{\textbf{Surrogate hyperparameters.} Default hyperparameters for both surrogate model architectures.}
\label{tab:hyperparameters}
\small
\begin{tabular}{llcc}
\toprule
Category & Parameter & FT-Transformer & ResNet \\
\midrule
\multirow{6}{*}{Architecture}
 & Number of blocks $L$ / $B$ & 3 & 2 \\
 & Block dimension $d$ / $d_{\text{block}}$ & 128 & 192 \\
 & Hidden multiplier $r_{\text{ff}}$ / $m$ & 2.0 & 2.0 \\
 & Number of heads $H$ & 4 & -- \\
 & Embedding dim per head & 32 & -- \\
 & Pooling & CLS & -- \\
\midrule
\multirow{3}{*}{Regularization}
 & Attention dropout $p_a$ & 0.1 & -- \\
 & FFN dropout $p_f$ & 0.05 & -- \\
 & Residual / block dropout $p_r$ / $p_1$ & 0.1 & 0.15 \\
 & Second dropout $p_2$ & -- & 0.0 \\
\midrule 
\multirow{2}{*}{Activation}
 & FFN / hidden & GELU & ReLU \\
 & Normalization & LayerNorm & BatchNorm \\
\midrule
\multirow{4}{*}{Training}
 & Optimizer & \multicolumn{2}{c}{Adam ($\eta = 0.001$)} \\
 & Batch size & \multicolumn{2}{c}{2048} \\
 & Epoch selection & \multicolumn{2}{c}{3-fold CV + early stopping} \\
 & Early stopping patience & \multicolumn{2}{c}{250 epochs} \\
\midrule
\multirow{2}{*}{Loss}
 & Objectives & \multicolumn{2}{c}{MSE} \\
 & Constraints & \multicolumn{2}{c}{Binary cross-entropy} \\
\midrule
\multirow{2}{*}{Normalization}
 & Inputs & \multicolumn{2}{c}{Bounds: $(x - x_{lb})/(x_{ub} - x_{lb})$} \\
 & Targets & \multicolumn{2}{c}{Range (min-max + log1p)} \\
\bottomrule
\end{tabular}
\end{table*}

\suppnotesec{Gradient-Based Feasibility Solving}{sec:supp-gradient}

This section details the gradient-based procedure by which the trained joint surrogate is used to transform parameter candidates toward feasible, high-performing regions of the search space. This supplements the gradient formulation presented in Equation~1 of the main text.

\subsubsection{Optimization Procedure}

Given a batch of parameter candidates $\mathbf{X} \in \mathds{R}^{B \times n}$ and a trained joint surrogate $f$, we perform iterative gradient descent on $\mathbf{X}$ while keeping the surrogate weights frozen. The procedure optimizes a composite loss over up to four targets:

\paragraph{1. Objective minimization (``objective'').}
The surrogate predicts objective values $\hat{\mathbf{y}} = f_{\mathbf{y}}(\mathbf{x})$, which are de-normalized to the original scale. The loss maximizes dominated hypervolume with respect to a dynamic nadir point $\mathbf{r}$:
\begin{equation}
    \mathcal{L}_{\text{obj}} = -\sum_{i=1}^{B} \prod_{j=1}^{q} \max\!\left(r_j - \hat{y}_{ij} / (\text{nadir}_j + \epsilon),\, 0\right),
\end{equation}
where $\text{nadir}_j = \max\!\left(\max_i \hat{y}_{ij},\, \max_i y_{ij}^{\text{train}}\right)$ and $r_j = 1.1$ is the reference point factor.

\paragraph{2. Constraint feasibility (``constraint'').}
The constraint head prediction $\hat{\mathbf{c}} = f_{\mathbf{c}}(\mathbf{x})$ is driven toward all-feasible via binary focal cross-entropy:
\begin{equation}
    \mathcal{L}_{\text{constr}} = \text{BinaryFocalCE}\!\left(\mathds{1}_{B \times k},\, \hat{\mathbf{c}}\right).
\end{equation}

When a constraint is satisfied by no training sample, its target column in $\mathds{1}_{B \times k}$ is uniformly zero and $\mathcal{L}_{\text{constr}}$ drives the corresponding logit downward everywhere, so that head's contribution to the composite gradient effectively vanishes. Steering toward such a constraint then relies on the remaining constraint heads, the objective term, and the diversity of the evolutionary population until a satisfying sample first appears.

\paragraph{3. Exploration (``distance'').}
To encourage diverse sampling, pairwise Euclidean distances between the current candidates and the training data are maximized:
\begin{equation}
    \mathcal{L}_{\text{dist}} = -\frac{1}{B} \sum_{i=1}^{B} \frac{1}{N} \sum_{j=1}^{N} \left\| \tilde{\mathbf{x}}_i - \tilde{\mathbf{x}}_j^{\text{train}} \right\|_2,
\end{equation}
where $\tilde{\mathbf{x}}$ denotes bounds-normalized parameters.

\paragraph{4. Non-negativity (``zero'').}
An optional penalty for negative predicted objective values:
\begin{equation}
    \mathcal{L}_{\text{zero}} = \sum_{i,j} \left[\text{ReLU}(-\hat{y}_{ij})\right]^2.
\end{equation}

\subsubsection{Multi-Loss Gradient Balancing}

When multiple targets are active, their gradients are balanced by rescaling each gradient to match the $L^2$ norm of the first (reference) gradient:
\begin{equation}
    \mathbf{g}_{\text{total}} = \sum_{t} \frac{\|\nabla_{\mathbf{X}} \mathcal{L}_1\|}{\|\nabla_{\mathbf{X}} \mathcal{L}_t\| + \epsilon} \cdot \nabla_{\mathbf{X}} \mathcal{L}_t.
\end{equation}

\subsubsection{Optimizer and Convergence}

\begin{itemize}
    \item \textbf{Optimizer:} Adam (learning rate $0.001$) for the standard case; SGD when using the ``zero-infeasible'' mode.
    \item \textbf{Maximum iterations:} 1000 (configurable).
    \item \textbf{Parameter bounds:} After each gradient step, parameters are clipped to $[x_{lb,j},\, x_{ub,j}]$.
    \item \textbf{Plateau detection:} The interquartile range (IQR) of the loss over the last 50 iterations is monitored. If IQR / $|\text{median}|$ $< 0.01$, optimization is terminated early.
\end{itemize}

\subsubsection{Integration with Black-Box Optimization}

The gradient-based feasibility solving does not replace the black-box optimizer but augments it. In each optimization epoch, the NSGA-II population is split: the elite half is preserved, and the remaining candidates are transformed via gradient descent on the surrogate. This hybrid strategy avoids the local-minima problem of pure gradient descent while exploiting smooth surrogate gradients for directed search.

\begin{algorithm}[h]
\caption{Surrogate-gradient-augmented optimization epoch}
\label{alg:hybrid}
\begin{algorithmic}[1]
\Require Population $\mathbf{P} = \{\mathbf{x}_1, \ldots, \mathbf{x}_M\}$, trained surrogate $f$
\State Sort $\mathbf{P}$ by non-dominated rank and crowding distance
\State $\mathbf{P}_{\text{elite}} \leftarrow \mathbf{P}[1:M/2]$ \Comment{preserve top half}
\State $\mathbf{P}_{\text{explore}} \leftarrow \mathbf{P}[M/2+1:M]$
\State $\mathbf{P}_{\text{explore}}' \leftarrow \textsc{MakeFeasible}(f, \mathbf{P}_{\text{explore}})$ \Comment{gradient descent on surrogate}
\State $\mathbf{P}' \leftarrow \mathbf{P}_{\text{elite}} \cup \mathbf{P}_{\text{explore}}'$
\State Evaluate $\mathbf{P}'$ on true simulator \Comment{expensive}
\State \Return $\mathbf{P}'$
\end{algorithmic}
\end{algorithm}

\subsubsection{Trace Sampling for Surrogate Bootstrapping}

During gradient descent, intermediate parameter states form a trajectory (``trace'') through the search space. To extract maximally informative samples from this trajectory for bootstrapping the surrogate model (as in Figure~4B of the main text), we apply a diversity filter:
\begin{enumerate}
    \item Predict objectives $\hat{\mathbf{y}}_t$ at each step $t$ of the trajectory.
    \item Normalize predictions via min-max scaling.
    \item Iteratively remove the point with the smallest minimum pairwise Euclidean distance to any remaining point, until $k$ points remain.
\end{enumerate}
This yields a set of $k$ diverse samples spanning the gradient descent path, which are then evaluated on the true simulator to augment the training set.

\suppnotesec{Sensitivity-Informed Sampling}{sec:supp-sensitivity}

\subsubsection{Surrogate Gradient Sensitivity}

The partial derivatives of the trained surrogate provide a natural estimate of parameter sensitivity. For each parameter $x_j$ and objective $\hat{y}_k$, the elasticity (relative sensitivity) is:
\begin{equation}
    S_j^{(k)} = \frac{\partial \hat{y}_k}{\partial x_j} \cdot x_j,
\end{equation}
computed via automatic differentiation (\texttt{tf.GradientTape}) and averaged (in absolute value) across all input samples:
\begin{equation}
    \bar{S}_j = \frac{1}{N} \sum_{i=1}^{N} |S_j^{(k)}(\mathbf{x}_i)|.
\end{equation}
This computation is batched (batch size 1024) for memory efficiency and is a by-product of the trained surrogate, requiring no additional simulation evaluations.

\subsubsection{Mapping to NSGA-II Distribution Indices}

Sensitivity values are converted to NSGA-II crossover and mutation distribution indices, which control the spread of offspring around parents:
\begin{equation}
    \eta_j^{\text{cross}} = \eta_j^{\text{mut}} = \text{clip}\!\left(1 + |\bar{S}_j| \times 20, \;[1, 30]\right).
\end{equation}
High sensitivity (large $|\bar{S}_j|$) yields a large distribution index, producing offspring close to the parent in that dimension (small perturbation). Low sensitivity yields a small distribution index, enabling larger exploration steps. This focuses the search on fine-tuning sensitive parameters while broadly exploring insensitive ones.

\subsubsection{Cross-Check Control Experiment}

To verify that the sensitivity-informed sampling genuinely improves optimization, we include a control experiment with inverted distribution indices:
\begin{equation}
    \eta_j^{\text{inv}} = 21 - \eta_j,
\end{equation}
which applies small perturbations to insensitive parameters and large perturbations to sensitive ones---the opposite of the intended strategy.

\subsubsection{Comparison with Established Methods}

We compare surrogate-gradient sensitivity against two established methods:
\begin{itemize}
    \item \textbf{FAST} (Fourier Amplitude Sensitivity Test)~\autocite{mcraeGlobalSensitivityAnalysis1982}: A variance-based global method implemented via SALib.
    \item \textbf{DGSM} (Derivative-based Global Sensitivity Measure)~\autocite{sobolDerivativeBasedGlobal2010}: Finite-difference derivatives, also via SALib.
\end{itemize}
Both FAST and DGSM require dedicated simulation budgets for their sampling designs. In contrast, surrogate-gradient sensitivity operates on the already-trained surrogate model, requiring only forward and backward passes over the training data.

\suppnotesec{Optimization Loop and Dynamic Sampling}{sec:supp-loop}

\subsubsection{Iterative Pipeline}

The optimization follows an iterative surrogate-assisted loop (Algorithm~\ref{alg:loop}).

\begin{algorithm}[h]
\caption{Surrogate-assisted optimization loop}
\label{alg:loop}
\begin{algorithmic}[1]
\Require Parameter bounds $[\mathbf{x}_{lb}, \mathbf{x}_{ub}]$, initial sample count $N_0$, evaluations per epoch $N_e$, number of epochs $E$
\State $\mathbf{X}_0 \leftarrow \text{SymmetricLatinHypercube}(N_0)$
\State $(\mathbf{Y}_0, \mathbf{C}_0) \leftarrow \text{Evaluate}(\mathbf{X}_0)$
\For{$e = 1, \ldots, E$}
    \State Train surrogate $f$ on $(\mathbf{X}_{0:e-1}, \mathbf{Y}_{0:e-1}, \mathbf{C}_{0:e-1})$ with automatic epoch selection
    \State Compute sensitivity $\bar{S}_j$ from $f$ \Comment{Optional}
    \State Run NSGA-II on $f$ with sensitivity-informed distribution indices to generate $N_e$ candidates
    \State \textbf{If} feasibility solving enabled: transform candidates via gradient descent on $f$
    \State $(\mathbf{Y}_e, \mathbf{C}_e) \leftarrow \text{Evaluate}(\text{candidates})$ \Comment{Expensive simulation}
    \State Append results to cumulative dataset
\EndFor
\end{algorithmic}
\end{algorithm}

\subsubsection{Configuration}

\paragraph{Evaluations per epoch.} Default: 100 problem evaluations per epoch. In the dynamic sampling variant, this is split into 4 sub-iterations of 25 samples each, with surrogate retraining between sub-iterations.

\paragraph{Initial sampling.} Symmetric Latin Hypercube (SLHC) sampling is used for the initial sample set. The default count is 100 samples \emph{per parameter}, so the initial sampling scales with dimensionality.

\paragraph{Stopping criteria.} Two modes are supported:
\begin{itemize}
    \item \textbf{Fixed budget:} A predetermined number of epochs $E$ (25 for single-cell problems, 50 for the network problem).
    \item \textbf{Dynamic:} A configurable convergence criterion evaluated as a Python expression on the optimization history, e.g., \texttt{"iteration > 3 and max(recent('ecov', 3)) < 0.1"}. If fewer than 3 unique constraint patterns are observed, the surrogate falls back from joint (\texttt{c+o}) to objective-only (\texttt{o}) mode to avoid degenerate classification.
\end{itemize}

\paragraph{Feasibility-model ranking (optional, not used).} Independently of the joint surrogate, \textsc{dmosopt} can integrate separate feasibility models in the ranking of elite samples. Candidates are ordered by non-domination rank first, followed by crowding distance, and finally predicted feasibility. The re-ranking cannot change the direction in which samples move so it presents the kind of heuristic coupling of separately trained models that the joint surrogate gradient is designed to replace. It was disabled in all experiments reported here.

\paragraph{Default configuration.} All experiments reported in this paper used the default configuration as defined in the \textsc{dmosopt} source code, including the default NaN handling (removal), no outlier filtering, no infeasible exclusion, and fixed-budget stopping. The only experiment-specific settings were the number of epochs, population size, and surrogate architecture, as detailed in the respective benchmark sections.

\suppnotesec{dmosopt Framework and Distributed Execution}{sec:supp-framework}

\subsubsection{Software Architecture}

\textsc{dmosopt} (\textbf{D}istributed \textbf{M}ulti-\textbf{O}bjective \textbf{S}urrogate-Assisted \textbf{Opt}imization) is an open-source Python framework built on top of MPI for high-performance computing scalability. The architecture follows a controller-worker model:
\begin{itemize}
    \item \textbf{Controller:} Manages the optimization state, trains the surrogate model, selects candidate solutions, and dispatches evaluation tasks.
    \item \textbf{Workers:} Receive parameter vectors from the controller, run the (potentially expensive) simulation, and return objective values, features, and constraint satisfaction.
\end{itemize}

Communication between controller and workers uses MPI point-to-point messaging. The surrogate model resides on the controller, avoiding data transfer overhead. Optionally, surrogate training can be offloaded to a remote GPU node via a (remote-)filesystem-based swap queue mechanism.

\subsubsection{Configurable Components}

\begin{itemize}
    \item \textbf{Optimizers:} NSGA-II~\autocite{debFastElitistMultiobjective2002a}, AGE-MOEA~\autocite{panichella_adaptive_2019}, SMPSO~\autocite{nebroSMPSONewPSObased2009}, CMA-ES~\autocite{hansenCMAEvolutionStrategy2006}.
    \item \textbf{Surrogates:} Gaussian Process Regression (GPR; fits an independent Mat\'ern 5/2 GP per objective, without cross-output correlations), multitask exact Gaussian process (MEGP; a single GP fitted jointly over all objectives, whose rank-1 coregionalization kernel models cross-output correlations), ResNet, FT-Transformer (this work), and custom models via a plugin interface. GP surrogates are trained exclusively on feasible samples, whereas neural network surrogates train on all available data including infeasible samples.
    \item \textbf{Samplers:} Symmetric Latin Hypercube (SLHC), Latin Hypercube (LHC), Monte Carlo (MC), Sobol sequences, Good Lattice Points (GLP).
    \item \textbf{Sensitivity methods:} Surrogate gradient, FAST, DGSM.
\end{itemize}

\subsubsection{Computational Environment}

All large-scale experiments (Figs.~\ref{fig:figure2-surrogate-opt}--\ref{fig:figure5-large-scale}) were executed on the Frontera supercomputer at the Texas Advanced Computing Center (TACC) using the software dependencies and versions detailed in Table~\ref{tab:software}. Each compute node contains two Intel Xeon Platinum 8280 (Cascade Lake) processors with 28 cores each (56 cores per node, 2.7\,GHz base clock, 192\,GB RAM). MPI communication used MVAPICH2.

For the CA1 network problem (Fig.~\ref{fig:figure5-large-scale}), each worker used 300 cores (approximately 5.4 nodes) for a single simulation evaluation, with a wall-clock time of ${\sim}42.3$\,s per evaluation. Surrogate training was performed on a Frontera GPU node (NVIDIA Quadro RTX 5000); for the single-cell problems, FT-Transformer training completed in ${\sim}150$-$235$\,s per epoch depending on cell type, which is negligible relative to the ${\sim}42.3$\,s $\times$ 100 evaluation budget per epoch.

\begin{table*}[h]
\centering
\caption{\textbf{Software environment.} Software dependencies and the exact versions used.}
\label{tab:software}
\small
\begin{tabular}{lll}
\toprule
Package & Version & Purpose \\
\midrule
Python & 3.12 & Runtime \\
dmosopt & 0.78 (commit \texttt{f034729}) & Optimization framework \\
MiV-Simulator & 0.2.1 (commit \texttt{e440eaa}) & Network simulation \\
NEURON & 8.2.6 & Biophysical neuron simulation \\
TensorFlow & 2.16.2 & Neural network surrogate training \\
PyTorch & 2.5.1 (CPU) & MEGP baseline (via GPyTorch) \\
GPyTorch & 1.14.2 & Gaussian process models \\
BoTorch & 0.18.1 & Constrained Bayesian optimization baseline \\
scikit-learn & 1.7.2 & GPR baseline, metrics, CV splitting \\
SALib & 1.5.1 & Sensitivity analysis (FAST, DGSM) \\
NumPy / SciPy & 1.26.4 / 1.16.2 & Numerics, statistics \\
mpi4py & 4.1.1 & MPI communication \\
h5py & 3.15.1 & HDF5 result storage \\
matplotlib / seaborn & 3.10.7 / 0.13.2 & Visualization \\
machinable & 4.10.6 & Experiment tracking \\
\bottomrule
\end{tabular}
\end{table*}

\suppnotesec{Benchmark Problem Specifications}{sec:supp-benchmarks}

This section provides the full specification of all benchmark problems, including parameter spaces, objectives, constraints, and simulation parameters. The main text (Methods) gives a summary; the details below are intended for reproducibility.

\subsubsection{CA1 Hippocampal Interneuron Single-Cell Models}\label{sec:supp-ca1-single}

We optimize 9 morphologically distinct CA1 hippocampal interneuron types using 2-compartment Pinsky-Rinzel neuron models~\autocite{pinskyIntrinsicNetworkRhythmogenesis1994a} implemented in the NEURON simulator~\autocite{jaeger_neuron_2022}. Each model consists of a somatic and dendritic compartment coupled by an adjustable conductance $g_c$, with Hodgkin-Huxley-type active conductances (Na$^+$, K$^+$, Ca$^{2+}$, KCa).

\subsubsection{Parameter Space}

The search space contains 10-15 parameters per cell type, governing compartment geometry, coupling conductance, maximal channel conductances, passive leak conductances, and calcium dynamics. Table~\ref{tab:ca1-params} lists the parameters for a representative cell type (BS, bistratified). Other cell types share the same parameter set with adjusted bounds; the complete specification for all cell types is available in the code repository.

\begin{table*}[h]
\centering
\caption{\textbf{CA1 interneuron parameter space.} Parameter space for the BS (bistratified) interneuron model.}
\label{tab:ca1-params}
\small
\begin{tabular}{llcc}
\toprule
Parameter & Description & Lower & Upper \\
\midrule
\texttt{pp} & Soma area fraction & 0.1 & 0.9 \\
\texttt{Ltotal} & Total length ($\mu$m) & 10 & 80 \\
\texttt{gc} & Coupling conductance (mS/cm$^2$) & 0.1 & 50 \\
\texttt{soma\_gmax\_Na} & Soma Na$^+$ conductance (S/cm$^2$) & 0.001 & 0.9 \\
\texttt{soma\_gmax\_K} & Soma K$^+$ conductance (S/cm$^2$) & 0.001 & 1.0 \\
\texttt{soma\_g\_pas} & Soma leak conductance (S/cm$^2$) & 0 & 0.01 \\
\texttt{dend\_gmax\_Ca} & Dendritic Ca$^{2+}$ conductance (S/cm$^2$) & 0.0001 & 0.9 \\
\texttt{dend\_gmax\_KCa} & Dendritic KCa conductance (S/cm$^2$) & 0 & 1.0 \\
\texttt{dend\_g\_pas} & Dendritic leak conductance (S/cm$^2$) & 0 & 0.01 \\
\texttt{cm\_ratio} & Capacitance ratio (dend/soma) & 1 & 50 \\
\texttt{dend\_d\_Caconc} & Ca$^{2+}$ diffusion depth ($\mu$m) & 0.1 & 20 \\
\texttt{dend\_beta\_Caconc} & Ca$^{2+}$ decay rate (ms$^{-1}$) & 0.01 & 0.1 \\
\midrule
\multicolumn{4}{l}{\textit{Fixed parameters:} \texttt{global\_diam} = 10\,$\mu$m, \texttt{global\_cm} = 3\,$\mu$F/cm$^2$, \texttt{e\_pas} = $-62$\,mV} \\
\bottomrule
\end{tabular}
\end{table*}

\subsubsection{Objectives}

Four objectives are minimized, each computed as the squared distance from the measured value to the target range $[l, u]$:

\begin{align}
    \text{obj}(v) &= d(v, [l, u])^2 \\
    d(v, [l,u]) &= \begin{cases}
        0 & \text{if } l \leq v \leq u, \\
        \min(|v-l|, |v-u|) & \text{otherwise.}
    \end{cases}
\end{align}

\begin{enumerate}
    \item \textbf{Input resistance error} ($R_{\text{in}}$): Distance from measured $R_{\text{in}}$ to target range.
    \item \textbf{Membrane time constant error} ($\tau$): Distance from measured $\tau$ to target range.
    \item \textbf{Frequency-current (f-I) curve error}: Mean squared range distance across $M$ current injection steps: $\frac{1}{M}\sum_{m=1}^{M} d(f_m, [f_m^l, f_m^u])^2$.
    \item \textbf{Spike amplitude error}: Mean squared range distance of spike amplitudes.
\end{enumerate}

\subsubsection{Constraints}\label{sec:supp-ca1-constraints}

Eight binary constraints determine feasibility:
\begin{enumerate}
    \item \textbf{Monotonic f-I:} $f(I_{m+1}) > f(I_m)$ for all consecutive current steps.
    \item \textbf{$R_{\text{in}}$ range:} $0 < R_{\text{in}} < 1000$\,M$\Omega$.
    \item \textbf{$\tau$ range:} $0 < \tau < 1000$\,ms.
    \item \textbf{Spike amplitude:} Spike amplitude is not NaN (i.e., spikes are detected).
    \item \textbf{First ISI:} All first inter-spike intervals exceed the lower bound.
    \item \textbf{ISI adaptation:} ISI adaptation ratio is not NaN.
    \item \textbf{Pre-stimulus spike count:} No spontaneous spikes before stimulus onset.
    \item \textbf{Initial voltage:} $|V_{\text{init}} - V_{\text{target}}| < 1.0$\,mV.
\end{enumerate}

\subsubsection{Electrophysiological Targets}

Electrophysiological targets for each cell type are derived from literature reports and matched to the corresponding experimental ranges. Table~\ref{tab:ca1-targets} summarizes key targets for all 9 populations.

\begin{table*}[h]
\centering
\caption{\textbf{CA1 electrophysiological targets.} Electrophysiological targets for CA1 interneuron populations. $R_{\text{in}}$: input resistance (M$\Omega$); $\tau$: membrane time constant (ms); $V_{\text{thr}}$: spike threshold (mV). f-I column shows the number of current injection steps.}
\label{tab:ca1-targets}
\small
\begin{tabular}{lccccc}
\toprule
Population & $R_{\text{in}}$ range & $\tau$ range & $V_{\text{thr}}$ & f-I steps & \# params \\
\midrule
Parvalbumin basket cells (PVBC) & [53, 179] & [5, 21] & $-37$ & 6 & 12 \\
CCK basket cells (CCKBC) & [240, 320] & [22, 29] & $-43$ & 4 & 12 \\
Ivy cells (IVY) & [167.5, 367.2] & [50, 70] & $-39$ & 4 & 12 \\
OLM cells (OLM) & [500.5, 600.1] & [30, 50] & $-44$ & 4 & 12 \\
Axo-axonic cells (AAC) & [65, 179] & [10, 14] & $-32$ & 5 & 12 \\
Bistratified cells (BS) & [30.5, 109.1] & [11, 15] & $-32$ & 3 & 12 \\
Neurogliaform cells (NGFC) & [189, 223.4] & [8, 10] & $-30$ & 6 & 12 \\
Schaffer collateral-associated cells (SCA) & [410, 550] & [23, 29] & $-43$ & 4 & 15 \\
Interneuron-specific cells (IS1,2,3) & [128.3, 226.1] & [11.6, 17.6] & $-44.3$ & 6 & 10 \\
\bottomrule
\end{tabular}
\end{table*}

\paragraph{Simulation parameters.} Integration time step $dt = 0.0125$\,ms, recording step $0.01$\,ms, simulation duration 2000\,ms, temperature 36\,$^\circ$C.

\paragraph{Passive property measurement.} Input resistance is measured as $R_{\text{in}} = \Delta V / \Delta I$ from a subthreshold current step. The membrane time constant $\tau$ is obtained by fitting $V(t) = V_0 + a \cdot \exp(-t/\tau)$ via \texttt{scipy.optimize.curve\_fit}, requiring a minimum signal-to-noise ratio of 20.

\subsubsection{Motoneuron Model}\label{sec:supp-motoneuron}

The motoneuron model is a 2-compartment (soma + dendrite) NEURON model with Na$^+$, K$^+$, KCa, CaN (soma), and CaL, CaN, KCa (dendrite) conductances, based on the experimental characterization by Booth et al.~\autocite{boothCompartmentalModelVertebrate1997} and Miles et al.~\autocite{milesFunctionalPropertiesMotoneurons2004}.

\begin{table*}[h]
\centering
\caption{\textbf{Motoneuron parameter space.} Narrow range (used in the standard optimization) and wide range (used in Fig.~\ref{fig:figure4-highly-constraint}).}
\label{tab:mn-params}
\small
\begin{tabular}{llcccc}
\toprule
& & \multicolumn{2}{c}{Narrow range} & \multicolumn{2}{c}{Wide range} \\
\cmidrule(lr){3-4} \cmidrule(lr){5-6}
Parameter & Description & Lower & Upper & Lower & Upper \\
\midrule
\texttt{gc} & Coupling conductance & 0.1 & 2 & 0 & 2 \\
\texttt{soma\_gmax\_Na} & Somatic Na$^+$ & 0.1 & 0.3 & 0 & 1 \\
\texttt{soma\_gmax\_K} & Somatic K$^+$ & 0.01 & 0.3 & 0 & 1 \\
\texttt{soma\_gmax\_KCa} & Somatic KCa & 0.0001 & 0.01 & 0 & 1 \\
\texttt{soma\_gmax\_CaN} & Somatic CaN & 0.00001 & 0.03 & 0 & 1 \\
\texttt{soma\_g\_pas} & Somatic leak & 0.00001 & 0.01 & 0 & 1 \\
\texttt{dend\_gmax\_CaL} & Dendritic CaL & 0.00001 & 0.001 & 0 & 1 \\
\texttt{dend\_gmax\_CaN} & Dendritic CaN & 0.00001 & 0.001 & 0 & 1 \\
\texttt{dend\_gmax\_KCa} & Dendritic KCa & 0.0001 & 0.005 & 0 & 1 \\
\texttt{dend\_g\_pas} & Dendritic leak & 0.00001 & 0.01 & 0 & 1 \\
\texttt{cm\_ratio} & Capacitance ratio & 1.0 & 40.0 & 1 & 40 \\
\bottomrule
\end{tabular}
\end{table*}

\paragraph{Parameters.} The optimized parameters are listed in Table~\ref{tab:mn-params}. Additionally, the model used the following fixed parameters: \texttt{soma\_f\_Caconc} = 0.004, \texttt{soma\_alpha\_Caconc} = 1, \texttt{soma\_kCa\_Caconc} = 8, \texttt{dend\_f\_Caconc} = 0.004, \texttt{dend\_alpha\_Caconc} = 1, \texttt{dend\_kCa\_Caconc} = 8, \texttt{global\_diam} = 5\,$\mu$m, \texttt{global\_cm} = 2\,$\mu$F/cm$^2$, \texttt{e\_pas} = $-62$\,mV, \texttt{pp} = 0.1, \texttt{Ltotal} = 120\,$\mu$m.

\paragraph{Targets.} $R_{\text{in}} \in [540, 598]$\,M$\Omega$; $\tau \in [16.308, 19.375]$\,ms; threshold $= -37$\,mV; 7 f-I current steps at $I \in \{20, 30, 40, 50, 60, 70, 80\}$\,pA with target rates $\{3.88, 9.09, 11.75, 14.29, 15.96, 16.58, 18.25\}$\,Hz; spike amplitude $\in [60, 80]$\,mV; ISI adaptation ratio $\in [1.2, 1.6]$.

\paragraph{Objectives and constraints.} Same as the CA1 models (\suppnotesub{sec:supp-ca1-single}).

\subsubsection{CA1 Hippocampal Network Model}\label{sec:supp-network}

The CA1 network model is a full-scale biophysical simulation of the hippocampal CA1 microcircuit based on previously published work~\autocite{bezaireInterneuronalMechanismsHippocampal2016}, implemented in the MiV Simulator~\autocite{upadhyayMiVSimulatorComputationalFramework2023} and simulated via NEURON with CoreNEURON.

\paragraph{Network architecture.} The model contains $836{,}970$ neurons across 15 cell populations, connected by $\sim$10 billion synaptic projections. Major populations include pyramidal cells (PYR, $N = 311{,}500$), CA3 afferents ($N = 204{,}700$), entorhinal cortex afferents (EC, $N = 250{,}000$), and 12 interneuron populations (PVBC, CCKBC, AAC, BS, OLM, IVY, NGFC, IS, SCA, and others).

\paragraph{Slice composition.}

Large numbers of evaluations of the full 836,970-neuron model within the surrogate optimization loop are infeasible on the HPC hardware allocation available to us. Optimization was therefore performed on a subset (``slice'') of the network comprising approximately 8,700 neurons distributed across all 15 cell populations, with per-population counts listed in Supplementary Table~\ref{tab:ca1-slice-counts}.

\begin{table}[h]
\centering
\caption{Neuron counts per population in the CA1 network slice used for surrogate optimization.}
\label{tab:ca1-slice-counts}
\begin{tabular}{lc}
\toprule
Population & Slice count \\
\midrule
Pyramidal (PYR)                              & {7,825}  \\
Parvalbumin basket cells (PVBC)              & {142} \\
CCK basket cells (CCKBC)                     & {104} \\
Ivy cells (IVY)                              & {217} \\
Neurogliaform cells (NGFC)                   & {93} \\
OLM cells (OLM)                              & {43} \\
Axo-axonic cells (AAC)                       & {37} \\
Bistratified cells (BS)                      & {63} \\
Schaffer collateral-associated cells (SCA)   & {12} \\
Interneuron-specific cells (IS1,2,3)         & {176} \\
\midrule
Total                               & \textbf{8{,}712} \\
\midrule
\multicolumn{2}{l}{Network inputs} \\
\midrule
CA2 afferents                                & {1,017} \\
CA3 afferents                                & {5,181} \\
Entorhinal cortex (EC) afferents             & {6,222} \\
\midrule
Total                               & \textbf{12{,}420} \\
\bottomrule
\end{tabular}
\end{table}

All projection pathway types and single-cell biophysical details were retained. The 142 optimizable synaptic weight parameters span the same projection pathways as in the full model.

\paragraph{Optimizable parameters.} The search space comprises $n = 142$ synaptic weight parameters governing connection strengths across all projection pathways. Each parameter controls the weight of a specific pre-synaptic population $\to$ post-synaptic compartment $\to$ receptor type projection (e.g., AMPA, NMDA, GABA\textsubscript{A}, GABA\textsubscript{B}), with bounds of the form $[0.1, u]$ where $u \in \{2, 10, 20\}$ depending on connection type.

\paragraph{Objectives.} Per-population firing rate, mean fraction of active cells, and standard deviation of the fraction of active cells, measured in 50\,ms bins, for each of 12 target populations (PYR and 11 interneuron types), yielding 36 objectives (3 metrics $\times$ 12 populations). Objectives are formulated as absolute deviations from experimentally determined target ranges.

\paragraph{Constraints.} Each of the 12 target populations must exhibit a positive firing rate, yielding 12 binary constraints.

\paragraph{Optimizer settings.} NSGA-II with population size 100, 10 generations per epoch, resampling fraction 0.1, up to 50 epochs. Surrogates compared: GPR (Mat\'ern 5/2), MEGP, and joint c+o FT-Transformer.

\paragraph{Computational cost.} Each simulation evaluation requires ${\sim}42.3$\,s on 300 CPU cores (${\sim}5.4$ Frontera nodes). At 100 evaluations per epoch and 50 epochs, the total compute cost ranges from ${\sim}60$ to ${\sim}300$ CPU-days depending on the surrogate method.

\suppnotesec{Evaluation Metrics}{sec:supp-metrics}

\subsubsection{Optimization Quality Metrics}

\paragraph{Normalized Hypervolume (HV).} The hypervolume indicator measures the volume of objective space dominated by a Pareto front approximation $A$ with respect to a reference point $\mathbf{r}$:
\begin{equation}
    \text{HV}(A) = \text{Vol}\!\left(\bigcup_{\mathbf{a} \in A} \prod_{j=1}^{q} [a_j,\, r_j]\right).
\end{equation}
We normalize the hypervolume by using a shared nadir point computed as the component-wise maximum across all compared methods for each problem instance. The reference point is set to $r_j = 1.1 \times \text{nadir}_j$. The theoretical maximum normalized hypervolume for $q = 4$ objectives is $1.1^4 \approx 1.4641$.

\paragraph{HV Area Under the Curve (HV-AUC).} The area under the hypervolume convergence curve over optimization epochs, computed via the trapezoidal rule (\texttt{sklearn.metrics.auc}). Higher values indicate faster convergence.

\paragraph{Inverted Generational Distance (IGD).} The average minimum Euclidean distance from each point in a reference front $R$ to the closest point in the approximation front $A$:
\begin{equation}
    \text{IGD}(A, R) = \frac{1}{|R|} \sum_{\mathbf{r} \in R} \min_{\mathbf{a} \in A} \|\mathbf{a} - \mathbf{r}\|_2.
\end{equation}
Lower values indicate better coverage and convergence to the reference front.

\paragraph{Additive $\varepsilon$-indicator.}
The minimum additive shift $\varepsilon$ such that every point in the reference front $B$ is weakly dominated by at least one point in the approximation $A$:
\begin{equation}
    I_{\varepsilon+}(A, B) = \max_{\mathbf{b} \in B} \min_{\mathbf{a} \in A} \max_{j=1,\ldots,q} (a_j - b_j).
\end{equation}
A value $\leq 0$ means $A$ dominates $B$; smaller values are better.

\paragraph{Set Coverage (C-metric).}
The fraction of solutions in $B$ that are dominated by at least one solution in $A$:
\begin{equation}
    C(A, B) = \frac{|\{\mathbf{b} \in B : \exists\, \mathbf{a} \in A,\; \mathbf{a} \preceq \mathbf{b}\}|}{|B|}.
\end{equation}

\subsubsection{Surrogate Quality Metrics}

\paragraph{Normalized Root Mean Squared Error (NRMSE).}
\begin{equation}
    \text{NRMSE} = \frac{1}{q} \sum_{j=1}^{q} \frac{\sqrt{\frac{1}{N}\sum_{i=1}^{N}(y_{ij} - \hat{y}_{ij})^2}}{\max_i y_{ij} - \min_i y_{ij}}.
\end{equation}

\suppnotesec{Trials, Error Bars, and Statistical Tests}{sec:supp-statistics}

\subsubsection{Independent Trials}

Each optimization experiment was repeated with independent random seeds to enable statistical comparison. The number of independent trials per experiment type is summarized in Table~\ref{tab:trials}.

\begin{table*}[h]
\centering
\caption{\textbf{Independent trials per experiment.} Number of independent trials (replicates) per experiment. Each trial uses a distinct random seed generated by the experiment management framework (\textsc{machinable}).}
\label{tab:trials}
\small
\begin{tabular}{llc}
\toprule
Experiment & Figures & Trials \\
\midrule
CA1 single-cell surrogate optimization & \ref{fig:figure2-surrogate-opt}, \ref{fig:figure3-joint-modelling}, \ref{fig:epsilon-performance}, \ref{fig:all-convergence} & 3 \\
CA1 no-surrogate baseline & \ref{fig:figure3-joint-modelling}, \ref{fig:all-convergence} & 3 \\
CA1 optimizer comparison (NSGA-II, AGE-MOEA, SMPSO) & -- (nadir estimation) & 3 \\
CA1 sensitivity analysis (surrogate gradient, FAST, DGSM) & \ref{fig:figure3-joint-modelling}, \ref{fig:sensitivity-all} & 5 (HV), 3 (IGD) \\
CA1 initial sampling strategy comparison & \ref{fig:figure2-surrogate-opt}, \ref{fig:samplers} & 3 \\
CA1 gradient-target comparison (BS) & \ref{fig:bs-gradient} & 3 \\
Motoneuron surrogate optimization (narrow and wide range) & \ref{fig:figure4-highly-constraint}, \ref{fig:baseline-sopt} & 3 \\
Motoneuron gradient-augmented optimization & \ref{fig:figure4-highly-constraint} & 3 \\
Motoneuron random sampling feasibility analysis & \ref{fig:figure4-highly-constraint}, \ref{fig:random-constraints} & 3 \\
Motoneuron surrogate gradient vs.\ black-box optimizers & \ref{fig:sgrad-vs-optimizer} & 3 \\
Motoneuron constrained Bayesian optimization & Table~\ref{tab:motoneuron-constrained-bo} & 3 \\
CA1 network optimization (full-scale) & \ref{fig:figure5-large-scale} & 1 \\
\bottomrule
\end{tabular}
\end{table*}

The CA1 network optimization (Fig.~\ref{fig:figure5-large-scale}) was run once per surrogate method due to the high computational cost (${\sim}60$--$300$ CPU-days per run). All single-cell and motoneuron experiments were replicated 3 times with independent random seeds. The sensitivity analysis was run with 5 replicates; its hypervolume is reported over all 5, whereas the reference-front comparison (IGD) is reported over the 3 replicates for which it was computed.

\subsubsection{Error Bars}

All error bars and shaded uncertainty bands in the figures represent the standard error of the mean (SEM) unless otherwise noted:
\begin{equation}
    \text{SEM} = \frac{\sigma}{\sqrt{n}},
\end{equation}
where $\sigma$ is the sample standard deviation and $n$ is the number of independent trials. Table~\ref{tab:errorbars} details the error representation used in each figure.

\begin{table*}[h]
\centering
\caption{\textbf{Error representation across figures.} Measure of centre, error representation and sample size $n$ for every panel that reports summary statistics. Individual observations are overlaid as dots or thin lines wherever a mean is shown.}
\label{tab:errorbars}
\small
\begin{tabular}{llp{8.5cm}}
\toprule
Figure & Panel / plot & Centre, error, $n$ \\
\midrule
\ref{fig:figure2-surrogate-opt}B & Sampling strategy comparison (HV) & Mean $\pm$ SEM; dots: individual runs; $n = 27$ per strategy (9 populations $\times$ 3 trials) \\
\ref{fig:figure2-surrogate-opt}C & Additive $\varepsilon$-indicator & Bars mean $\pm$ SEM; dots: individual runs (runs beyond the axis drawn at its edge); $n = 27$ per method \\
\ref{fig:figure2-surrogate-opt}D & NRMSE / MAPE box plots & Median, IQR box, 1.5\,IQR whiskers, dots individual fits within whiskers (fliers not drawn); $n = 72$--$77$ per model after outlier filtering of 81 (9 populations $\times$ 9 checkpoints, one run) \\
\ref{fig:figure2-surrogate-opt}E & Inference time & Box plot as in D with individual timings; $n = 81$ timed predictions per model \\
\ref{fig:figure3-joint-modelling}A & Convergence curves (HV vs.\ epoch) & Mean, thin lines individual trials; $n = 3$ trials; inset IGD mean $\pm$ SEM with dots \\
\ref{fig:figure3-joint-modelling}B & Mean rank plot & Mean $\pm$ SEM over $n = 10$ problems; dots: per-problem ranks \\
\ref{fig:figure3-joint-modelling}C & Sensitivity methods & Mean $\pm$ SEM, dots individual trials; $n = 3$ (IGD), $n = 5$ (HV) \\
\ref{fig:figure4-highly-constraint}A & Motoneuron convergence curves & Mean, shaded $\pm 1$ SEM; $n = 3$ trials \\
\ref{fig:figure4-highly-constraint}B & Pareto-front prediction error & Single surrogate fit per condition ($n = 1$), no error bars \\
\ref{fig:figure5-large-scale} & Network results & Single run per method ($n = 1$), no error bars; proxy SEM for panel C from $n = 3$ replicates of the first 5 epochs reported in the caption \\
\ref{fig:samplers} & Per-population sampling comparison & Mean $\pm$ SEM, dots individual trials; $n = 3$ \\
\ref{fig:epsilon-performance} & Per-population $\varepsilon$-indicator & Mean $\pm$ SEM, dots individual trials; $n = 3$ \\
\ref{fig:all-convergence} & All-population convergence & Mean, shaded $\pm 1$ SEM; inset mean $\pm$ SEM with dots; $n = 3$ \\
\ref{fig:bs-gradient} & BS gradient-target comparison & Mean, shaded $\pm 1$ SEM; $n = 3$ \\
\ref{fig:sensitivity-all} & Sensitivity methods, all populations & Mean $\pm$ SEM, dots; $n = 3$ (IGD), $n = 5$ (HV) \\
\ref{fig:baseline-sopt} & Narrow-range motoneuron convergence & Mean, shaded $\pm 1$ SEM; inset mean $\pm$ SEM with dots; $n = 3$ \\
\ref{fig:sgrad-vs-optimizer} & HV-AUC heatmap & Mean $\pm$ s.d.; $n = 3$ \\
\bottomrule
\end{tabular}
\end{table*}

Convergence curves (Figures~\ref{fig:figure3-joint-modelling}, \ref{fig:figure4-highly-constraint} and \ref{fig:all-convergence}) plot the mean normalized hypervolume across trials as the central line with $\pm 1$ SEM shaded bands and, where legible, the individual trials as thin lines. For the HV-AUC heatmaps (Figure~\ref{fig:sgrad-vs-optimizer}), cell annotations report mean $\pm$ standard deviation.

\subsubsection{Statistical Tests}

Significance levels in figures are reported as: $^{*}p < 0.05$, $^{**}p < 0.01$, $^{***}p < 0.001$.

\paragraph{Outlier filtering for pooled metrics.}
Before pooling NRMSE values across populations for statistical testing, per-population z-score filtering is applied: values with $|z| > 2$ (where $z$ is the z-score computed within each population) are excluded. This removes extreme outliers arising from occasional degenerate surrogate fits without discarding population-level variation.

\paragraph{Superfront Composition}

The superfront analysis (Figure~\ref{fig:superfront}) is deterministic and does not involve error bars. Pareto-optimal solutions from all trials of all surrogate methods are pooled into a single merged set for each cell population, non-dominated sorting is applied, and the percentage contribution of each method is computed from the resulting non-dominated front. Since all trials are combined, the superfront reflects the best solutions found across all replicates.

\paragraph{CA1 Network Experiment}

The CA1 network optimization (Fig.~\ref{fig:figure5-large-scale}) was conducted as a single run per surrogate method due to the computational cost (${\sim}42.3$\,s per evaluation $\times$ 300~cores $\times$ up to 5{,}000 total evaluations). Consequently, no error bars or statistical tests are reported for the network-level results. The convergence curves show the hypervolume of the single run for each method. As a proxy for run-to-run variability, three independent replicates (distinct random seeds) of the c+o-FTTransformer configuration were run for the first 5 epochs under a revised objective formulation; their spread, transferred to the main-text curve as the SEM of the fraction of attainable hypervolume gained and as the relative SEM of the hypervolume gain, is reported in the caption of Fig.~\ref{fig:figure5-large-scale}.

\paragraph{Wall-clock time.}
For the CA1 network optimization, the c+o-FTTransformer surrogate reached a solution quality comparable to GPR's final hypervolume 39 epochs earlier, and comparable to MEGP 26 epochs earlier. In wall-clock terms (at 300 cores per evaluation, ${\sim}42.3$\,s per evaluation, 100 evaluations per epoch), this corresponds to a reduction from ${\sim}55$ hours (GPR/MEGP for 49 additional epochs) to ${\sim}10$-$25$ hours (c+o-FTTransformer), a saving of ${\sim}300$-$500$ CPU-days.

\suppnotesec{Additional Results and Analyses}{sec:supp-additional}

\subsubsection{Per-Population Random Sampling and Surrogate Baseline}

Before comparing advanced surrogate architectures, we establish baseline performance using random sampling strategies and Gaussian Process Regression (GPR)-based surrogate optimization. The main text (Figure~2B) presents these baselines averaged across all CA1 interneuron types; here we provide the underlying per-population data to characterize problem-specific variation.

\begin{figure*}[htbp]
\centering
\includegraphics[width=\textwidth]{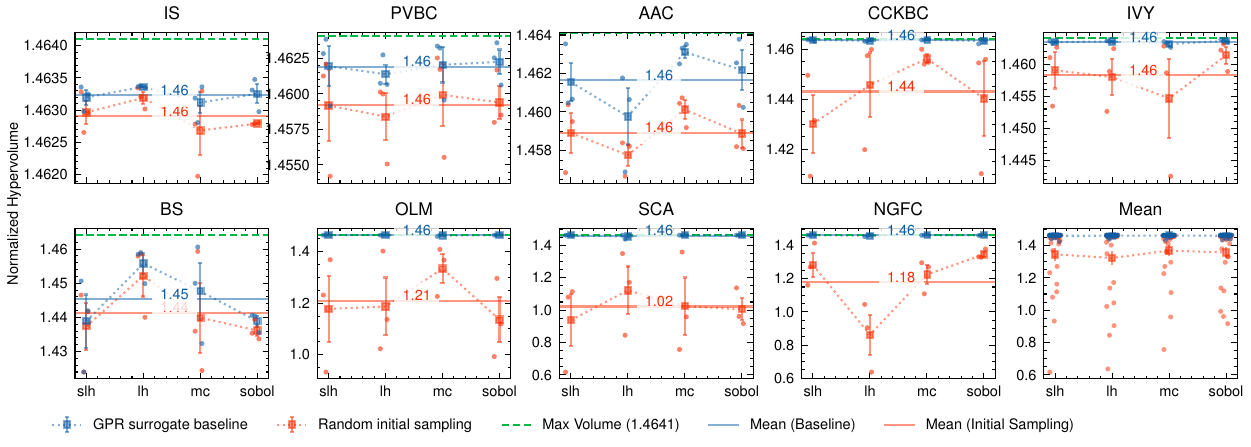}
\caption{\textbf{Per-population sampling and GPR surrogate baseline.} Normalized hypervolume for 9 CA1 interneuron populations (panels) plus the cross-population mean (rightmost panel). Each panel compares four initial sampling strategies (MC: Monte Carlo, LHC: Latin Hypercube, SLHC: Symmetric Latin Hypercube, Sobol) under two conditions: initial random sampling only (red circles) and after 25 epochs of GPR surrogate-assisted optimization (blue circles). Horizontal colored lines show the mean hypervolume across sampling strategies for each condition. The green dashed line indicates the theoretical maximum hypervolume ($1.1^4 \approx 1.4641$). Error bars represent standard deviation across $n = 3$ independent trials. The GPR surrogate consistently improves hypervolume over random sampling across all populations, with final performance varying by cell type due to differences in constraint satisfaction rates and objective landscape complexity. Markers show the mean and error bars $\pm 1$ SEM over $n = 3$ independent trials per population and strategy; dots show the individual trials. The rightmost panel pools all populations ($n = 27$ runs per strategy).}
\label{fig:samplers}
\end{figure*}

Figure~\ref{fig:samplers} shows the normalized hypervolume achieved by four initial sampling strategies - Monte Carlo (MC), Latin Hypercube (LHC), Symmetric Latin Hypercube (SLHC), and Sobol sequences - on each of the 9 CA1 interneuron populations. Two conditions are compared: (1)~the hypervolume of the initial sample set alone (red markers), representing pure random search without optimization, and (2)~the hypervolume after 25 epochs of GPR-based surrogate-assisted optimization starting from each initial sample set (blue markers). Horizontal lines indicate the mean across sampling strategies; the green dashed line marks the theoretical maximum hypervolume ($1.1^4 \approx 1.4641$).

Several observations emerge from this analysis:
\begin{itemize}
    \item \textbf{Initial sampling quality:} The choice of initial sampling strategy has a modest effect on the starting hypervolume. Symmetric Latin Hypercube (SLHC) and Sobol sequences tend to provide slightly better initial coverage than Monte Carlo, though the differences are small relative to the total optimization improvement.
    \item \textbf{Population-specific difficulty:} Populations differ substantially in baseline optimization difficulty. IS and PVBC achieve near-optimal hypervolume ($>1.46$) even with the GPR baseline, while NGFC and OLM show more modest final performance, reflecting harder constraint landscapes or more complex objective trade-offs.
    \item \textbf{Consistent improvement:} Across all populations and sampling strategies, GPR-based surrogate optimization consistently improves hypervolume over the initial random sample, demonstrating the value of surrogate-assisted search even with a simple GPR model.
\end{itemize}

Table~\ref{tab:single-cells-initial} summarizes the normalized hypervolume metrics achieved by each surrogate strategy across all CA1 cell types. The four surrogate categories compared are: GPR (Gaussian Process Regression), MEGP (multitask exact Gaussian process), ``o'' (objective-only neural network surrogates, averaging ResNet and FTTransformer), and ``c+o'' (joint constraint-and-objective neural network surrogates). Populations are sorted by increasing mean hypervolume to highlight the range of optimization difficulty across cell types.

\begin{table*}[htbp]
\centering
\caption{\textbf{CA1 single-cell surrogate comparison.} Normalized hypervolume (mean $\pm$ std across $n = 3$ trials) for each surrogate strategy and cell type after initial sampling. GPR: Gaussian Process Regression; MEGP: multitask exact Gaussian process; o: objective-only neural networks (mean of o-ResNet and o-FTTransformer); c+o: joint constraint-and-objective neural networks (mean of c+o-ResNet and c+o-FTTransformer). Higher values indicate better Pareto front quality; the theoretical maximum is $1.1^4 \approx 1.4641$. Populations are sorted by mean hypervolume (ascending). NGFC exhibits the lowest performance due to stringent constraint satisfaction requirements, while IS approaches the theoretical optimum across all methods.}
\label{tab:single-cells-initial}
\begin{tabular*}{\textwidth}{llllll}
\toprule%
 & gpr                 & megp                & o                     & c+o               & Mean   \\
\midrule
 NGFC    & $0.0112 \pm 0.0076$ & $0.0133 \pm 0.0085$ & $0.0081 \pm 0.0021$   & $0.044 \pm 0.028$ & $0.019 \pm 0.012$    \\
 BS      & $0.484 \pm 0.055$   & $0.553 \pm 0.055$   & $0.549 \pm 0.082$     & $0.520 \pm 0.046$ & $0.527 \pm 0.059$    \\
 SCA     & $0.85 \pm 0.25$     & $0.98 \pm 0.14$     & $0.81 \pm 0.10$       & $0.87 \pm 0.36$   & $0.88 \pm 0.21$      \\
 OLM     & $1.27 \pm 0.14$     & $1.22 \pm 0.20$     & $1.18 \pm 0.19$       & $1.23 \pm 0.14$   & $1.23 \pm 0.17$      \\
 PVBC    & $1.288 \pm 0.045$   & $1.266 \pm 0.068$   & $1.258 \pm 0.052$     & $1.284 \pm 0.056$ & $1.274 \pm 0.055$    \\
 AAC     & $1.364 \pm 0.026$   & $1.367 \pm 0.033$   & $1.365 \pm 0.037$     & $1.366 \pm 0.026$ & $1.366 \pm 0.031$    \\
 CCKBC   & $1.431 \pm 0.042$   & $1.422 \pm 0.031$   & $1.422 \pm 0.019$     & $1.35 \pm 0.14$   & $1.406 \pm 0.059$    \\
 IVY     & $1.441 \pm 0.030$   & $1.441 \pm 0.018$   & $1.442 \pm 0.018$     & $1.436 \pm 0.027$ & $1.440 \pm 0.023$    \\
 IS      & $1.4562 \pm 0.0030$ & $1.4564 \pm 0.0019$ & $1.45770 \pm 0.00098$ & $1.449 \pm 0.018$ & $1.4549 \pm 0.0059$  \\
 Optimal & $1.46 \pm 0$        & $1.46 \pm 0$        & $1.46 \pm 0$          & $1.46 \pm 0$      & $1.46 \pm 0$         \\
\bottomrule
\end{tabular*}
\end{table*}

\subsubsection{Per-Population Epsilon Indicator}

The additive epsilon indicator ($I_{\varepsilon+}$, \suppnote{sec:supp-metrics}) provides a complementary view to hypervolume by measuring how much each surrogate-assisted Pareto front must be shifted to dominate the no-surrogate baseline front. Negative values indicate that the surrogate front dominates the baseline; values closer to zero indicate better performance.

\begin{figure*}[htbp]
\centering
\includegraphics[width=\textwidth]{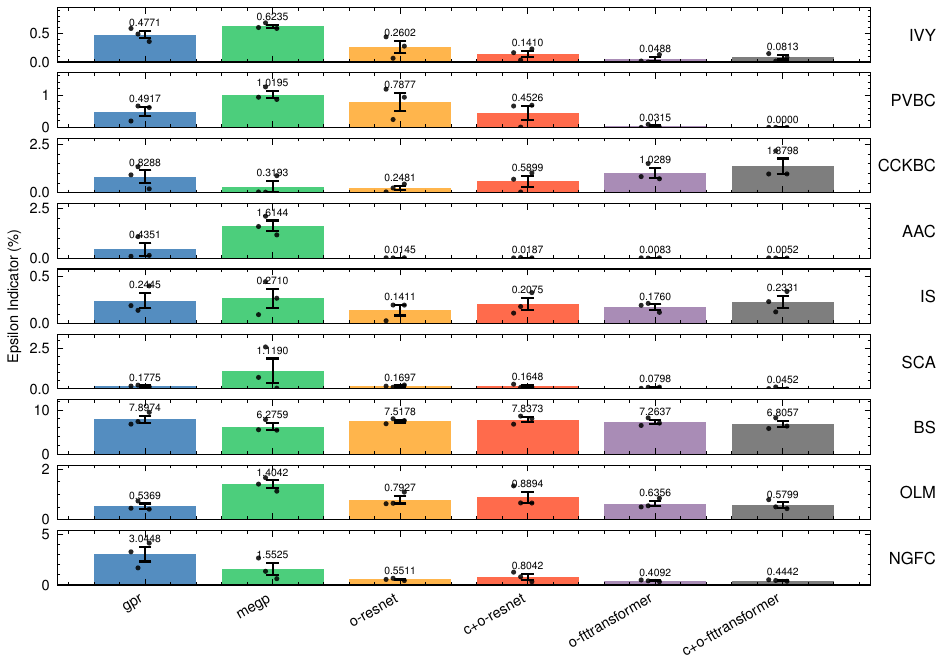}
\caption{\textbf{Per-population epsilon indicator.} Additive epsilon indicator (\%) for each surrogate method (GPR, MEGP, o-ResNet, c+o-ResNet, o-FTTransformer, c+o-FTTransformer) across all 9 CA1 interneuron populations. Each panel shows one population; bar heights indicate the mean epsilon indicator across trials, with error bars showing standard deviation. Lower values indicate better performance (the surrogate front is closer to dominating the baseline). The c+o-FTTransformer achieves consistently low epsilon values ($<2\%$) across all populations, indicating that its Pareto fronts nearly dominate those of the no-surrogate baseline. Bars show the mean $\pm 1$ SEM over $n = 3$ independent trials per population and method; dots show the individual trials.}
\label{fig:epsilon-performance}
\end{figure*}

Figure~\ref{fig:epsilon-performance} shows the epsilon indicator for each surrogate method on each CA1 interneuron population (for global summary refer to Figure 2C in the main text). The indicator is computed by comparing the normalized Pareto front from each surrogate run against the corresponding no-surrogate baseline run. Error bars represent standard deviation across the $n = 3$ independent trials.

\subsubsection{Per-Constraint Feasibility Prediction Accuracy}\label{sec:supp-constraint-accuracy}

Aggregating prediction performance over all constraints of a problem (\suppnotesub{sec:supp-ca1-constraints}) can mask accuracy differences, particularly as the feasibility rate of each constraint can vary widely. Table~\ref{tab:constraint-accuracy} therefore reports per-constraint accuracy, with the metrics computed within each cell type and averaged over the nine populations. To account for the significant class imbalance, we report accuracy in terms of Area Under the Receiver Operating Characteristic curve (AUROC) and the Matthews correlation coefficient (MCC). The breakdown reveals significant differences in the prediction performance, with firing behaviour-related constraints being predicted accurately while the remaining ones remain at or near chance level. Because AUROC can be optimistic when the positive class is rare, we note that the three constraints with high AUROC are close to balanced (48-52\% violations).

\begin{table*}[htbp]
\centering
\caption{\textbf{Per-constraint feasibility prediction accuracy.} Feasibility detection performance of the joint constraint head across the 9 CA1 interneuron populations  for both backbones fitted on the same samples (computed within cells, then averaged). The joint surrogate is re-fit offline at nine epoch checkpoints of a completed optimization run and scored on the candidates the optimizer proposed in the following epoch. ``Violations'' are infeasible samples (and percentage) for that constraint across the $18{,}299$ training samples. ``all (conjunction)'' scores the overall feasibility prediction. AUROC is threshold-free (chance = $0.5$) whereas for MCC $0$ is no better than constant prediction.}
\label{tab:constraint-accuracy}
\begin{tabular}{lr@{\hskip 12pt}rr@{\hskip 12pt}rr}
\toprule
& & \multicolumn{2}{c}{c+o-FTTransformer} & \multicolumn{2}{c}{c+o-ResNet} \\
\cmidrule(lr){3-4}\cmidrule(lr){5-6}
Constraint & Violations & AUROC & MCC & AUROC & MCC \\
\midrule
\multicolumn{6}{l}{\itshape Firing behaviour} \\
\quad Monotonic f-I & 10704 (58.5\%) & 0.875 & 0.622 & 0.847 & 0.581 \\
\quad First ISI & 11619 (63.5\%) & 0.881 & 0.652 & 0.855 & 0.616 \\
\quad ISI adaptation & 11746 (64.2\%) & 0.875 & 0.638 & 0.853 & 0.610 \\
\quad Pre-stimulus spike count & 314 (1.7\%) & 0.367 & 0.000 & 0.583 & 0.000 \\
\multicolumn{6}{l}{\itshape Measurement validity} \\
\quad $R_{\text{in}}$ range & 2127 (11.6\%) & 0.536 & 0.081 & 0.644 & 0.050 \\
\quad $\tau$ range & 2282 (12.5\%) & 0.499 & 0.030 & 0.603 & 0.004 \\
\quad Initial voltage & 1982 (10.8\%) & 0.377 & 0.000 & 0.617 & 0.000 \\
\multicolumn{6}{l}{\itshape Feature definedness} \\
\quad Spike amplitude & 0 (0.0\%) & -- & -- & -- & -- \\
\midrule
all (conjunction) & 13076 (71.5\%) & 0.852 & 0.549 & 0.828 & 0.522 \\
\bottomrule
\end{tabular}
\end{table*}

\subsubsection{Gradient Geometry of Feasibility Solving}\label{sec:supp-gradient-geometry}

The gradient-based feasibility solving (\suppnote{sec:supp-gradient}) must implicitly trade off its representation of objective and constraints. To understand what drives convergence, we can compute the gradient of each feasibility loss term with respect to the parameter input and measure the angle between them. 
For the purpose of this analysis, we re-fit the surrogate on evaluations of a completed run, as in \suppnotesub{sec:supp-constraint-accuracy}.

We find that objective and constraint gradients are positively aligned, with a mean cosine similarity of $0.61$ across the nine populations and opposing directions at only $0$--$19\%$ of candidates depending on cell type. Improving the predicted objectives and predicted feasibility therefore point in broadly compatible directions, and the joint gradient does not require trading one against the other.

For the individual constraints, on the other hand, the gradients separate into the same two groups that emerge in the prediction accuracies (Table~\ref{tab:constraint-accuracy}). The constraints describing firing behaviour align, as do those describing measurement validity, while the two groups oppose each other (Table~\ref{tab:gradient-geometry}). This points to a gradient trade-off that reflects the intrinsic characteristic of the problem specification. The practical consequence is small since the constraint probabilities along the feasibility-solving trajectory improve consistently. Note that the analysis excludes two gradients as the spike amplitude constraint was never violated and initial voltage constraint collapses to a constant prediction with zero gradient.

\begin{table*}[htbp]
\centering
\caption{\textbf{Gradient geometry of the constraint predictions.} Mean cosine similarity  $\pm$ standard deviation between the input gradients of the per-constraint feasibility losses, evaluated at the candidates the optimizer proposed and averaged over 3 random initializations $\times$ 9 populations within each group of constraints. \emph{Joint surrogate} is the \texttt{c+o-ResNet} used throughout while \emph{one model per constraint} fits a separate standalone constraint surrogate. Constraints are grouped as in Table~\ref{tab:constraint-accuracy}.}
\label{tab:gradient-geometry}
\begin{tabular}{l@{\hskip 12pt}rr}
\toprule
Constraint pairs & Joint surrogate & One model per constraint \\
\midrule
Within firing behaviour & $+0.733 \pm 0.194$ & $+0.892 \pm 0.070$ \\
Within measurement validity & $+0.894 \pm 0.059$ & $+0.698 \pm 0.166$ \\
Across the two groups & $-0.655 \pm 0.132$ & $-0.368 \pm 0.195$ \\
\bottomrule
\end{tabular}
\end{table*}

\subsubsection{Per-Population Convergence Curves}

\begin{figure*}[htbp]
\centering
\includegraphics[width=0.8\textwidth]{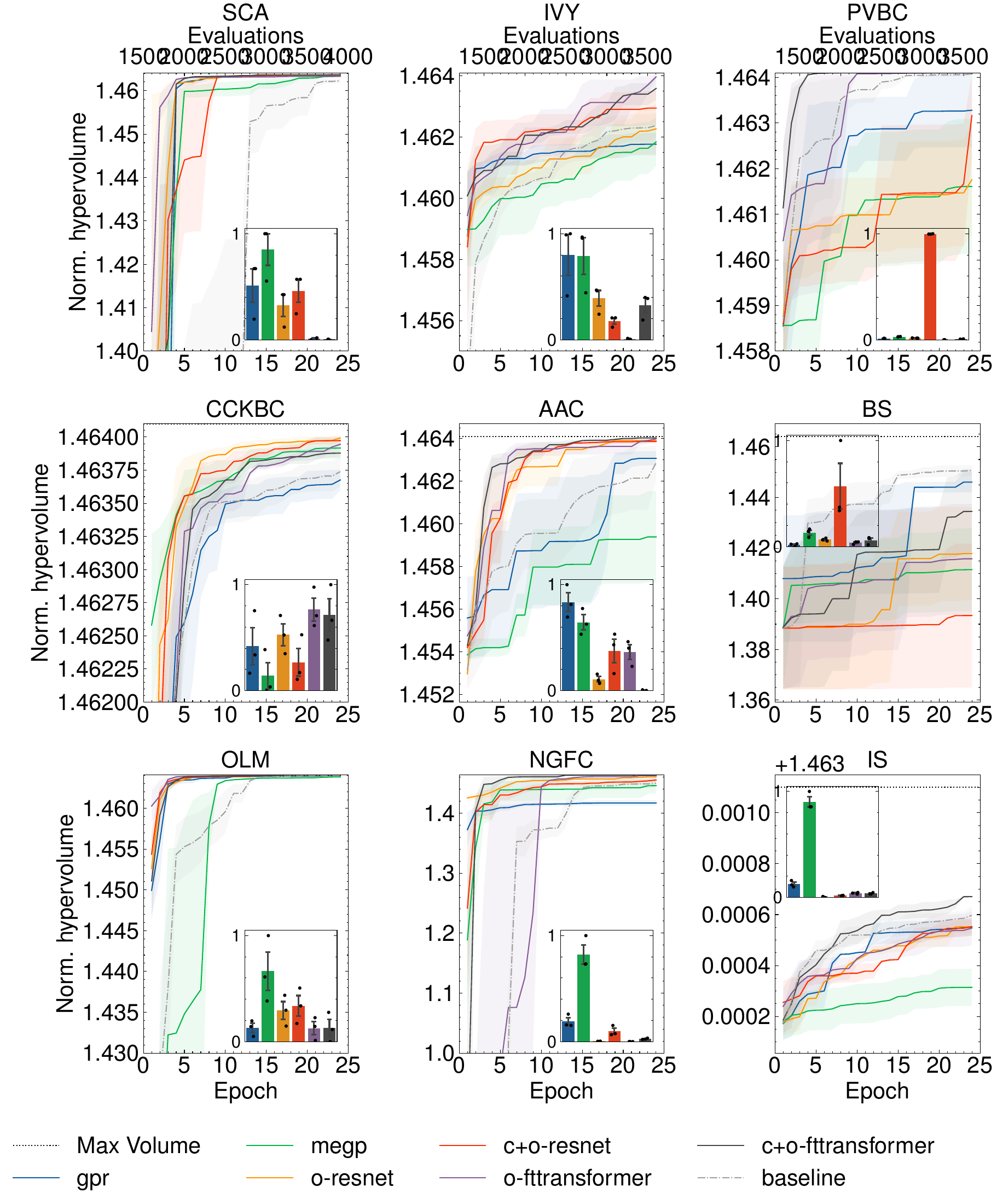}
\caption{\textbf{Per-population convergence curves for all 9 CA1 interneuron types.} $3 \times 3$ grid showing normalized hypervolume (HV) as a function of optimization epoch for SCA, IVY, PVBC, CCKBC, AAC, BS, OLM, NGFC, and IS populations. Each panel plots all 6 surrogate methods (GPR, MEGP, o-ResNet, c+o-ResNet, o-FTTransformer, c+o-FTTransformer; solid lines) and the no-surrogate baseline (dash-dotted). Shaded regions indicate standard error of the mean across replicates. Inset bar plots show the normalized inverted generational distance (IGD) for each method. The secondary $x$-axis (top) shows the cumulative number of simulation evaluations. The dashed horizontal line marks the theoretical maximum HV~($1.1^4 \approx 1.4641$). Surrogate-assisted methods consistently reach higher HV in fewer epochs across all populations, with the c+o variants achieving the fastest convergence. The only exception is the bi-stratified (BS) cell where the surrogates converge noticeably slower than the no-surrogate baseline. For an extended discussion of this failure mode and how to overcome it, see \suppnotesub{sec:bs-fail}. Lines show the mean over $n = 3$ independent trials and shaded bands $\pm 1$ SEM; inset bars show the mean min--max-normalized IGD $\pm 1$ SEM with dots for the individual trials.}
\label{fig:all-convergence}
\end{figure*}

The main text (Fig.~\ref{fig:figure3-joint-modelling}) shows representative convergence curves for selected cell types. Extended convergence curves for all 9 CA1 interneuron populations over 25 epochs are provided in Figure~\ref{fig:all-convergence}.

Tables~\ref{tab:single-cells-final-hv}, \ref{tab:single-cells-final-igd}, \ref{tab:single-cells-hv-auc} report for every population and optimization strategy the final hypervolume, normalized IGD, and HV-AUC convergence speed respectively.

\begin{table*}[htbp]
\centering
\caption{\textbf{Final normalized hypervolume by population.} Mean $\pm$ standard deviation across three trials at epoch 25 (highest in bold).}
\label{tab:single-cells-final-hv}
\resizebox{\textwidth}{!}{%
\begin{tabular}{lccccccccc}
\toprule
Model & IVY & PVBC & CCKBC & AAC & IS & SCA & BS & OLM & NGFC \\
\midrule
No surrogate & $1.4624 \pm 0.0007$ & $1.4641 \pm 0.0000$ & $1.4637 \pm 0.0002$ & $1.4629 \pm 0.0014$ & $1.4636 \pm 0.0001$ & $1.4622 \pm 0.0008$ & $\mathbf{1.4506 \pm 0.0037}$ & $1.4640 \pm 0.0001$ & $1.4499 \pm 0.0129$ \\
GPR & $1.4618 \pm 0.0006$ & $1.4633 \pm 0.0014$ & $1.4637 \pm 0.0001$ & $1.4631 \pm 0.0006$ & $1.4635 \pm 0.0001$ & $1.4635 \pm 0.0003$ & $1.4461 \pm 0.0057$ & $1.4640 \pm 0.0001$ & $1.4171 \pm 0.0078$ \\
MEGP & $1.4618 \pm 0.0014$ & $1.4616 \pm 0.0026$ & $1.4639 \pm 0.0001$ & $1.4594 \pm 0.0037$ & $1.4633 \pm 0.0001$ & $1.4634 \pm 0.0003$ & $1.4115 \pm 0.0283$ & $1.4638 \pm 0.0000$ & $1.4459 \pm 0.0234$ \\
\texttt{o}-ResNet & $1.4623 \pm 0.0014$ & $1.4618 \pm 0.0021$ & $\mathbf{1.4640 \pm 0.0000}$ & $1.4639 \pm 0.0001$ & $1.4636 \pm 0.0000$ & $1.4636 \pm 0.0001$ & $1.4178 \pm 0.0319$ & $1.4640 \pm 0.0000$ & $1.4604 \pm 0.0018$ \\
\texttt{c+o}-ResNet & $1.4629 \pm 0.0010$ & $1.4632 \pm 0.0015$ & $1.4640 \pm 0.0001$ & $1.4639 \pm 0.0002$ & $1.4636 \pm 0.0001$ & $1.4636 \pm 0.0001$ & $1.3934 \pm 0.0492$ & $1.4640 \pm 0.0001$ & $1.4553 \pm 0.0069$ \\
\texttt{o}-FTTransformer & $\mathbf{1.4640 \pm 0.0001}$ & $1.4641 \pm 0.0000$ & $1.4639 \pm 0.0000$ & $1.4640 \pm 0.0001$ & $1.4635 \pm 0.0001$ & $1.4633 \pm 0.0001$ & $1.4158 \pm 0.0375$ & $\mathbf{1.4641 \pm 0.0000}$ & $1.4625 \pm 0.0014$ \\
\texttt{c+o}-FTTransformer & $1.4636 \pm 0.0006$ & $\mathbf{1.4641 \pm 0.0000}$ & $1.4639 \pm 0.0001$ & $\mathbf{1.4640 \pm 0.0000}$ & $\mathbf{1.4637 \pm 0.0000}$ & $\mathbf{1.4637 \pm 0.0001}$ & $1.4345 \pm 0.0290$ & $1.4640 \pm 0.0001$ & $\mathbf{1.4638 \pm 0.0001}$ \\
\bottomrule
\end{tabular}%
}
\end{table*}

\begin{table*}[htbp]
\centering
\caption{\textbf{Normalized inverted generational distance by population.} Mean $\pm$ standard deviation of the no-surrogate reference-front comparisons, normalized by the Euclidean norm of the population-specific nadir and reported as a percentage (lowest in bold).}
\label{tab:single-cells-final-igd}
\resizebox{\textwidth}{!}{%
\begin{tabular}{lccccccccc}
\toprule
Model & IVY & PVBC & CCKBC & AAC & IS & SCA & BS & OLM & NGFC \\
\midrule
No surrogate & -- & -- & -- & -- & -- & -- & -- & -- & -- \\
GPR & $19.1274 \pm 7.9034$ & $0.0496 \pm 0.0097$ & $0.5835 \pm 0.3656$ & $4.3584 \pm 0.7938$ & $0.6985 \pm 0.1340$ & $5.8518 \pm 3.1411$ & $\mathbf{0.1448 \pm 0.0258}$ & $0.7515 \pm 0.2890$ & $1.5082 \pm 0.4431$ \\
MEGP & $18.8594 \pm 7.0845$ & $0.1178 \pm 0.0091$ & $\mathbf{0.2477 \pm 0.2582}$ & $3.3730 \pm 0.6306$ & $4.0507 \pm 0.3621$ & $9.7238 \pm 2.9401$ & $0.3608 \pm 0.0762$ & $2.8427 \pm 1.2167$ & $6.1695 \pm 1.1426$ \\
\texttt{o}-ResNet & $9.4390 \pm 3.0480$ & $0.0661 \pm 0.0067$ & $0.7147 \pm 0.2173$ & $0.5974 \pm 0.2301$ & $\mathbf{0.1695 \pm 0.0220}$ & $3.7030 \pm 2.0149$ & $0.2370 \pm 0.0257$ & $1.3929 \pm 0.5676$ & $0.1316 \pm 0.0031$ \\
\texttt{c+o}-ResNet & $4.4141 \pm 1.2012$ & $4.0812 \pm 0.0079$ & $0.4014 \pm 0.2736$ & $1.9805 \pm 0.9929$ & $0.2327 \pm 0.0064$ & $5.2827 \pm 2.1386$ & $1.1835 \pm 0.7002$ & $1.5669 \pm 0.6529$ & $0.8262 \pm 0.3534$ \\
\texttt{o}-FTTransformer & $\mathbf{0.3824 \pm 0.1547}$ & $\mathbf{0.0007 \pm 0.0004}$ & $1.0022 \pm 0.2273$ & $1.9209 \pm 0.6244$ & $0.3261 \pm 0.0349$ & $0.1459 \pm 0.1171$ & $0.1820 \pm 0.0258$ & $\mathbf{0.7436 \pm 0.4216}$ & $\mathbf{0.1061 \pm 0.0025}$ \\
\texttt{c+o}-FTTransformer & $7.8910 \pm 2.8374$ & $0.0240 \pm 0.0108$ & $0.9392 \pm 0.3196$ & $\mathbf{0.0561 \pm 0.0228}$ & $0.3154 \pm 0.0332$ & $\mathbf{0.0513 \pm 0.0188}$ & $0.2252 \pm 0.0677$ & $0.7555 \pm 0.5419$ & $0.2965 \pm 0.0480$ \\
\bottomrule
\end{tabular}%
}
\end{table*}

\begin{table*}[htbp]
\centering
\caption{\textbf{Hypervolume area under the curve (HV-AUC) by population.} Mean $\pm$ standard deviation through epoch 25 ($n=3$). Higher values indicate faster convergence (highest in bold).}
\label{tab:single-cells-hv-auc}
\resizebox{\textwidth}{!}{%
\begin{tabular}{lccccccccc}
\toprule
Model & IVY & PVBC & CCKBC & AAC & IS & SCA & BS & OLM & NGFC \\
\midrule
No surrogate & $35.053 \pm 0.022$ & $35.116 \pm 0.022$ & $35.095 \pm 0.021$ & $35.032 \pm 0.060$ & $35.124 \pm 0.001$ & $32.570 \pm 1.628$ & $\mathbf{34.435 \pm 0.285}$ & $34.616 \pm 0.335$ & $26.998 \pm 1.232$ \\
GPR & $35.069 \pm 0.009$ & $35.095 \pm 0.040$ & $35.107 \pm 0.016$ & $35.022 \pm 0.084$ & $35.123 \pm 0.001$ & $34.333 \pm 0.792$ & $34.107 \pm 0.629$ & $34.962 \pm 0.105$ & $33.183 \pm 0.147$ \\
MEGP & $35.050 \pm 0.042$ & $35.054 \pm 0.052$ & $\mathbf{35.121 \pm 0.000}$ & $34.972 \pm 0.046$ & $35.118 \pm 0.003$ & $34.463 \pm 0.049$ & $33.766 \pm 0.653$ & $34.651 \pm 0.133$ & $33.465 \pm 0.923$ \\
\texttt{o}-ResNet & $35.063 \pm 0.023$ & $35.063 \pm 0.038$ & $35.119 \pm 0.013$ & $35.088 \pm 0.024$ & $35.122 \pm 0.001$ & $34.735 \pm 0.369$ & $33.615 \pm 0.865$ & $34.972 \pm 0.100$ & $\mathbf{34.098 \pm 0.066}$ \\
\texttt{c+o}-ResNet & $35.088 \pm 0.029$ & $35.058 \pm 0.037$ & $35.118 \pm 0.009$ & $35.086 \pm 0.002$ & $35.122 \pm 0.001$ & $34.446 \pm 0.324$ & $33.354 \pm 1.050$ & $34.975 \pm 0.105$ & $33.679 \pm 0.496$ \\
\texttt{o}-FTTransformer & $\mathbf{35.089 \pm 0.025}$ & $35.116 \pm 0.030$ & $35.111 \pm 0.015$ & $35.093 \pm 0.025$ & $35.122 \pm 0.002$ & $\mathbf{34.827 \pm 0.151}$ & $33.771 \pm 1.061$ & $\mathbf{34.982 \pm 0.107}$ & $29.002 \pm 5.353$ \\
\texttt{c+o}-FTTransformer & $35.087 \pm 0.015$ & $\mathbf{35.131 \pm 0.004}$ & $35.114 \pm 0.011$ & $\mathbf{35.102 \pm 0.011}$ & $\mathbf{35.124 \pm 0.001}$ & $34.555 \pm 0.251$ & $33.874 \pm 0.805$ & $34.968 \pm 0.124$ & $33.649 \pm 0.751$ \\
\bottomrule
\end{tabular}%
}
\end{table*}

\subsubsection{Gradient Feasibility Solving Bi-stratified Cell}\label{sec:bs-fail}

\begin{figure*}[htbp]
\centering
\includegraphics[width=0.7\textwidth]{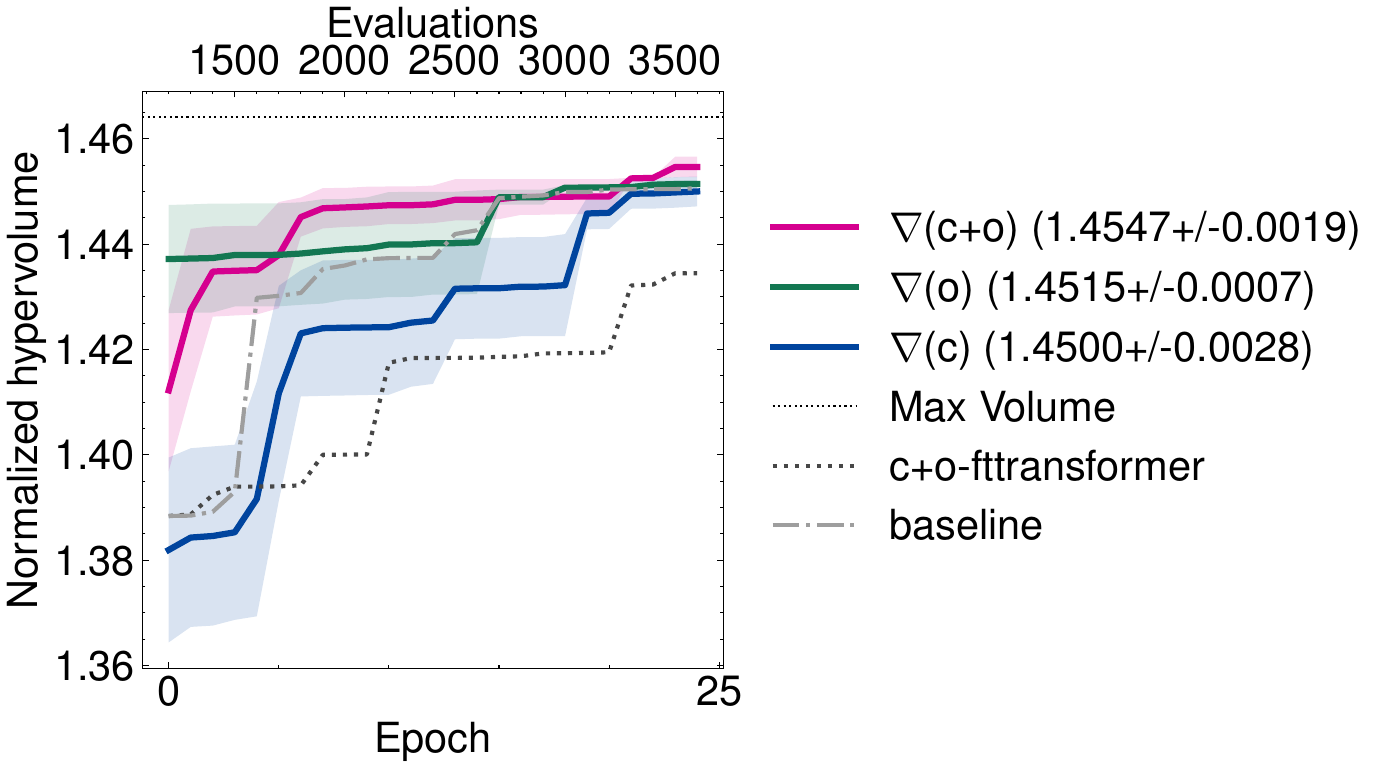}
\caption{\textbf{Gradient target comparison on the BS population.} Normalized hypervolume convergence over 25 epochs comparing three gradient descent target configurations: $\nabla$(c+o) (joint objective and constraint optimization, magenta), $\nabla$(o) (objective-only, green), and $\nabla$(c) (constraint-only, blue). The c+o-FTTransformer surrogate without gradient augmentation (dotted line) and the no-surrogate NSGA-II baseline (dash-dotted line) are included for reference. Legend entries report final HV as mean $\pm$ SEM ($n = 3$ trials). The joint $\nabla$(c+o) configuration achieves the highest final hypervolume, demonstrating that simultaneously optimizing objectives and constraints during gradient descent produces the best solutions. The constraint-only $\nabla$(c) variant converges more slowly, as it does not directly optimize the objectives. Shaded regions indicate $\pm 1$ SEM.}
\label{fig:bs-gradient}
\end{figure*}

\begin{figure*}[htbp]
\centering
\includegraphics[width=0.7\textwidth]{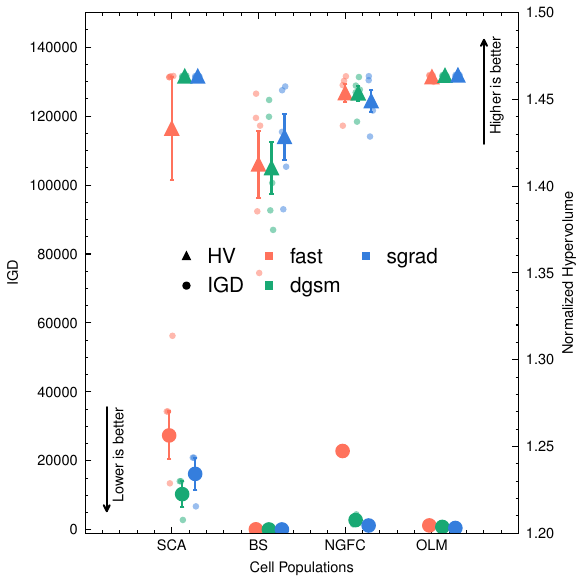}
\caption{\textbf{Sensitivity analysis method comparison across CA1 populations.} Inverted generational distance (IGD, circles, left $y$-axis; lower is better) and normalized hypervolume (HV, triangles, right $y$-axis; higher is better) for three sensitivity-informed sampling strategies: FAST (red), DGSM (green), and sgrad (blue). Each marker shows the mean and error bars $\pm 1$ SEM; dots show the individual trials ($n = 3$ trials for IGD, $n = 5$ trials for HV). The surrogate gradient method (sgrad) achieves competitive or superior IGD and HV compared to the established FAST and DGSM methods, while incurring no additional simulation evaluations - the sensitivity estimates are computed directly from the trained surrogate via automatic differentiation. Across most populations, all three methods yield similar performance, indicating that the optimization is robust to the choice of sensitivity method once sensitivity-informed sampling is enabled.}
\label{fig:sensitivity-all}
\end{figure*}

With limited number of initial samples to bootstrap the surrogate, models may struggle to accurately capture the most promising regions of the search space, leading to slower convergence than direct optimization without surrogate. This is the case for the Bi-stratified Interneuron model (BS, see Figure~\ref{fig:all-convergence}), making it a suitable test case for our gradient-based feasibility solving procedure described in \suppnote{sec:supp-gradient}.  The strategy can be configured to optimize different target combinations: objectives only ($\nabla$(o)), constraints only ($\nabla$(c)), or both jointly ($\nabla$(c+o)). Figure~\ref{fig:bs-gradient} presents the resulting convergence as measured by normalized hypervolume. Leveraging the surrogate gradient enables the surrogate optimization to converge faster than the no-surrogate baseline, confirming that surrogate-gradient guidance meaningfully improves surrogate optimization.

\subsubsection{Extended Sensitivity Analysis}

The main text (Figure~3C) compares sensitivity-informed sampling methods for the representative population Bi-stratified (BS). Here we provide the full results for the SCA, BS, NGFC, OLM interneuron populations, comparing three methods: FAST (Fourier Amplitude Sensitivity Test), DGSM (Derivative-based Global Sensitivity Measure), and sgrad (surrogate gradient sensitivity).

\subsubsection{Motoneuron Narrow-Range Surrogate Optimization}

The main text (Fig.~\ref{fig:figure4-highly-constraint}) presents surrogate-assisted optimization results on the motoneuron benchmark with an expanded (wide-range) search space. Here we show the complementary narrow-range setting, where the parameter bounds are tighter and more biologically informed.

\begin{figure*}[htbp]
\centering
\includegraphics[width=0.4\textwidth]{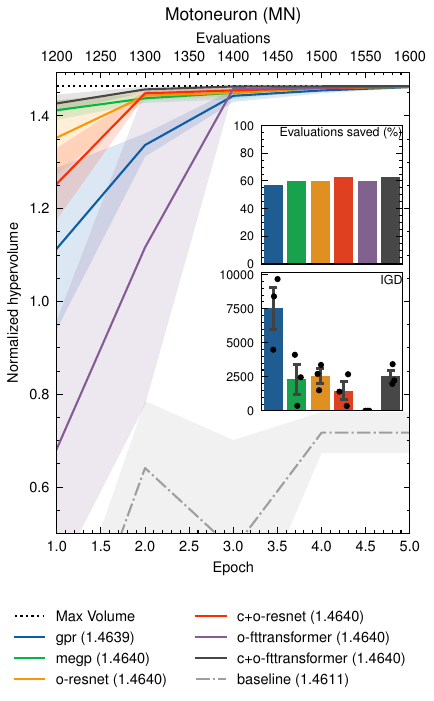}
\caption{\textbf{Surrogate optimization on the narrow-range motoneuron benchmark.} Normalized hypervolume vs. epoch for 6 surrogate methods (GPR, MEGP, o-ResNet, c+o-ResNet, o-FTTransformer, c+o-FTTransformer; solid lines) and the no-surrogate baseline (dash-dotted). The secondary $x$-axis (top) shows cumulative evaluations. Shaded regions indicate $\pm 1$ SEM across $n = 3$ independent trials; inset IGD bars show the mean $\pm 1$ SEM with dots for the individual trials. Inset bar plots show IGD (lower right) and the percentage of evaluations saved relative to the baseline (upper right). The dashed horizontal line marks the theoretical maximum HV~($1.1^4 \approx 1.4641$). All surrogate methods converge faster than the baseline, with c+o variants achieving the highest final HV.}
\label{fig:baseline-sopt}
\end{figure*}

Figure~\ref{fig:baseline-sopt} reports normalized hypervolume convergence over 25 epochs for all 6 surrogate methods and the no-surrogate NSGA-II baseline. In this easier setting, all surrogate methods reach near-optimal hypervolume within 3-5 epochs, compared to 10-15 epochs for the baseline. The inset bar plots show (i) the inverted generational distance (IGD) of each method relative to the no-surrogate Pareto front, and (ii) the percentage of simulation evaluations saved by converging earlier. Even in the narrow-range configuration, surrogate-assisted methods reduce the number of required evaluations by 40-60\%, confirming that the efficiency gains are not limited to difficult, wide-range problems.

\subsubsection{Motoneuron Random Sampling Constraint Satisfaction}

The motoneuron (MN) benchmark defines 8 biophysical constraints that candidate parameter vectors must satisfy (monotonic f-I, resistance range, time constant range, spike amplitude, first ISI, ISI adaptation, pre-spike count, initial voltage). Before surrogate-assisted optimization begins, the initial sampling phase draws parameter vectors uniformly at random from the search space. Whether a sample satisfies each constraint depends critically on the extent of the search space.

\begin{figure*}[htbp]
\centering
\includegraphics[width=\textwidth]{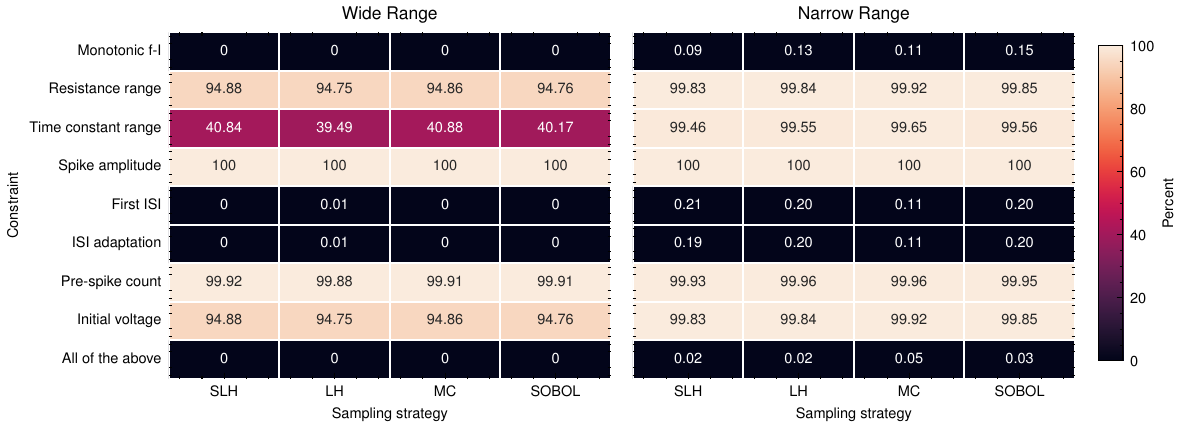}
\caption{\textbf{Constraint satisfaction under random sampling for the motoneuron benchmark.} Heatmaps showing the percentage of random samples that satisfy each individual constraint (rows) and all constraints jointly (bottom row) for four sampling strategies (columns) in wide-range (left) and narrow-range (right) search-space configurations. Cell annotations give the percentage; color scale runs from 0\% (dark) to 100\% (light). Wide-range sampling yields near-zero joint feasibility, whereas narrow-range sampling achieves 0.02--0.05\% joint satisfaction. Sampling strategy has a comparatively minor effect relative to search-space extent. Percentages are means over $n = 3$ independent sampling runs.}
\label{fig:random-constraints}
\end{figure*}

Figure~\ref{fig:random-constraints} compares constraint satisfaction rates for four sampling strategies (LHS, Saltelli, Sobol, uniform) under two search-space configurations. In the \emph{wide range} setting, only individual constraints with loose bounds (e.g., initial voltage) are frequently met; the joint satisfaction rate (``All of the above'') is below 1\%, reflecting the combinatorial difficulty of simultaneously satisfying all constraints in a large parameter space. In the \emph{narrow range} setting, per-constraint satisfaction rates are substantially higher, and joint satisfaction rises to approximately 10--20\%. The choice of sampling strategy (LHS vs.\ Sobol vs.\ Saltelli vs.\ uniform) has a relatively minor effect compared to the search-space extent, indicating that the constraint landscape - not the quasi-random sequence - is the primary bottleneck for initial feasibility. These results motivate the gradient-based feasibility-solving procedure (\suppnote{sec:supp-gradient}), which actively steers samples toward the feasible region rather than relying on random coverage.

\subsubsection{Extended Surrogate Gradient Descent Traces}

\begin{figure*}[htbp]
\centering
\includegraphics[width=0.85\textwidth]{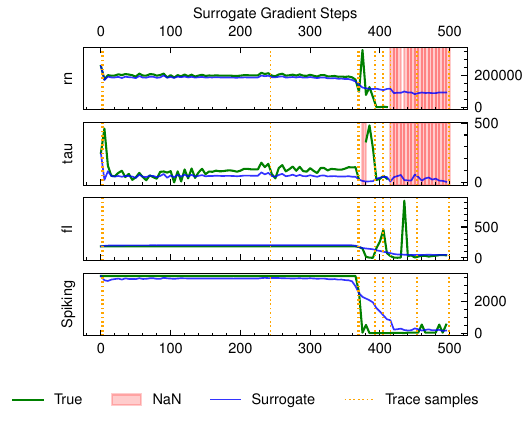}
\caption{\textbf{Extended gradient descent traces for all motoneuron objectives.} Four vertically stacked panels showing the surrogate-guided trajectory for each objective: input resistance (rn), membrane time constant (tau), f-I curve (fI), and spike amplitude (Spiking). Blue lines: surrogate predictions at each gradient step. Green lines: corresponding true simulation values. Orange vertical dotted lines: trace samples selected by the diversity filter and evaluated on the full NEURON model. Red-shaded bands: gradient steps where the true simulation returned NaN. The secondary x-axis gives the gradient step count. Close agreement between surrogate and simulation confirms that the surrogate gradient provides a reliable descent direction for all objectives simultaneously.}
\label{fig:sopt-grad-sampling}
\end{figure*}

The main text (Figure~4B) shows the surrogate-guided gradient descent trajectory for a single objective. Figure~\ref{fig:sopt-grad-sampling} extends this visualization to all four motoneuron objectives simultaneously: input resistance error, membrane time constant error, f-I curve error, and spike amplitude error. Each panel plots the surrogate-predicted objective value (blue) alongside the true simulation value (green) over gradient descent steps. Orange vertical lines mark the trace samples selected by the diversity filter (\suppnote{sec:supp-gradient}), and red-shaded regions indicate steps where the true simulation produced NaN (non-viable parameter configurations).

The close tracking between surrogate predictions and true simulation values across all four objectives demonstrates that the surrogate provides a reliable multi-objective gradient signal. Occasional divergence between predicted and true values particularly near NaN regions highlights the importance of the diversity-filtered trace sampling, which re-anchors the surrogate to informative points along the trajectory.

\subsubsection{Surrogate Gradient vs. Black-Box Optimizers}

The dmosopt framework defaults to NSGA-II as the black-box optimizer that operates on the surrogate-predicted objective landscape. An alternative approach is to bypass the black-box optimizer entirely and use \emph{direct surrogate gradient optimization} (sgrad), where candidate solutions are generated by following the surrogate gradient directly via backpropagation rather than through evolutionary search. We compare sgrad against several black-box optimizers across all surrogate architectures on the motoneuron benchmark.

\begin{figure*}[htbp]
\centering
\includegraphics[width=\textwidth]{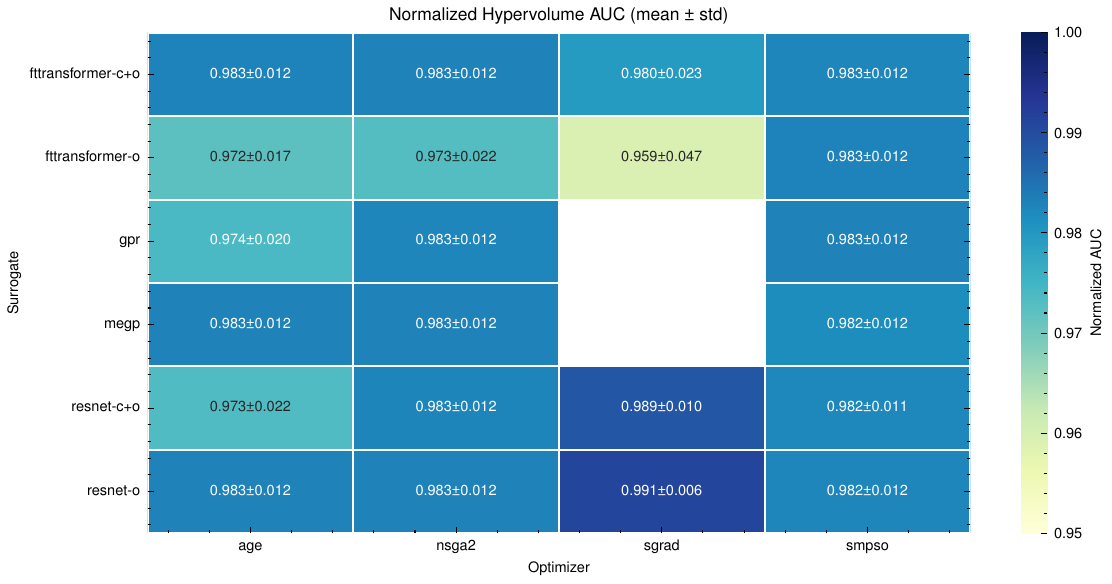}
\caption{\textbf{Surrogate gradient (sgrad) vs. black-box optimizer comparison.} Heatmap of normalized hypervolume AUC (higher is better) for each combination of surrogate architecture (rows) and optimizer (columns) on the motoneuron benchmark over 10 epochs. Cell annotations report mean $\pm$ s.d.\ across $n = 3$ independent trials. Color scale ranges from 0.95 to 1.0. ``sgrad'' denotes direct surrogate gradient optimization via backpropagation; ``nsga2'' denotes the standard NSGA-II black-box optimizer used in the main experiments. The sgrad approach achieves competitive AUC across surrogate architectures, demonstrating that for this benchmark, direct gradient-based candidate generation on the differentiable surrogate is a viable alternative to population-based black-box search. However, success of direct surrogate descent may be undermined if the solutions are found in regions of the search space the model predicts confidently wrong.}
\label{fig:sgrad-vs-optimizer}
\end{figure*}

Figure~\ref{fig:sgrad-vs-optimizer} shows the normalized hypervolume AUC (area under the convergence curve over 10 epochs) for each surrogate-optimizer combination. The sgrad approach achieves competitive or superior performance compared to conventional black-box optimizers, particularly when paired with the c+o-FTTransformer surrogate. This result demonstrates that the differentiable surrogate models trained in dmosopt provide sufficiently smooth objective landscapes for gradient-based optimization to succeed, eliminating the need for population-based search entirely in some configurations. However, the relative advantage of sgrad varies by surrogate architecture, suggesting that the smoothness and accuracy of the surrogate landscape are important prerequisites for direct gradient optimization.

\subsubsection{Training Gaussian Process Surrogates on Infeasible Samples}\label{sec:supp-gp-infeasible}

The GP baselines in the main text are fitted using feasible samples only (Methods, \emph{Gaussian process baselines}). To quantify the effect of this modeling choice, we additionally ran the MEGP surrogate using all available samples, i.e.\ with \textsc{dmosopt}'s feasibility filter switched off and otherwise identical settings. Both variants were run as full online optimizations on the nine CA1 interneuron problems (25 epochs of 100 evaluations, $n = 3$ independent trials each, sharing the initial samples of the main comparison). For reference, the tables also include a rank-1 linear model of coregionalization (LMC) with a single latent Mat\'ern kernel, fitted on the feasible subset by L-BFGS in BoTorch~\autocite{balandatBoTorchFrameworkEfficient2020}.

Tables~\ref{tab:single-cells-megp-ablation-hv}, \ref{tab:single-cells-megp-ablation-igd} and \ref{tab:single-cells-megp-ablation-hv-auc} report the final normalized hypervolume, normalized IGD and HV-AUC. The all-samples MEGP reached an equal or higher final hypervolume in every population, a lower IGD in 5 of 9 and a higher HV-AUC in 4 of 9 populations. Notably, the differences lie within one standard deviation across trials so the feasible-only restriction costs little solution quality, but comes with computational benefits. Fitting the exact GP on all samples took a median $11.6\times$ longer and prediction $8.4\times$ longer than on the feasible subset on the same data (Table~\ref{tab:megp-all-samples-timing}), reflecting the cubic cost in the number of training points. To see what the infeasible samples do to the surrogate itself, we re-fitted MEGP offline on the samples evaluated up to epochs 4 and 12 of the completed feasible-only runs (trial 0 of each population), scored on the candidates the optimizer proposed in the following epoch (Table~\ref{tab:megp-offline-ablation}). Across all candidates, the all-samples fit ranks the proposals better early on (median Spearman correlation $0.51$ versus $0.21$ at epoch 5), because it has seen the infeasible region that the feasible-only model never encounters. On the feasible candidates, however, which are the only ones that can enter the Pareto front, it is markedly less accurate with median NRMSE $1.95$ versus $0.26$ at epoch 5 and $10.6$ versus $0.49$ at epoch 13, worse in 8 of 9 populations at both checkpoints (rank correlation on feasible candidates falling to $0.05$ by epoch 13). The very large objective errors of the infeasible samples dominate the fit of the stationary kernel, and the effect grows with the size of the training set. That the online outcome is nevertheless unchanged is consistent with the optimizer relying on the surrogate mainly to rank candidates. The better discrimination of the all-samples model between feasible and infeasible regions offsets its poorer accuracy within the feasible set, at roughly ten times the fitting cost. 

Naively enlarging the GP training set is therefore not worthwhile on these problems. Mapping the objectives to a standard normal distribution or clipping the very large errors of infeasible samples might yield improvements, but would confound the effect of the training-set size with that of the transform and would reintroduce the problem-specific target engineering that the joint neural network surrogate is designed to avoid (Methods, \emph{Gaussian process baselines}).

\begin{table*}[htbp]
\centering
\caption{\textbf{Final normalized hypervolume for the MEGP ablations.} Values are mean $\pm$ standard deviation across three trials at epoch 25. The highest mean in each population is shown in bold.}
\label{tab:single-cells-megp-ablation-hv}
\resizebox{\textwidth}{!}{%
\begin{tabular}{lccccccccc}
\toprule
Model & IVY & PVBC & CCKBC & AAC & IS & SCA & BS & OLM & NGFC \\
\midrule
MEGP (feasible only) & $1.4618 \pm 0.0014$ & $1.4616 \pm 0.0026$ & $\mathbf{1.4639 \pm 0.0001}$ & $1.4594 \pm 0.0037$ & $1.4633 \pm 0.0001$ & $1.4634 \pm 0.0003$ & $1.4115 \pm 0.0283$ & $1.4638 \pm 0.0000$ & $1.4459 \pm 0.0234$ \\
MEGP (all samples) & $\mathbf{1.4623 \pm 0.0010}$ & $1.4616 \pm 0.0019$ & $1.4639 \pm 0.0001$ & $\mathbf{1.4617 \pm 0.0013}$ & $\mathbf{1.4636 \pm 0.0000}$ & $\mathbf{1.4635 \pm 0.0003}$ & $\mathbf{1.4176 \pm 0.0269}$ & $\mathbf{1.4639 \pm 0.0001}$ & $\mathbf{1.4514 \pm 0.0073}$ \\
Rank-1 LMC & $1.4615 \pm 0.0014$ & $\mathbf{1.4630 \pm 0.0009}$ & $1.4635 \pm 0.0002$ & $1.4586 \pm 0.0046$ & $1.4635 \pm 0.0001$ & $1.4634 \pm 0.0002$ & $1.4144 \pm 0.0211$ & $1.4636 \pm 0.0001$ & -- \\
\bottomrule
\end{tabular}%
}
\end{table*}

\begin{table*}[htbp]
\centering
\caption{\textbf{Normalized inverted generational distance for the MEGP ablations.} Values are mean $\pm$ standard deviation across three paired trial/reference-front comparisons, normalized by the Euclidean norm of the population-specific nadir and reported as a percentage. The lowest mean in each population is shown in bold.}
\label{tab:single-cells-megp-ablation-igd}
\resizebox{\textwidth}{!}{%
\begin{tabular}{lccccccccc}
\toprule
Model & IVY & PVBC & CCKBC & AAC & IS & SCA & BS & OLM & NGFC \\
\midrule
MEGP (feasible only) & $22.1571 \pm 8.4526$ & $0.1339 \pm 0.0362$ & $\mathbf{0.3457 \pm 0.3568}$ & $\mathbf{4.0268 \pm 1.1657}$ & $3.2049 \pm 2.3933$ & $\mathbf{4.6331 \pm 6.0008}$ & $0.2789 \pm 0.0894$ & $\mathbf{2.4267 \pm 0.7917}$ & $1.9597 \pm 3.0832$ \\
MEGP (all samples) & $\mathbf{11.8538 \pm 8.6729}$ & $\mathbf{0.1183 \pm 0.0518}$ & $0.9549 \pm 0.7954$ & $4.0327 \pm 0.6143$ & $\mathbf{0.4153 \pm 0.1145}$ & $7.8556 \pm 5.8246$ & $\mathbf{0.1849 \pm 0.0199}$ & $4.4460 \pm 3.2251$ & $\mathbf{1.1435 \pm 1.2977}$ \\
Rank-1 LMC & $12.6766 \pm 3.4653$ & $0.1432 \pm 0.0105$ & $0.5509 \pm 0.4716$ & $4.6108 \pm 0.9797$ & $2.6264 \pm 3.4252$ & $8.0956 \pm 2.0168$ & $0.2512 \pm 0.0764$ & $4.4332 \pm 3.1316$ & -- \\
\bottomrule
\end{tabular}%
}
\end{table*}

\begin{table*}[htbp]
\centering
\caption{\textbf{Hypervolume area under the curve for the MEGP ablations.} Values are mean $\pm$ standard deviation across three trials through epoch 25. Higher values indicate faster convergence; the highest mean in each population is shown in bold.}
\label{tab:single-cells-megp-ablation-hv-auc}
\resizebox{\textwidth}{!}{%
\begin{tabular}{lccccccccc}
\toprule
Model & IVY & PVBC & CCKBC & AAC & IS & SCA & BS & OLM & NGFC \\
\midrule
MEGP (feasible only) & $35.050 \pm 0.042$ & $35.054 \pm 0.052$ & $\mathbf{35.121 \pm 0.000}$ & $\mathbf{34.972 \pm 0.046}$ & $35.118 \pm 0.003$ & $34.463 \pm 0.049$ & $\mathbf{33.766 \pm 0.653}$ & $34.651 \pm 0.133$ & $33.465 \pm 0.923$ \\
MEGP (all samples) & $\mathbf{35.072 \pm 0.014}$ & $35.052 \pm 0.048$ & $35.106 \pm 0.032$ & $34.969 \pm 0.048$ & $\mathbf{35.124 \pm 0.001}$ & $34.297 \pm 0.573$ & $33.671 \pm 0.750$ & $\mathbf{34.826 \pm 0.156}$ & $\mathbf{33.822 \pm 0.197}$ \\
Rank-1 LMC & $35.054 \pm 0.018$ & $\mathbf{35.079 \pm 0.029}$ & $35.088 \pm 0.032$ & $34.971 \pm 0.074$ & $35.120 \pm 0.003$ & $\mathbf{34.801 \pm 0.189}$ & $33.585 \pm 0.793$ & $34.515 \pm 0.088$ & -- \\
\bottomrule
\end{tabular}%
}
\end{table*}

\begin{table*}[htbp]
\centering
\caption{\textbf{MEGP compute cost by training-set policy.} Fit and prediction times are mean $\pm$ standard deviation in seconds across the benchmarked population--epoch pairs. Slowdowns are medians of the paired ratios relative to feasible-only MEGP over the $n = 9$ population datasets (one fit per population at the final checkpoint of the Figure~\ref{fig:figure2-surrogate-opt} inference analysis).}
\label{tab:megp-all-samples-timing}
\begin{tabular}{lcccc}
\toprule
Model & Fit (s) & Prediction (s) & Fit slowdown & Prediction slowdown \\
\midrule
MEGP (feasible only) & $35.21 \pm 29.36$ & $0.35 \pm 0.32$ & -- & -- \\
MEGP (all samples) & $330.69 \pm 68.53$ & $2.00 \pm 0.33$ & $11.6\times$ & $8.4\times$ \\
\bottomrule
\end{tabular}
\end{table*}

\begin{table*}[htbp]
\centering
\caption{\textbf{Offline prediction accuracy of MEGP by training-set policy.} At each checkpoint epoch the surrogate is re-fitted on all samples evaluated up to that epoch of the completed MEGP runs, either on the feasible subset only (the default) or on all samples including infeasible ones, and scored on the candidates the optimizer proposed in the following epoch. NRMSE (lower is better) and Spearman rank correlation (higher is better) are reported on all finite test points and on the feasible test points only; values are the median [interquartile range] over $n = 9$ population--trial pairs (9 populations $\times$ 1 trial). $N_{\text{train}}$ is the mean number of training samples used by each ablation.}
\label{tab:megp-offline-ablation}
\small
\begin{tabular}{llrcccc}
\toprule
& & & \multicolumn{2}{c}{All test points} & \multicolumn{2}{c}{Feasible test points} \\
\cmidrule(lr){4-5} \cmidrule(lr){6-7}
Epoch & Training set & $N_{\text{train}}$ & NRMSE $\downarrow$ & Spearman $\uparrow$ & NRMSE $\downarrow$ & Spearman $\uparrow$ \\
\midrule
5 & MEGP (feasible only) & 372 & $0.44$ [0.32, 0.56] & $0.21$ [0.13, 0.27] & $0.26$ [0.21, 0.50] & $0.46$ [0.34, 0.54] \\
 & MEGP (all samples) & 1533 & $0.36$ [0.20, 0.67] & $0.51$ [0.30, 0.55] & $1.95$ [1.44, 3.68] & $0.31$ [0.20, 0.44] \\
\addlinespace
13 & MEGP (feasible only) & 607 & $0.61$ [0.40, 1.15] & $0.24$ [0.23, 0.37] & $0.49$ [0.34, 1.20] & $0.55$ [0.45, 0.58] \\
 & MEGP (all samples) & 2333 & $3.97$ [1.47, 9.90] & $0.29$ [0.28, 0.52] & $10.60$ [4.30, 82.92] & $0.05$ [-0.02, 0.28] \\
\bottomrule
\end{tabular}
\end{table*}

\subsubsection{Comparison with Constrained Bayesian Optimization}\label{sec:supp-cbo}

Constrained Bayesian optimization (BO) is the established prior art for expensive optimization under unknown constraints~\autocite{gardnerBayesianOptimizationInequality2014,gelbartBayesianOptimizationUnknown2014,hernandez-lobatoPredictiveEntropySearch2015}. It structurally differs in its use of parallel hardware. \textsc{dmosopt} produces all 100 candidates of an epoch in a single NSGA-II pass over the surrogate, followed by a single batch of gradient refinement, enabling concurrent simulation. BO assembles a batch of $q$ candidates by $q$ \emph{successive} acquisition optimizations, limiting parallelism~\autocite{wilsonMaximizingAcquisitionFunctions2018,daultonParallelBayesianOptimization2021}. Reducing $q$ restores the sample efficiency for which BO is designed but serializes the simulations, which for the network problem would leave all but one of 100 workers of 300 cores each idle. No single configuration is therefore fair in terms of compute budget, and we instead compare optimization outcomes across three batch sizes that span the trade-off.
On the wide range, no BO managed to escape the infeasible initial design across three trials at each of $q=1$, $16$ and $100$. None of the up to 2{,}400 new evaluations satisfied all eight constraints simultaneously, and the hypervolume never rose above the initial $0.032 \pm 0.007$.

On the narrow range, where each constraint is individually satisfiable, the same implementation converged to the theoretical maximum hypervolume. The first feasible sample took $126$ new evaluations on average at $q=1$ (range $27$--$251$), $805$ at $q=16$ ($64$--$1{,}936$), and $933$ at $q=100$ ($400$--$1{,}300$), against $200$--$400$ evaluations for the black-box strategies to reach $99\%$ of the maximum.

\begin{table*}[htbp]
\centering
\caption{\textbf{Constrained Bayesian optimization vs. black-box strategies.} Hypervolume area under the curve (HV-AUC; higher is faster convergence) and final normalized hypervolume. Values are mean $\pm$ standard deviation across trials (best in bold). The $q=1$ case runs a reduced budget by construction, so it is scored over its own shorter window (marked $^{\dagger}$) and excluded from the bold comparison. For reference, the last two rows ($^{\ddagger}$) also list the constraint-gradient-augmented strategies of Figure~\ref{fig:figure4-highly-constraint}A, only run on the wide range configuration ($n = 3$ trials). Since they use the surrogate gradient, which none of the other methods do, they are also excluded from the bold comparison.}
\label{tab:motoneuron-constrained-bo}
\resizebox{\textwidth}{!}{%
\begin{tabular}{lcccc}
\toprule
& \multicolumn{2}{c}{Narrow range (MN)} & \multicolumn{2}{c}{Wide range (MN-r)} \\
\cmidrule(lr){2-3} \cmidrule(lr){4-5}
Method & HV-AUC $\uparrow$ & Final HV $\uparrow$ & HV-AUC $\uparrow$ & Final HV $\uparrow$ \\
\midrule
Constrained BO ($q=1$) & $2.983 \pm 1.114$$^{\dagger}$ & $1.4126 \pm 0.0726$$^{\dagger}$ & $0.129 \pm 0.026$$^{\dagger}$ & $0.0322 \pm 0.0066$$^{\dagger}$ \\
Constrained BO ($q=16$) & $5.538 \pm 7.831$ & $0.7320 \pm 1.0352$ & $0.773 \pm 0.159$ & $0.0322 \pm 0.0066$ \\
Constrained BO ($q=100$) & $5.077 \pm 5.468$ & $0.9417 \pm 0.8172$ & $0.773 \pm 0.159$ & $0.0322 \pm 0.0066$ \\
\midrule
GPR & $14.911 \pm 0.351$ & $1.4624 \pm 0.0024$ & $1.420 \pm 0.141$ & $0.0638 \pm 0.0044$ \\
MEGP & $15.332 \pm 0.101$ & $1.4640 \pm 0.0000$ & $17.431 \pm 8.416$ & $1.0236 \pm 0.6091$ \\
\texttt{o}-ResNet & $15.278 \pm 0.252$ & $1.4640 \pm 0.0001$ & $\mathbf{18.050 \pm 10.715}$ & $\mathbf{1.4493 \pm 0.0246}$ \\
\texttt{c+o}-ResNet & $15.190 \pm 0.192$ & $1.4639 \pm 0.0000$ & $15.668 \pm 13.099$ & $0.9209 \pm 0.7502$ \\
\texttt{o}-FT-Transformer & $14.292 \pm 0.933$ & $\mathbf{1.4640 \pm 0.0000}$ & $11.383 \pm 8.792$ & $0.7956 \pm 0.7016$ \\
\texttt{c+o}-FT-Transformer & $\mathbf{15.385 \pm 0.140}$ & $1.4640 \pm 0.0000$ & $1.715 \pm 0.465$ & $0.1022 \pm 0.0419$ \\
No surrogate & $9.524 \pm 1.135$ & $1.3450 \pm 0.1053$ & $12.082 \pm 8.704$ & $0.9844 \pm 0.4181$ \\
\midrule
$\partial x_c$-\texttt{c+o}-FT-Transformer$^{\ddagger}$ & -- & -- & $27.240 \pm 5.250$ & $1.4640 \pm 0.0002$ \\
$\partial x_c$-\texttt{c}-ResNet$^{\ddagger}$ & -- & -- & $30.555 \pm 3.060$ & $1.4640 \pm 0.0001$ \\
\bottomrule
\end{tabular}%
}
\end{table*}

\subsubsection{Parameter Importance Analysis}

The surrogate gradient sensitivity analysis (\suppnote{sec:supp-sensitivity}) yields per-parameter importance scores that quantify how strongly each model parameter influences the objectives. These scores are derived from the NSGA-II crossover distribution indices ($\eta_j^{\text{cross}}$), which are set proportionally to the absolute sensitivity: parameters with higher importance receive larger distribution indices, resulting in smaller perturbations during crossover to preserve good values.

\begin{figure*}[htbp]
\centering
\includegraphics[width=0.9\textwidth]{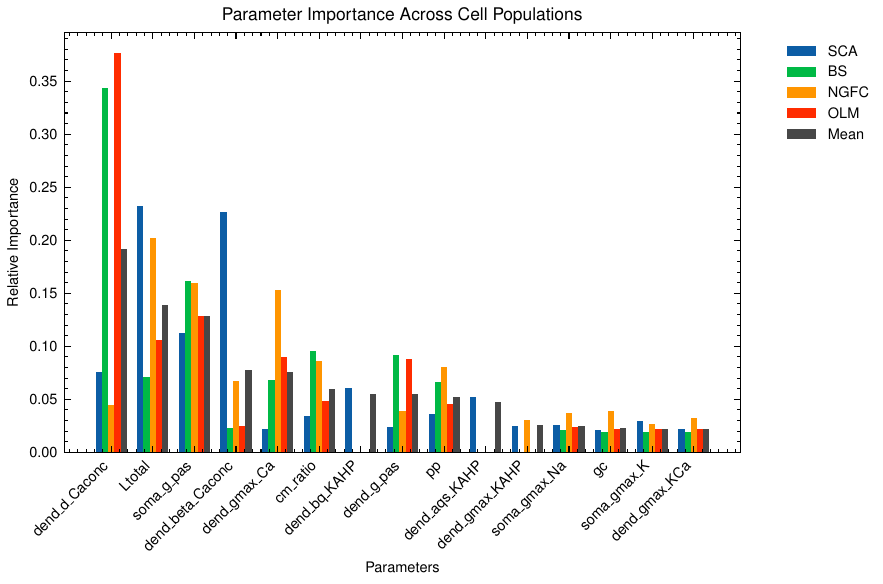}
\caption{\textbf{Parameter importance across CA1 interneuron populations.} Relative importance of each model parameter, computed from the surrogate gradient sensitivity analysis as the normalized mean crossover distribution index. Parameters are sorted by mean importance (black bars) across all populations. Colored bars show importance for individual populations (SCA, BS, NGFC, OLM). The sensitivity analysis identifies coupling conductance and compartment length as the most influential parameters, consistent with the known biophysics of these two-compartment models.}
\label{fig:param-importance}
\end{figure*}

\begin{figure*}[htbp]
\centering
\includegraphics[width=0.55\textwidth]{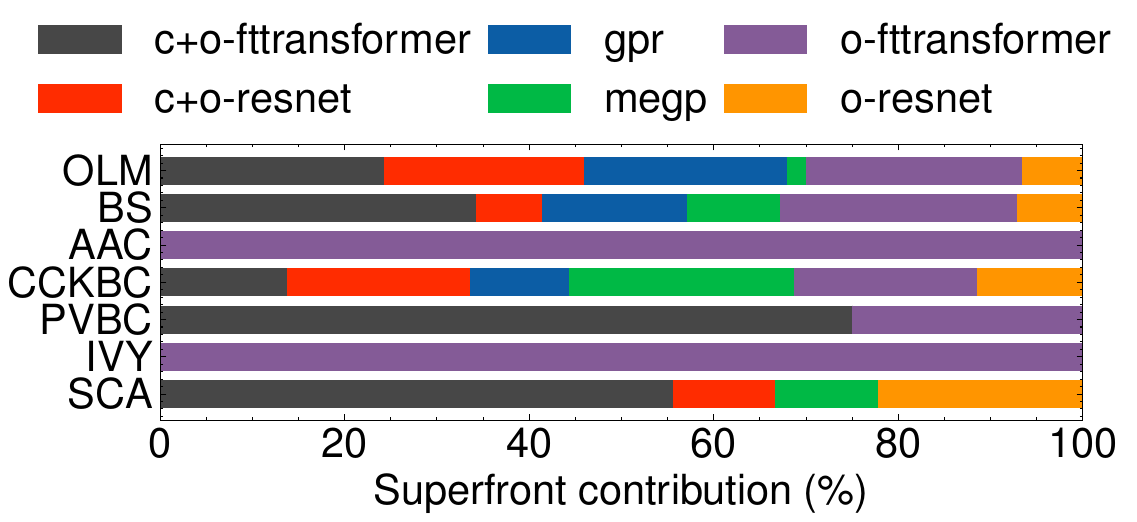}
\caption{\textbf{Superfront composition across CA1 interneuron populations.} Stacked horizontal bar chart showing the percentage contribution of each surrogate method (GPR, MEGP, o-ResNet, c+o-ResNet, o-FTTransformer, c+o-FTTransformer) to the merged non-dominated superfront for each cell type. Longer bars indicate a greater share of Pareto-optimal solutions originating from that method. The c+o-FTTransformer consistently contributes the largest fraction across populations, reflecting its ability to produce high-quality solutions that remain competitive even when compared against solutions from all other methods.}
\label{fig:superfront}
\end{figure*}

Figure~\ref{fig:param-importance} shows the relative importance of each parameter across four representative CA1 interneuron populations (SCA, BS, NGFC, OLM), computed as the mean distribution index normalized to sum to 1. The coupling conductance (\texttt{gc}) and somatic sodium conductance (\texttt{soma\_gmax\_Na}) consistently emerge as the most important parameters across all populations, reflecting their dominant role in shaping action potential generation and propagation between compartments. In contrast, calcium-related parameters (\texttt{dend\_gmax\_Ca}, \texttt{dend\_beta\_Caconc}) and passive leak conductances show lower importance, suggesting that the optimization is less sensitive to their precise values within the search bounds.

\subsubsection{Superfront Composition Analysis}

To quantify which surrogate strategy produces the most competitive solutions across all methods, we compute the superfront, i.e. the combined non-dominated front obtained by merging Pareto-optimal solutions from all surrogate methods for each cell population. The contribution of each method is measured as the fraction of superfront members originating from that method.

For each of the 7 CA1 interneuron populations analyzed (SCA, IVY, PVBC, CCKBC, AAC, BS, OLM; NGFC and IS are excluded because NGFC exhibits near-zero hypervolume, making superfront analysis uninformative, and IS already approaches the theoretical maximum across all methods):
\begin{enumerate}
    \item Collect Pareto-optimal solutions from all 6 surrogate experiments (GPR, MEGP, o-ResNet, c+o-ResNet, o-FTTransformer, c+o-FTTransformer).
    \item Merge into a single point set and compute non-dominated sorting.
    \item Record the originating method for each superfront member.
    \item Report the percentage contribution of each method.
\end{enumerate}

\FloatBarrier

\onecolumn

\printbibliography

@article{amentUnexpectedImprovementsExpected2023,
  title = {Unexpected Improvements to Expected Improvement for Bayesian Optimization},
  author = {Ament, Sebastian and Daulton, Samuel and Eriksson, David and Balandat, Maximilian and Bakshy, Eytan},
  year = 2023,
  journal = {Advances in neural information processing systems},
  volume = {36},
  pages = {20577--20612},
  urldate = {2026-09-06}
}

@article{atherton_bifurcation_2016,
  title = {Bifurcation Analysis of a Two-Compartment Hippocampal Pyramidal Cell Model},
  author = {Atherton, Laura A. and Prince, Luke Y. and {Tsaneva-Atanasova}, Krasimira},
  year = 2016,
  month = aug,
  journal = {Journal of Computational Neuroscience},
  volume = {41},
  number = {1},
  pages = {91--106},
  issn = {0929-5313, 1573-6873},
  doi = {10.1007/s10827-016-0606-8},
  urldate = {2021-11-25},
  langid = {english}
}

@inproceedings{balandatBoTorchFrameworkEfficient2020,
  title = {{{BoTorch}}: {{A Framework}} for {{Efficient Monte-Carlo Bayesian Optimization}}},
  shorttitle = {{{BoTorch}}},
  booktitle = {Advances in {{Neural Information Processing Systems}}},
  author = {Balandat, Maximilian and Karrer, Brian and Jiang, Daniel and Daulton, Samuel and Letham, Ben and Wilson, Andrew G and Bakshy, Eytan},
  year = 2020,
  volume = {33},
  pages = {21524--21538},
  publisher = {Curran Associates, Inc.},
  urldate = {2026-02-14}
}

@article{bezaireInterneuronalMechanismsHippocampal2016,
  title = {Interneuronal Mechanisms of Hippocampal Theta Oscillations in a Full-Scale Model of the Rodent {{CA1}} Circuit},
  author = {Bezaire, Marianne J. and Raikov, Ivan and Burk, Kelly and Vyas, Dhrumil and Soltesz, Ivan},
  year = 2016,
  journal = {elife},
  volume = {5},
  pages = {e18566},
  publisher = {eLife Sciences Publications, Ltd},
  urldate = {2026-02-14}
}

@article{bhosekarAdvancesSurrogateBased2018,
  title = {Advances in Surrogate Based Modeling, Feasibility Analysis, and Optimization: {{A}} Review},
  shorttitle = {Advances in Surrogate Based Modeling, Feasibility Analysis, and Optimization},
  author = {Bhosekar, Atharv and Ierapetritou, Marianthi},
  year = 2018,
  journal = {Computers \& Chemical Engineering},
  volume = {108},
  pages = {250--267},
  publisher = {Elsevier},
  urldate = {2026-02-14}
}

@article{boothCompartmentalModelVertebrate1997,
  title = {Compartmental {{Model}} of {{Vertebrate Motoneurons}} for {{Ca2}}+-{{Dependent Spiking}} and {{Plateau Potentials Under Pharmacological Treatment}}},
  author = {Booth, Victoria and Rinzel, John and Kiehn, Ole},
  year = 1997,
  month = dec,
  journal = {Journal of Neurophysiology},
  volume = {78},
  number = {6},
  pages = {3371--3385},
  publisher = {American Physiological Society},
  issn = {0022-3077},
  doi = {10.1152/jn.1997.78.6.3371},
  urldate = {2025-05-22}
}

@inproceedings{chenGradnormGradientNormalization2018,
  title = {Gradnorm: {{Gradient}} Normalization for Adaptive Loss Balancing in Deep Multitask Networks},
  shorttitle = {Gradnorm},
  booktitle = {International Conference on Machine Learning},
  author = {Chen, Zhao and Badrinarayanan, Vijay and Lee, Chen-Yu and Rabinovich, Andrew},
  year = 2018,
  pages = {794--803},
  publisher = {PMLR},
  urldate = {2026-02-14}
}

@article{chughSurrogateAssistedReferenceVector2018,
  title = {A {{Surrogate-Assisted Reference Vector Guided Evolutionary Algorithm}} for {{Computationally Expensive Many-Objective Optimization}}},
  author = {Chugh, Tinkle and Jin, Yaochu and Miettinen, Kaisa and Hakanen, Jussi and Sindhya, Karthik},
  year = 2018,
  month = feb,
  journal = {IEEE Transactions on Evolutionary Computation},
  volume = {22},
  number = {1},
  pages = {129--142},
  issn = {1941-0026},
  doi = {10.1109/TEVC.2016.2622301},
  urldate = {2026-02-14}
}

@article{daultonParallelBayesianOptimization2021,
  title = {Parallel Bayesian Optimization of Multiple Noisy Objectives with Expected Hypervolume Improvement},
  author = {Daulton, Samuel and Balandat, Maximilian and Bakshy, Eytan},
  year = 2021,
  journal = {Advances in neural information processing systems},
  volume = {34},
  pages = {2187--2200},
  urldate = {2026-09-06}
}

@article{debFastElitistMultiobjective2002a,
  title = {A Fast and Elitist Multiobjective Genetic Algorithm: {{NSGA-II}}},
  shorttitle = {A Fast and Elitist Multiobjective Genetic Algorithm},
  author = {Deb, Kalyanmoy and Pratap, Amrit and Agarwal, Sameer and Meyarivan, {\relax TAMT}},
  year = 2002,
  journal = {IEEE transactions on evolutionary computation},
  volume = {6},
  number = {2},
  pages = {182--197},
  publisher = {Ieee},
  urldate = {2025-05-22}
}

@article{druckmannNovelMultipleObjective2007,
  title = {A Novel Multiple Objective Optimization Framework for Constraining Conductance-Based Neuron Models by Experimental Data},
  author = {Druckmann, Shaul and Banitt, Yoav and Gidon, Albert A. and Sch{\"u}rmann, Felix and Markram, Henry and Segev, Idan},
  year = 2007,
  month = oct,
  journal = {Frontiers in Neuroscience},
  volume = {1},
  publisher = {Frontiers},
  issn = {1662-453X},
  doi = {10.3389/neuro.01.1.1.001.2007},
  urldate = {2026-02-14},
  langid = {english}
}

@article{dura-bernalNetPyNEToolDatadriven2019,
  title = {{{NetPyNE}}, a Tool for Data-Driven Multiscale Modeling of Brain Circuits},
  author = {{Dura-Bernal}, Salvador and Suter, Benjamin A and Gleeson, Padraig and Cantarelli, Matteo and Quintana, Adrian and Rodriguez, Facundo and Kedziora, David J and Chadderdon, George L and Kerr, Cliff C and Neymotin, Samuel A and McDougal, Robert A and Hines, Michael and Shepherd, Gordon MG and Lytton, William W},
  editor = {Bhalla, Upinder Singh and Calabrese, Ronald L and Sterratt, David and W{\'o}jcik, Daniel K},
  year = 2019,
  month = apr,
  journal = {eLife},
  volume = {8},
  pages = {e44494},
  publisher = {eLife Sciences Publications, Ltd},
  issn = {2050-084X},
  doi = {10.7554/eLife.44494},
  urldate = {2026-02-14}
}

@book{forresterEngineeringDesignSurrogate2008,
  title = {Engineering {{Design}} via {{Surrogate Modelling}}: {{A Practical Guide}}},
  shorttitle = {Engineering {{Design}} via {{Surrogate Modelling}}},
  author = {Forrester, Alexander and S{\'o}bester, Andr{\'a}s and Keane, Andy},
  year = 2008,
  month = sep,
  publisher = {John Wiley \& Sons},
  googlebooks = {ulMHmeMnRCcC},
  isbn = {978-0-470-77079-5},
  langid = {english}
}

@article{ganaieEnsembleDeepLearning2022,
  title = {Ensemble Deep Learning: {{A}} Review},
  shorttitle = {Ensemble Deep Learning},
  author = {Ganaie, Mudasir A. and Hu, Minghui and Malik, Ashwani Kumar and Tanveer, Muhammad and Suganthan, Ponnuthurai N.},
  year = 2022,
  journal = {Engineering Applications of Artificial Intelligence},
  volume = {115},
  pages = {105151},
  publisher = {Elsevier},
  urldate = {2026-02-14}
}

@inproceedings{gardnerBayesianOptimizationInequality2014,
  title = {Bayesian Optimization with Inequality Constraints.},
  booktitle = {{{ICML}}},
  author = {Gardner, Jacob R. and Kusner, Matt J. and Xu, Zhixiang Eddie and Weinberger, Kilian Q. and Cunningham, John P.},
  year = 2014,
  volume = {2014},
  pages = {937--945},
  urldate = {2026-02-14}
}

@article{gardnerGpytorchBlackboxMatrixmatrix2018,
  title = {Gpytorch: {{Blackbox}} Matrix-Matrix Gaussian Process Inference with Gpu Acceleration},
  shorttitle = {Gpytorch},
  author = {Gardner, Jacob and Pleiss, Geoff and Weinberger, Kilian Q. and Bindel, David and Wilson, Andrew G.},
  year = 2018,
  journal = {Advances in neural information processing systems},
  volume = {31},
  urldate = {2026-02-15}
}

@misc{garnelo_neural_2018,
  title = {Neural {{Processes}}},
  author = {Garnelo, Marta and Schwarz, Jonathan and Rosenbaum, Dan and Viola, Fabio and Rezende, Danilo J. and Eslami, S. M. Ali and Teh, Yee Whye},
  year = 2018,
  month = jul,
  number = {arXiv:1807.01622},
  eprint = {1807.01622},
  primaryclass = {cs.LG},
  publisher = {arXiv},
  doi = {10.48550/arXiv.1807.01622},
  urldate = {2026-09-27},
  archiveprefix = {arXiv}
}

@misc{gelbartBayesianOptimizationUnknown2014,
  title = {Bayesian {{Optimization}} with {{Unknown Constraints}}},
  author = {Gelbart, Michael A. and Snoek, Jasper and Adams, Ryan P.},
  year = 2014,
  month = mar,
  number = {arXiv:1403.5607},
  eprint = {1403.5607},
  primaryclass = {stat.ML},
  publisher = {arXiv},
  doi = {10.48550/arXiv.1403.5607},
  urldate = {2026-09-06},
  archiveprefix = {arXiv}
}

@phdthesis{gibbs_bayesian_1998,
  title = {Bayesian {{Gaussian}} Processes for Regression and Classification},
  author = {Gibbs, Mark Nicholas},
  year = 1998,
  urldate = {2026-09-07},
  school = {University of Cambridge Cambridge, UK}
}

@article{gomez-bombarelliAutomaticChemicalDesign2018,
  title = {Automatic {{Chemical Design Using}} a {{Data-Driven Continuous Representation}} of {{Molecules}}},
  author = {{G{\'o}mez-Bombarelli}, Rafael and Wei, Jennifer N. and Duvenaud, David and {Hern{\'a}ndez-Lobato}, Jos{\'e} Miguel and {S{\'a}nchez-Lengeling}, Benjam{\'i}n and Sheberla, Dennis and {Aguilera-Iparraguirre}, Jorge and Hirzel, Timothy D. and Adams, Ryan P. and {Aspuru-Guzik}, Al{\'a}n},
  year = 2018,
  month = feb,
  journal = {ACS Central Science},
  volume = {4},
  number = {2},
  pages = {268--276},
  issn = {2374-7943, 2374-7951},
  doi = {10.1021/acscentsci.7b00572},
  urldate = {2026-02-14},
  copyright = {http://pubs.acs.org/page/policy/authorchoice\_termsofuse.html},
  langid = {english}
}

@misc{goodfellow_deep_2016,
  title = {Deep Learning},
  author = {Goodfellow, Ian},
  year = 2016,
  publisher = {MIT press},
  urldate = {2024-10-01}
}

@inproceedings{gorishniyRevisitingDeepLearning2021,
  title = {Revisiting {{Deep Learning Models}} for {{Tabular Data}}},
  booktitle = {Advances in {{Neural Information Processing Systems}}},
  author = {Gorishniy, Yury and Rubachev, Ivan and Khrulkov, Valentin and Babenko, Artem},
  year = 2021,
  volume = {34},
  pages = {18932--18943},
  publisher = {Curran Associates, Inc.},
  urldate = {2024-10-02}
}

@article{griffithsConstrainedBayesianOptimization2020,
  title = {Constrained {{Bayesian}} Optimization for Automatic Chemical Design Using Variational Autoencoders},
  author = {Griffiths, Ryan-Rhys and {Hern{\'a}ndez-Lobato}, Jos{\'e} Miguel},
  year = 2020,
  journal = {Chemical science},
  volume = {11},
  number = {2},
  pages = {577--586},
  publisher = {Royal Society of Chemistry},
  urldate = {2026-02-14}
}

@incollection{hansenCMAEvolutionStrategy2006,
  title = {The {{CMA Evolution Strategy}}: {{A Comparing Review}}},
  shorttitle = {The {{CMA Evolution Strategy}}},
  booktitle = {Towards a {{New Evolutionary Computation}}},
  author = {Hansen, Nikolaus},
  editor = {Lozano, Jose A. and Larra{\~n}aga, Pedro and Inza, I{\~n}aki and Bengoetxea, Endika},
  year = 2006,
  volume = {192},
  pages = {75--102},
  publisher = {Springer Berlin Heidelberg},
  address = {Berlin, Heidelberg},
  issn = {1434-9922},
  doi = {10.1007/3-540-32494-1_4},
  urldate = {2026-02-14},
  isbn = {978-3-540-29006-3 978-3-540-32494-2},
  langid = {english}
}

@inproceedings{hernandez-lobatoPredictiveEntropySearch2015,
  title = {Predictive {{Entropy Search}} for {{Bayesian Optimization}} with {{Unknown Constraints}}},
  booktitle = {Proceedings of the 32nd {{International Conference}} on {{Machine Learning}}},
  author = {{Hernandez-Lobato}, Jose Miguel and Gelbart, Michael and Hoffman, Matthew and Adams, Ryan and Ghahramani, Zoubin},
  year = 2015,
  month = jun,
  pages = {1699--1707},
  publisher = {PMLR},
  issn = {1938-7228},
  urldate = {2026-02-14},
  langid = {english}
}

@incollection{jaeger_neuron_2022,
  title = {{{NEURON Simulation Environment}}},
  booktitle = {Encyclopedia of {{Computational Neuroscience}}},
  author = {Hines, Michael and Carnevale, Ted and McDougal, Robert A.},
  editor = {Jaeger, Dieter and Jung, Ranu},
  year = 2022,
  pages = {2355--2361},
  publisher = {Springer New York},
  address = {New York, NY},
  doi = {10.1007/978-1-0716-1006-0_795},
  urldate = {2025-05-07},
  isbn = {978-1-0716-1004-6 978-1-0716-1006-0},
  langid = {english}
}

@article{jinSurrogateassistedEvolutionaryComputation2011,
  title = {Surrogate-Assisted Evolutionary Computation: {{Recent}} Advances and Future Challenges},
  shorttitle = {Surrogate-Assisted Evolutionary Computation},
  author = {Jin, Yaochu},
  year = 2011,
  journal = {Swarm and Evolutionary Computation},
  volume = {1},
  number = {2},
  pages = {61--70},
  publisher = {Elsevier},
  urldate = {2026-02-14}
}

@article{jonesEfficientGlobalOptimization1998,
  title = {Efficient {{Global Optimization}} of {{Expensive Black-Box Functions}}},
  author = {Jones, Donald R. and Schonlau, Matthias and Welch, William J.},
  year = 1998,
  month = dec,
  journal = {Journal of Global Optimization},
  volume = {13},
  number = {4},
  pages = {455--492},
  issn = {0925-5001, 1573-2916},
  doi = {10.1023/A:1008306431147},
  urldate = {2026-02-14},
  copyright = {https://www.springernature.com/gp/researchers/text-and-data-mining},
  langid = {english}
}

@article{knowlesParEGOHybridAlgorithm2006,
  title = {{{ParEGO}}: {{A}} Hybrid Algorithm with on-Line Landscape Approximation for Expensive Multiobjective Optimization Problems},
  shorttitle = {{{ParEGO}}},
  author = {Knowles, Joshua},
  year = 2006,
  journal = {IEEE transactions on evolutionary computation},
  volume = {10},
  number = {1},
  pages = {50--66},
  publisher = {IEEE},
  urldate = {2026-02-14}
}

@book{kozielSurrogateBasedModelingOptimization2013,
  title = {Surrogate-{{Based Modeling}} and {{Optimization}}: {{Applications}} in {{Engineering}}},
  shorttitle = {Surrogate-{{Based Modeling}} and {{Optimization}}},
  editor = {Koziel, Slawomir and Leifsson, Leifur},
  year = 2013,
  publisher = {Springer New York},
  address = {New York, NY},
  doi = {10.1007/978-1-4614-7551-4},
  urldate = {2026-02-14},
  copyright = {https://www.springernature.com/gp/researchers/text-and-data-mining},
  isbn = {978-1-4614-7550-7 978-1-4614-7551-4},
  langid = {english}
}

@article{liuWhenGaussianProcess2020,
  title = {When {{Gaussian}} Process Meets Big Data: {{A}} Review of Scalable {{GPs}}},
  shorttitle = {When {{Gaussian}} Process Meets Big Data},
  author = {Liu, Haitao and Ong, Yew-Soon and Shen, Xiaobo and Cai, Jianfei},
  year = 2020,
  journal = {IEEE transactions on neural networks and learning systems},
  volume = {31},
  number = {11},
  pages = {4405--4423},
  publisher = {IEEE},
  urldate = {2026-02-14}
}

@article{lookmanActiveLearningMaterials2019,
  title = {Active Learning in Materials Science with Emphasis on Adaptive Sampling Using Uncertainties for Targeted Design},
  author = {Lookman, Turab and Balachandran, Prasanna V. and Xue, Dezhen and Yuan, Ruihao},
  year = 2019,
  journal = {npj Computational Materials},
  volume = {5},
  number = {1},
  pages = {21},
  publisher = {Nature Publishing Group UK London},
  urldate = {2026-02-14}
}

@article{markramReconstructionSimulationNeocortical2015,
  title = {Reconstruction and Simulation of Neocortical Microcircuitry},
  author = {Markram, Henry and Muller, Eilif and Ramaswamy, Srikanth and Reimann, Michael W. and Abdellah, Marwan and Sanchez, Carlos Aguado and Ailamaki, Anastasia and {Alonso-Nanclares}, Lidia and Antille, Nicolas and Arsever, Selim},
  year = 2015,
  journal = {Cell},
  volume = {163},
  number = {2},
  pages = {456--492},
  publisher = {Elsevier},
  urldate = {2026-02-14}
}

@article{mcraeGlobalSensitivityAnalysis1982,
  title = {Global Sensitivity Analysis---a Computational Implementation of the {{Fourier}} Amplitude Sensitivity Test ({{FAST}})},
  author = {McRae, Gregory J. and Tilden, James W. and Seinfeld, John H.},
  year = 1982,
  journal = {Computers \& Chemical Engineering},
  volume = {6},
  number = {1},
  pages = {15--25},
  publisher = {Elsevier},
  urldate = {2026-02-14}
}

@article{milesFunctionalPropertiesMotoneurons2004,
  title = {Functional {{Properties}} of {{Motoneurons Derived}} from {{Mouse Embryonic Stem Cells}}},
  author = {Miles, Gareth B. and Yohn, Damien C. and Wichterle, Hynek and Jessell, Thomas M. and Rafuse, Victor F. and Brownstone, Robert M.},
  year = 2004,
  month = sep,
  journal = {Journal of Neuroscience},
  volume = {24},
  number = {36},
  pages = {7848--7858},
  publisher = {Society for Neuroscience},
  issn = {0270-6474, 1529-2401},
  doi = {10.1523/JNEUROSCI.1972-04.2004},
  urldate = {2025-08-26},
  chapter = {Development/Plasticity/Repair},
  copyright = {Copyright \copyright{} 2004 Society for Neuroscience 0270-6474/04/247848-11.00/0},
  langid = {english},
  pmid = {15356197}
}

@inproceedings{nebroSMPSONewPSObased2009,
  title = {{{SMPSO}}: {{A}} New {{PSO-based}} Metaheuristic for Multi-Objective Optimization},
  shorttitle = {{{SMPSO}}},
  booktitle = {2009 {{IEEE Symposium}} on {{Computational Intelligence}} in {{Milti-Criteria Decision-Making}}},
  author = {Nebro, A.J. and Durillo, J.J. and {Garcia-Nieto}, J. and Coello Coello, C.A. and Luna, F. and Alba, E.},
  year = 2009,
  month = mar,
  pages = {66--73},
  publisher = {IEEE},
  address = {Nashville, TN, USA},
  doi = {10.1109/MCDM.2009.4938830},
  urldate = {2021-11-30},
  isbn = {978-1-4244-2764-2}
}

@inproceedings{paciorek_nonstationary_2003,
  title = {Nonstationary {{Covariance Functions}} for {{Gaussian Process Regression}}},
  booktitle = {Advances in {{Neural Information Processing Systems}}},
  author = {Paciorek, Christopher and Schervish, Mark},
  year = 2003,
  volume = {16},
  publisher = {MIT Press},
  urldate = {2026-09-07}
}

@inproceedings{panichella_adaptive_2019,
  title = {An Adaptive Evolutionary Algorithm Based on Non-Euclidean Geometry for Many-Objective Optimization},
  booktitle = {Proceedings of the {{Genetic}} and {{Evolutionary Computation Conference}}},
  author = {Panichella, Annibale},
  year = 2019,
  month = jul,
  pages = {595--603},
  publisher = {ACM},
  address = {Prague Czech Republic},
  doi = {10.1145/3321707.3321839},
  urldate = {2021-11-25},
  isbn = {978-1-4503-6111-8},
  langid = {english}
}

@article{pedregosaScikitlearnMachineLearning2011,
  title = {Scikit-Learn: {{Machine}} Learning in {{Python}}},
  shorttitle = {Scikit-Learn},
  author = {Pedregosa, Fabian and Varoquaux, Ga{\"e}l and Gramfort, Alexandre and Michel, Vincent and Thirion, Bertrand and Grisel, Olivier and Blondel, Mathieu and Prettenhofer, Peter and Weiss, Ron and Dubourg, Vincent},
  year = 2011,
  journal = {the Journal of machine Learning research},
  volume = {12},
  pages = {2825--2830},
  publisher = {JMLR. org},
  urldate = {2026-09-04}
}

@article{pinskyIntrinsicNetworkRhythmogenesis1994a,
  title = {Intrinsic and Network Rhythmogenesis in a Reduced Traub Model for {{CA3}} Neurons},
  author = {Pinsky, Paul F. and Rinzel, John},
  year = 1994,
  month = jun,
  journal = {Journal of Computational Neuroscience},
  volume = {1},
  number = {1-2},
  pages = {39--60},
  publisher = {{Springer Science and Business Media LLC}},
  issn = {0929-5313, 1573-6873},
  doi = {10.1007/bf00962717},
  urldate = {2025-05-22},
  copyright = {http://www.springer.com/tdm},
  langid = {english}
}

@article{prinzAlternativeHandTuningConductanceBased2003,
  title = {Alternative to {{Hand-Tuning Conductance-Based Models}}: {{Construction}} and {{Analysis}} of {{Databases}} of {{Model Neurons}}},
  shorttitle = {Alternative to {{Hand-Tuning Conductance-Based Models}}},
  author = {Prinz, Astrid A. and Billimoria, Cyrus P. and Marder, Eve},
  year = 2003,
  month = dec,
  journal = {Journal of Neurophysiology},
  volume = {90},
  number = {6},
  pages = {3998--4015},
  issn = {0022-3077, 1522-1598},
  doi = {10.1152/jn.00641.2003},
  urldate = {2026-02-14},
  langid = {english}
}

@article{sensoyEvidentialDeepLearning2018,
  title = {Evidential Deep Learning to Quantify Classification Uncertainty},
  author = {Sensoy, Murat and Kaplan, Lance and Kandemir, Melih},
  year = 2018,
  journal = {Advances in neural information processing systems},
  volume = {31},
  urldate = {2026-02-14}
}

@article{shahriariTakingHumanOut2015,
  title = {Taking the Human out of the Loop: {{A}} Review of {{Bayesian}} Optimization},
  shorttitle = {Taking the Human out of the Loop},
  author = {Shahriari, Bobak and Swersky, Kevin and Wang, Ziyu and Adams, Ryan P. and De Freitas, Nando},
  year = 2015,
  journal = {Proceedings of the IEEE},
  volume = {104},
  number = {1},
  pages = {148--175},
  publisher = {IEEE},
  urldate = {2026-02-14}
}

@article{snelson_warped_2003,
  title = {Warped Gaussian Processes},
  author = {Snelson, Edward and Ghahramani, Zoubin and Rasmussen, Carl},
  year = 2003,
  journal = {Advances in neural information processing systems},
  volume = {16},
  urldate = {2026-09-07}
}

@inproceedings{snoek_scalable_2015,
  title = {Scalable {{Bayesian Optimization Using Deep Neural Networks}}},
  booktitle = {Proceedings of the 32nd {{International Conference}} on {{Machine Learning}}},
  author = {Snoek, Jasper and Rippel, Oren and Swersky, Kevin and Kiros, Ryan and Satish, Nadathur and Sundaram, Narayanan and Patwary, Mostofa and Prabhat, Mr and Adams, Ryan},
  year = 2015,
  month = jun,
  pages = {2171--2180},
  publisher = {PMLR},
  issn = {1938-7228},
  urldate = {2026-09-07},
  langid = {english}
}

@article{sobolDerivativeBasedGlobal2010,
  title = {Derivative Based Global Sensitivity Measures},
  author = {Sobol, Ilya M. and Kucherenko, Sergei},
  year = 2010,
  journal = {Procedia-Social and Behavioral Sciences},
  volume = {2},
  number = {6},
  pages = {7745--7746},
  publisher = {Elsevier},
  urldate = {2026-02-14}
}

@inproceedings{springenberg_bayesian_2016,
  title = {Bayesian {{Optimization}} with {{Robust Bayesian Neural Networks}}},
  booktitle = {Advances in {{Neural Information Processing Systems}}},
  author = {Springenberg, Jost Tobias and Klein, Aaron and Falkner, Stefan and Hutter, Frank},
  year = 2016,
  volume = {29},
  publisher = {Curran Associates, Inc.},
  urldate = {2026-09-07}
}

@article{upadhyayMiVSimulatorComputationalFramework2023,
  title = {{{MiV-Simulator}}: {{A}} Computational Framework to Simulate Exact Scale in-Vitro Neuronal Networks},
  shorttitle = {{{MiV-Simulator}}},
  author = {Upadhyay, Gaurav and Kim, Seung Hyun and Gressmann, Frithjof and Raikov, Ivan and Dou, Zhi and Zhang, Xiaotian and Soltesz, Ivan and Rauchwerger, Lawrence and Gazzola, Mattia},
  year = 2023,
  month = jan,
  volume = {2023},
  pages = {D11.010},
  urldate = {2024-04-01}
}

@article{vangeitBluePyOptLeveragingOpen2016,
  title = {{{BluePyOpt}}: {{Leveraging Open Source Software}} and {{Cloud Infrastructure}} to {{Optimise Model Parameters}} in {{Neuroscience}}},
  shorttitle = {{{BluePyOpt}}},
  author = {Van Geit, Werner and Gevaert, Michael and Chindemi, Giuseppe and R{\"o}ssert, Christian and Courcol, Jean-Denis and Muller, Eilif B. and Sch{\"u}rmann, Felix and Segev, Idan and Markram, Henry},
  year = 2016,
  month = jun,
  journal = {Frontiers in Neuroinformatics},
  volume = {10},
  publisher = {Frontiers},
  issn = {1662-5196},
  doi = {10.3389/fninf.2016.00017},
  urldate = {2026-02-14},
  langid = {english}
}

@incollection{voutchkov_multi-objective_2010,
  title = {Multi-{{Objective Optimization Using Surrogates}}},
  booktitle = {Computational {{Intelligence}} in {{Optimization}}: {{Applications}} and {{Implementations}}},
  author = {Voutchkov, Ivan and Keane, Andy},
  editor = {Tenne, Yoel and Goh, Chi-Keong},
  year = 2010,
  pages = {155--175},
  publisher = {Springer},
  address = {Berlin, Heidelberg},
  doi = {10.1007/978-3-642-12775-5_7},
  urldate = {2024-07-23},
  isbn = {978-3-642-12775-5},
  langid = {english}
}

@book{williams_gaussian_2006,
  title = {Gaussian Processes for Machine Learning},
  author = {Williams, Christopher KI and Rasmussen, Carl Edward},
  year = 2006,
  volume = {2},
  publisher = {MIT press Cambridge, MA},
  urldate = {2026-09-07}
}

@inproceedings{wilson_bayesian_2020,
  title = {Bayesian {{Deep Learning}} and a {{Probabilistic Perspective}} of {{Generalization}}},
  booktitle = {Advances in {{Neural Information Processing Systems}}},
  author = {Wilson, Andrew G and Izmailov, Pavel},
  year = 2020,
  volume = {33},
  pages = {4697--4708},
  publisher = {Curran Associates, Inc.},
  urldate = {2026-09-07}
}

@inproceedings{wilson_deep_2016,
  title = {Deep Kernel Learning},
  booktitle = {Artificial Intelligence and Statistics},
  author = {Wilson, Andrew Gordon and Hu, Zhiting and Salakhutdinov, Ruslan and Xing, Eric P.},
  year = 2016,
  pages = {370--378},
  publisher = {PMLR},
  urldate = {2026-09-07}
}

@article{wilsonMaximizingAcquisitionFunctions2018,
  title = {Maximizing Acquisition Functions for {{Bayesian}} Optimization},
  author = {Wilson, James and Hutter, Frank and Deisenroth, Marc},
  year = 2018,
  journal = {Advances in neural information processing systems},
  volume = {31},
  urldate = {2026-09-06}
}

@software{gressmann2026msoCode,
  author    = {Gressmann, Frithjof and Raikov, Ivan Georgiev and Kim, Seung Hyun and Gazzola, Mattia and Rauchwerger, Lawrence and Soltesz, Ivan},
  title     = {{MultiObjectiveSurrogateOptimization}: analysis code and figure data for ``Joint surrogate learning of objectives, constraints, and sensitivities for efficient multi-objective optimization of neural dynamical systems'' (this paper)},
  year      = {2026},
  publisher = {Zenodo},
  version   = {1.0.0},
  doi       = {10.5281/zenodo.23000796},
  url       = {https://github.com/GazzolaLab/MultiObjectiveSurrogateOptimization}
}

\end{document}